\documentclass[twoside]{article}
\usepackage[accepted]{aistats2024}
\usepackage[round]{natbib}

\usepackage{amsmath,amssymb,amsfonts}
\usepackage[pdftex]{graphicx}
\usepackage{caption}
\usepackage{subcaption}
\usepackage{xspace}
\usepackage{multirow}
\usepackage{booktabs}
\usepackage{bbm}
\usepackage{microtype}
\usepackage{mathtools}
\usepackage{xcolor}
\usepackage{url}

\newtheorem{theorem}{\textbf{Theorem}}
\newtheorem{lemma}{\textbf{Lemma}}
\newtheorem{definition}{Definition}

\makeatletter
\def\@opargbegintheorem#1#2#3{\trivlist
   \item[]{\bfseries #1\ #2\ (#3)} \itshape}
\makeatother

\newtheorem{corollary}{\textbf{Corollary}}
\makeatletter
\def\@opargbegintheorem#1#2#3{\trivlist
   \item[]{\bfseries #1\ #2\ (#3)} \itshape}
\makeatother

\newcommand{\thetaretrain}{\theta_{D'}^*}
\newcommand{\thetaunlearn}{\theta_{D'}^-}
\newcommand{\thetafull}{\theta_{D}^*}

\begin{document}

%
\runningtitle{Fair Machine Unlearning}

%

\twocolumn[

\aistatstitle{Fair Machine Unlearning: \\ Data Removal while Mitigating Disparities}

\aistatsauthor{Alex Oesterling \And Jiaqi Ma \And Flavio P. Calmon \And Himabindu Lakkaraju }

\aistatsaddress{Harvard University\And UIUC\And Harvard University\And Harvard University}

]

\begin{abstract}
The Right to be Forgotten is a core principle outlined by regulatory frameworks such as the EU's General Data Protection Regulation (GDPR). This principle allows individuals to request that their personal data be deleted from deployed machine learning models. While ``forgetting" can be naively achieved by retraining on the remaining dataset, it is computationally expensive to do to so with each new request. As such, several \textit{machine unlearning} methods have been proposed as efficient alternatives to retraining. These methods aim to approximate the predictive performance of retraining, but fail to consider how unlearning impacts other properties critical to real-world applications such as fairness. In this work, we demonstrate that most efficient unlearning methods cannot accommodate popular fairness interventions, and we propose the first \emph{fair machine unlearning} method that can  efficiently unlearn data instances from a fair objective. We derive theoretical results which demonstrate that our method can provably unlearn data and provably maintain fairness performance. Extensive experimentation with real-world datasets highlight the efficacy of our method at unlearning data instances while preserving fairness. Code is provided at \url{https://github.com/AI4LIFE-GROUP/fair-unlearning}.
\end{abstract}

\section{INTRODUCTION}

Machine learning applications, ranging from recommender systems to credit scoring, frequently involve training sophisticated models using large quantities of personal data. As a means to prevent misuse of such information, recent regulatory policies have included provisions\footnote{For example, Article 17 in the European Union's General Data Protection Regulation (GDPR)~\citep{GDPR} or Section~1798.105 in the California Consumer Privacy Act (CCPA)~\citep{CCPA}.} that allow data subjects to delete their information from databases and models used by organizations. These policies have given rise to a general regulatory principle known as the \emph{right to be forgotten}. However, achieving this principle in practice is non-trivial; for instance, the naive approach of retraining a model from scratch each time a deletion request occurs is computationally prohibitive. As a result, a variety of \emph{machine unlearning} (or simply \emph{unlearning}) methods have been developed to effectively approximate the retraining process with reduced computational costs~\citep{guoCertifiedDataRemoval2020,bourtouleMachineUnlearning2020,izzoApproximateDataDeletion,neelDescenttoDeleteGradientBasedMethods2020a}.

In critical real-world applications that use personal data we often seek to achieve other key properties, such as algorithmic fairness, in addition to unlearning. For example, if a social media recommender system systematically under-ranks posts from a particular demographic group, it could substantially suppress the public voices of that group~\citep{beutel2019fairness, jiang2019degenerate}. Similarly, if a machine learning model used for credit scoring exhibits bias, it could unjustly deny loans to certain demographic groups~\citep{corbett2018measure}. In medicine, the fair allocation of healthcare services is also critical to prevent the exacerbation of systemic racism and improve the quality of care for marginalized groups~\citep{chen2021algorithm}. Both protection against algorithmic discrimination and the right to be forgotten have been spelled out in various regulatory documents~\citep{aibillofrights,GDPR,CCPA}, highlighting a need to achieve them simultaneously in practice.

To adhere to the above mentioned regulatory principles, practitioners must satisfy the right to be forgotten while still preventing discrimination in deployed settings. However, most prior work in unlearning only considers models optimized for predictive performance, and fails to consider additional objectives. In fact, we find that most existing machine unlearning methods are not compatible with common fairness interventions. Specifically, popular unlearning methods~\citep{guoCertifiedDataRemoval2020,neelDescenttoDeleteGradientBasedMethods2020a, izzoApproximateDataDeletion} provide theoretical analysis over training objectives that are convex and can be decomposed into sums of losses on individual training data points. On the other hand, existing fairness interventions generally have nonconvex objectives \citep{lowyStochasticOptimizationFramework2022} or involve pairwise or group comparisons between samples that make unlearning a datapoint nontrivial due to each instance's entangled influence on all other samples in the objective function\footnote{See Appendix~\ref{sec:incompatible} for a detailed explanation.}~\citep{berk2017convex, beutel2019fairness}. As such, current unlearning methods cannot be naively combined with fairness objectives, highlighting a need from real-world practitioners for a method that achieves fairness and unlearning simultaneously.

To fill in this critical gap, we formalize the problem of fair unlearning, and propose the first \emph{fair unlearning} method
that can provably unlearn over a fairness-constrained objective. Our fair unlearning approach takes a convex fairness regularizer with pairwise comparisons and unrolls these comparisons into an unlearnable form while ensuring the preservation of theoretical guarantees on unlearning. We provide theoretical results that our method can achieve a common notion of unlearning, \emph{statistical indistinguishability}~\citep{guoCertifiedDataRemoval2020,neelDescenttoDeleteGradientBasedMethods2020a}, while preserving a measure of \emph{equality of odds}~\citep{hardt2016equality}, a popular group fairness metric. Furthermore, we empirically evaluate the proposed method on a variety of real-world datasets and verify its effectiveness across various data deletion request settings, such as deletion at random and deletion from minority and majority subgroups. We show that our method well approximates retraining over fair machine learning objectives in terms of both accuracy and fairness. 

Our main contributions include:
\begin{itemize}
    \item We formalize the problem of fair unlearning, i.e., how to develop models we can train fairly and unlearn from.
    \item We introduce the first fair unlearning method that can provably unlearn requested data points while preserving popular notions of fairness.
    \item We provide theoretical analysis at the intersection of fairness and unlearning that can be applied to a broad set of unlearning methods.
    \item We empirically verify the effectiveness of the proposed method through extensive experiments on a variety of real-world datasets and across different settings of data deletion requests.
\end{itemize}
 
\section{RELATED WORK}

\paragraph{Unlearning.}

The naive approach to machine unlearning is to retrain a model from scratch with each data deletion request. However, retraining is not feasible for companies with many large models or organizations with limited resources. Thus, the primary objective of machine unlearning is to provide efficient approximations to retraining. Early approaches in security and privacy attempt to achieve exact removal, where an unlearned model is identical to retraining, but are limited in model class~\citep{caoMakingSystemsForget2015, ginartMakingAIForget2019}. \citet{bourtouleMachineUnlearning2020} propose SISA, a flexible approach to exact unlearning that ``shards" a dataset, dividing it and training an ensemble of models where each can be retrained separately. More recent approaches propose approximate removal, requiring the unlearned model to be ``close" to the output of retraining. Some approximate removal methods focus on improving efficiency~\citep{wuDeltaGradRapidRetraining2020} and others try to preserve performance~\citep{wuPUMAPerformanceUnchanged2022}. While these methods apply to a large class of models, they have no formal guarantees on data removal. A second group of approximate approaches provide theoretical guarantees on the statistical indistinguishability of unlearned and retrained models. These noise-based methods leverage convex loss functions to guarantee unlearning with gradient updates~\citep{neelDescenttoDeleteGradientBasedMethods2020a} and Hessian methods~\citep{guoCertifiedDataRemoval2020, sekhariRememberWhatYou, izzoApproximateDataDeletion}. We augment this second set of approximate methods to simultaneously provide strong guarantees on data protection and preserve fairness performance while targeting a common class of models.

\paragraph{Fairness.}

There are a multitude of definitions for fairness in machine learning, such as individual fairness, multicalibration or multiaccuracy, and group fairness. Individual fairness~\citep{dwork2012fairness} posits that ``similar individuals should be treated similarly" by a model. 
On the other hand, recent work has focused on multicalibration and multiaccuracy~\citep{hebert2018multicalibration, kearns2018preventing, deng2023happymap}, where predictions are required to be calibrated across subpopulations. These subpopulation definitions can be highly expressive, containing many intersectional identities from protected groups. In this work, however, we focus on the most commonly studied form of fairness, group fairness, which seeks to balance certain statistical metrics across predefined subgroups. Group fairness literature has proposed various definitions of fairness, but the three most common definitions are Demographic Parity~\citep{zafar2017fairness, feldman2015certifying, zliobaite2015relation, calders2009building}, Equalized Odds, and Equality of Opportunity~\citep{hardt2016equality}. To achieve these definitions, there are generally three approaches to achieving group fairness: \emph{preprocessing} which attempts to correct dataset imbalance to ensure fairness~\citep{calmon2017optimized}, \emph{in-processing} which occurs during training by modifying traditional empirical risk minimization objectives to include fairness constraints~\citep{lowyStochasticOptimizationFramework2022, berk2017convex, agarwal2018reductions, martinez2020minimax}, and \emph{postprocessing} which modifies predictions to ensure fair treatment~\citep{alghamdi2022beyond, hardt2016equality}. 
In this work we focus on in-processing algorithms because they simply modify an objective to account for fairness rather than requiring an additional operation before or after each unlearning request which would also have to be made unlearnable.

\paragraph{Intersections.} Despite advancements in machine unlearning, the literature still lacks sufficient consideration of the downstream impacts of unlearning methods. While recent papers have explored the compatibility of the right to be forgotten with the right to explanation~\citep{krishna2023towards}, there is little work at the intersection of unlearning and fairness. In privacy literature, a thread of work has shown the incompatibility of group fairness with privacy~\citep{esipova2022disparate, bagdasaryan2019differential, cummings2019compatibility} but these incompatibilities arise due to privacy-specific methods, such as gradient clipping and differences in neighboring datasets. Fairness literature has studied the related problem of the influence of training data on fairness~\citep{wang2022understanding}, but does not provide any methods for unlearning. In unlearning literature, recent empirical studies have shown that unlearning can increase disparity~\citep{zhang2023forgotten}, other works have demonstrated the incompatibility of fairness and unlearning for the SISA algorithm \citep{kochno}, and one work~\citep{wang2023inductive} has provided a method to achieve removal and fairness but uses a sharding and retraining algorithm over fairness-corrected graph data for GNNs. In this paper, we propose the first efficient method which achieves fairness while being provably unlearnable without requiring retraining.

\section{PRELIMINARIES} \label{III}

Let $D = \{(x_1, y_1, s_1), \dots, (x_n, y_n, s_n)\} \in \mathcal{D}$ be a training dataset with size $n$, where $\mathcal{D}$ is the set of all possible datasets. For $i=1,2,\ldots, n$, $x_i \in \mathcal{X} \subseteq \mathbb{R}^d$ is a feature vector; $y_i \in \mathcal{Y} = \{0, 1\}$ is a binary label; and $s_i \in \mathcal{S} = \{a, b\}$ is a binary sensitive attribute. A learning algorithm is represented as a mapping $A: \mathcal{D} \rightarrow \mathcal{H}_\Theta$ that maps a dataset $D\in \mathcal{D}$ to a model $A(D)$ in a hypothesis space $\mathcal{H}$ parametrized by some $\theta \in \Theta$. 

\paragraph{Unlearning.} Given set of samples to be deleted $R \subseteq D$, the goal of unlearning is to efficiently find a model that resembles the model retrained on $D\setminus R$ from scratch. We define  an unlearning mechanism as a mapping $M: \mathcal{H}_\Theta \times \mathcal{D} \times \mathcal{D} \rightarrow \mathcal{H}_\Theta$ that maps a learned model $A(D)$, the original dataset $D$, and a subset $R\subseteq D$ of data points to be deleted, to a new unlearned model $M(A(D), D, R)$. Ideally, the unlearned model $M(A(D), D, R)$ is \emph{statistically indistinguishable} from $A(D\setminus R)$~\citep{guoCertifiedDataRemoval2020,neelDescenttoDeleteGradientBasedMethods2020a}. When both the learning algorithm $A$ and the unlearning mechanism $M$ are randomized, i.e., their outputs produce a probability distribution over the hypothesis $\mathcal{H}_\Theta$, the notion of statistical indistinguishability can be formalized as follows.

\begin{definition}[$(\epsilon, \delta)$-Statistical Indistinguishability~\citep{guoCertifiedDataRemoval2020, neelDescenttoDeleteGradientBasedMethods2020a}.]\label{unlearning_definition}

For given $\epsilon, \delta > 0$, an unlearning mechanism $M$ achieves \emph{$(\epsilon, \delta)$-Statistical Indistinguishability} with respect to $A$, if for all $ \mathcal{M} \subseteq \mathcal{H}_\Theta, D\in \mathcal{D}, R \subseteq D$ and $|R| = 1$, we have
\begin{align*}
    \Pr(M(A(D), D, R) \in \mathcal{M}) \leq e^\epsilon \Pr(A(D \setminus R) \in \mathcal{M}) + \delta, \\
    \Pr(A(D \setminus R) \in \mathcal{M}) \leq e^\epsilon \Pr(M(A(D), D, R) \in \mathcal{M}) + \delta.
\end{align*}
\end{definition}
Note that this definition is based on deletion of one data point ($|R| = 1$). We include results for unlearning over multiple requests ($|R| = m$) in the Appendix.

\paragraph{Fairness.} We focus on a popular notion of group fairness, \emph{Equalized Odds}~\citep{hardt2016equality}, and present results for other popular metrics such as \emph{Demographic Parity}, \emph{Equality of Opportunity}, and \emph{Subgroup Accuracy} in the Appendix. Consider a data point $(X, Y, S)$ randomly drawn from the data distribution, where $X\in \mathcal{X}, Y\in \mathcal{Y}$, and $S \in \mathcal{S}$. Note that $X$, $Y$, and $S$ are random variables. Equalized Odds is then defined as follows.
\begin{definition}[Equalized Odds~\citep{hardt2016equality}.] \label{equalizedodds}
A model $h_\theta: \mathcal{X} \rightarrow \mathcal{Y}$ satisfies \emph{Equalized Odds} with respect to the sensitive attribute $S$ and the outcome $Y$ if the model prediction $h_\theta(X)$ and $S$ are independent conditional on $Y$. More formally, for all $y \in \{0, 1\},$
    \begin{align*}
        \Pr(h_\theta(X) = 1 \mid S = a, Y = y) = \\
        \Pr(h_\theta(X) = 1 \mid S = b, Y = y).
    \end{align*}
\end{definition} 

In practice, when exact Equalized Odds is not achieved, we can use \emph{Absolute Equalized Odds Difference}, the absolute difference for both false positive rates and true positive rates between the two subgroups, as a quantitative measure for unfairness. The Absolute Equalized Odds Difference (AEOD) is defined as:
\begin{align}
    \text{AEOD}(\theta) := \frac{1}{2}&\sum_{y\in \{0, 1\}}\big|\Pr(h_\theta(X) = 1 \mid S = a, Y=y) - \nonumber \\
    &\Pr(h_\theta(X) = 1 \mid S = b, Y=y)\big|. \label{eq:aeod}
\end{align} 

We build on prior approaches by incorporating fairness goals into the classic unlearning problem. In the next section, we propose the first method to satisfy fairness and unlearning simultaneously.

\section{FAIR UNLEARNING}

We provide an algorithm that can efficiently unlearn requested data points from a model trained with fair loss. We also prove theoretical guarantees on the unlearnability of our method, and show novel fairness bounds that can be applied to our method as well as prior work.

\subsection{A Convex Fair Loss Function}
To simultaneously achieve fairness and provable unlearning, we leverage a convex fair learning loss introduced by \citet{berk2017convex}. This fair loss can effectively optimize for several popular notions of group fairness, including Equalized Odds. In addition, the convexity of this loss offers mathematical convenience in the development of our provable unlearning algorithm. 

We begin with a binary cross-entropy (BCE) loss for binary classification:
\begin{align}
    \mathcal{L}_\textrm{BCE}(\theta, D) = \frac{1}{n}\sum_{i = 1}^{n} \ell(\theta,x_i, y_i) + \frac{\lambda}{2}||\theta||_2^2, \label{bceloss}
\end{align}
where $\ell(\theta,x_i, y_i) = -y_i\log(\langle x_i,\theta\rangle) - (1-y_i) \log(1 - \langle x_i,\theta\rangle)$ is the logistic loss.

We then add a fairness regularizer. Let $G_a := \{i : s_i = a\}$, $G_b := \{i : s_i = b\}$ be sets of indices indicating subgroup membership for each sample in dataset $D$ with $n_a = |G_a|$ and $n_b = |G_b|$. The fairness regularizer is defined as the following:
\begin{align}
    &\mathcal{L}_\textrm{fair}(\theta, D) := \nonumber \\
    &\qquad\left(\frac{1}{n_an_b} \sum_{i \in G_a}\sum_{j \in G_b} \mathbbm{1}[y_i = y_j](\langle x_i,\theta\rangle - \langle x_j, \theta\rangle) \right)^2. \label{eqn:pairwisefair}
\end{align}
In simple terms, $\mathcal{L}_\textrm{fair}$ penalizes the pairwise difference in logits for samples with the same label between subgroups. Because $\mathcal{L}_\textrm{fair}$ considers samples that share a label regardless of label value, this term directly optimizes for Equalized Odds (Def.~\ref{equalizedodds}). The full fair loss becomes
\begin{align}
    \mathcal{L}(\theta, D) = \mathcal{L}_\textrm{BCE}(\theta, D) + \gamma\mathcal{L}_\textrm{fair}(\theta, D), \label{fairloss}
\end{align}
where $\gamma$ is the fairness regularization parameter.

Most existing unlearning methods~\citep{guoCertifiedDataRemoval2020,izzoApproximateDataDeletion,neelDescenttoDeleteGradientBasedMethods2020a} are specifically designed to unlearn over objective functions like Eq.~(\ref{bceloss}), where the objective constitutes a sum of independent functions on each data point. However, the introduction of the fairness regularizer $\mathcal{L}_\textrm{fair}(\theta, D)$ (and most other fairness regularizers~\citep{beutel2019fairness}) entangles the influence of data points, rendering existing unlearning methods inapplicable. In the following section, we propose a method to address these issues and the first fair unlearning algorithm.

\subsection{The Fair Unlearning Algorithm}
Recall the goal of an unlearning algorithm $M$ is to efficiently update the model $A(D)$, such that $M(A(D), D, R)$ is close to $A(D\setminus R)$. With a concrete definition of our fair loss, we can represent the fully trained model $A(D)$ by its parameters $\theta_D^* := \arg\min_{\theta} \mathcal{L}(\theta, D)$. Similarly, the retrained model $A(D\setminus R)$ corresponds to parameters $\theta_{D'}^* := \arg\min_{\theta} \mathcal{L}(\theta, D')$, where we define $D' := D\setminus R$ for simplicity.

Similar to several existing unlearning methods~\citep{guoCertifiedDataRemoval2020,neelDescenttoDeleteGradientBasedMethods2020a}, our fair unlearning method consists of two steps. First we conduct an efficient update on $\theta_D^*$ to obtain updated model parameters $\theta_{D'}^-$. The goal of this part is to make the updated model parameters $\theta_{D'}^-$ close to the retrained model parameters $\theta_{D'}^*$. Then we add noise to the loss function such that both $A(D)$ and $A(D')$ are randomized, from which we can establish a statistical indistinguishability result between $M(\theta_D^*, D, R)$ and $A(D')$.

\paragraph{Efficient Model Update.}
We first introduce the efficient model update. To simplify notations, we can unroll the squared double sum in Eq.~(\ref{eqn:pairwisefair}) into a single sum using a Cartesian product. Define $N := G_a \times G_b \times G_a \times G_b$. Then we can write $\mathcal{L}_\textrm{fair}$ as
\begin{align}
    \mathcal{L}_\textrm{fair}(\theta, D) = \frac{1}{|N|} \sum_{(i, j, k, l) \in N} \ell_\textrm{fair}(\theta, \{(x_f, y_f)\}_{f \in \{i,j,k,l\}}), \nonumber\\[-15pt]
\end{align}
where
\begin{align}
    &\ell_\textrm{fair}(\theta, \{(x_f, y_f)\}_{f \in \{i,j,k,l\}}) := \nonumber\\
    &\mathbbm{1}[y_i = y_j](\langle x_i, \theta\rangle - \langle x_j, \theta\rangle)\mathbbm{1}[y_k = y_l](\langle x_k, \theta\rangle - \langle x_l, \theta\rangle). 
\end{align}

Without loss of generality, assume that we want to remove one single data point, the final sample $(x_n, y_n, s_n)$, and that this sample comes from subgroup $a$, i.e., $n \in G_a$. We propose an algorithm that updates $\theta_D^*$ to approximately minimize $\mathcal{L}(\theta, D')$ over the remaining dataset $D'= D\setminus R$ where $R = \{(x_n, y_n, s_n)\}$. Our algorithm takes a second-order Newton update from $\theta_D^*$, which results in $\theta_{D'}^-$. Let $d'_a := \{i \in G_a : x_i \in D'\}$, $C_{D'} := d_a' \times G_b \times d_a' \times G_b$. We define a quantity $\Delta$ related to the residual difference between gradients over $D$ and $D'$ as follows:
\begin{align}
    &\Delta := \ell'(\thetafull, x_n, y_n) + \lambda \thetafull \nonumber \\
    &\quad +\gamma\frac{n}{|N|} \sum_{(i,j,k,l) \in N}\ell_\textrm{fair}'(\thetafull, \{(x_f, y_f)\}_{f \in \{i,j,k,l\}})  \nonumber\\
    &\quad -\gamma\frac{n-1}{|C_{D'}|}\sum_{(i,j,k,l)\in C_{D'}}\ell_\textrm{fair}'(\thetafull, \{(x_f, y_f)\}_{f \in \{i,j,k,l\}}). \label{eq:delta}
\end{align}
Intuitively, the first and second term correspond to taking a gradient step in the direction of the BCE loss and $\ell_2$ penalty respectively, while the third term counteracts the entangled influence the sample has on the remaining data and the fourth term takes a gradient step over the fairness penalty. By constructing $\Delta$ in this way, our unlearning algorithm becomes the following: 
\begin{align}
    \theta_{D'}^- = \theta_D^* + H_{\theta_D^*}^{-1}\Delta, \label{fairnewton}
\end{align}
where $H_{\theta_D^*}=\nabla^2 \mathcal{L}(\theta_D^*, D')$ is the Hessian of the full loss. In Theorem \ref{thm:eps_delta}, we achieve a bound on the gradient $||\nabla \mathcal{L}(\thetaunlearn; D')||_2$ and then apply this bound to perform $(\epsilon, \delta)$-unlearning.

\paragraph{Noisy Loss Perturbation.}

To achieve $(\epsilon, \delta)$-statistical indistinguishability, we add Gaussian noise to both the original training and the retraining process. Specifically, we modify the loss function by an additional term $\textbf{b}^T\theta$, where $\textbf{b} \sim N(0, \sigma I)^d$, resulting in the perturbed loss function $\mathcal{L}^\textbf{b}(\theta, D) = \mathcal{L}(\theta, D) + \textbf{b}^T\theta$. 

\paragraph{Runtime Complexity.}

Hessian inversion takes $O(d^2n)$ time, but this can be precomputed and cached as in~\citep{izzoApproximateDataDeletion, guoCertifiedDataRemoval2020}. We can also rearrange terms in Eq.~\eqref{eq:delta} to precompute the double sum over all elements in $N$ which is $O(n^2)$ and then subtract terms accordingly from $N\setminus C$ which is $O(kn+k^2)$, where $k$ is the size of our unlearning request and $k << n$. Thus, at unlearning time, our runtime is of order $O(kn+k^2)$. 

\paragraph{Trade-off between fairness and unlearning.} Although not directly related to each other, both fairness (in the form of increased regularization, $\gamma$) and unlearning (in the form of increased noise, $\sigma$), form a Pareto frontier with accuracy. Thus, for a given accuracy budget, fairness and unlearning come at a cost to one another.

\subsection{Theoretical Guarantees}

\paragraph{Theoretical Guarantee on Statistical Indistinguishability.}
When both the original training and retraining are randomized, the following Theorem~\ref{thm:eps_delta} provides a guarantee that our unlearning method in Eq.~(\ref{fairnewton}) is $(\epsilon, \delta)$-indistinguishable from retraining for some $\epsilon, \delta > 0$.

\begin{theorem}[$(\epsilon, \delta)$-unlearning.\footnote{For proof and generalization to unlearning $m$ samples, see Appendix \ref{sec:thm1_proof}}] \label{thm:eps_delta}
    Let $\thetafull$ be the output of algorithm $A$ trained on $\mathcal{L}^\textbf{b}$. Assume the loss $\ell$ is $\psi$-Lipschitz in its second derivative ($\ell''$ is $\psi$-Lipschitz), and bounded in its first derivative by a constant $||\ell'(\theta, x, y)||_2 \leq g$. Assume the data is bounded such that for all $i \in n$, $||x_i||_2 \leq 1$. Then, 
    \begin{align}
        &||\nabla \mathcal{L}(\thetaunlearn; D')||_2 \leq  \nonumber \\
        &\qquad \frac{\psi}{\lambda^2(n-1)}\left(2g + ||\thetafull||_2\left|\left|\frac{8(n-1)}{n_a^2} \right|\right|_2\right)^2 ,
    \end{align}
    and if $\textbf{b} \sim N(0, k\epsilon'/\epsilon)^d$ with k $>$ 0 and $||\nabla \mathcal{L}(\thetaunlearn; D')||_2 \leq \epsilon'$, then the unlearning algorithm defined by Eq.~\ref{fairnewton} is $(\epsilon, \delta)$-unlearnable with $\delta = 1.5\exp(-k^2/2)$.
\end{theorem}
For logistic loss, we know $||\ell'(\theta, x, y)||_2 \leq 1$ and that $\ell''$ is $1/4$-Lipschitz.

Intuitively, our goal is to achieve Definition~\ref{unlearning_definition} by making the likelihood of a model being output by training approximately equal to the likelihood of that same model being output by unlearning. We do this by 1) ensuring the gradient of our unlearned model is bounded, which for strongly convex loss directly means our unlearned parameter $\thetaunlearn$ is close to the optimal retrained parameter $\thetaretrain$, and 2) adding a sufficient amount of noise to the training objective such that any parameter close to $\thetaretrain$ could have been generated by $A$.

\paragraph{Theoretical Guarantee on Fairness.}

Finally, we show that the model output by our method, $\thetaunlearn$, has a bounded change in Absolute Equalized Odds Difference (Eq.~\eqref{eq:aeod}) in comparison to the model output by retraining, $\thetaretrain$. We formalize that $\text{AEOD}(\thetaunlearn)$ is bounded in the following theorem.

\begin{theorem}[AEOD is bounded.] \label{thm:fairnesG_bound}
    Assume both $\ell_\textrm{BCE}(\theta, x, y) = \ell(\theta, x, y) + \frac{2\lambda}{n}||\theta||_2^2$ and $\ell_\textrm{fair}$ are $L$-Lipschitz. Suppose $\forall x \in \mathcal{X} \subseteq \mathbb{R}^d, ||x||_2 \leq 1$, and for all $\mathcal{Z} \subseteq \mathcal{X}$, the conditional probability $P(\mathcal{Z} | S, Y) \leq c\frac{|\mathcal{Z}|}{|\mathcal{X}|}$ for some $c\ge 1$, which controls the concentration of the probability distribution. Then, with the results from Thm.~\ref{thm:eps_delta} we can show that $||\thetaretrain-\thetaunlearn||_2 \leq \kappa$ for some $\kappa$ being $O(1/n^2)$. Furthermore, we have that the absolute equalized odds difference for $\thetaunlearn$ is bounded by
    \begin{align}
        &\text{AEOD}(\thetaunlearn) \leq \text{AEOD}(\thetaretrain) \nonumber \\
        & \qquad + c\left(1- \frac{2\int_{\mu}^{1}\frac{\pi^{\frac{d-1}{2}}}{\Gamma(\frac{d+1}{2})}(1-y^2)^{\frac{d-1}{2}} dy}{V}\right), \label{eq:aeodbound}
    \end{align}
    where $V$ is the volume of the unit $d$-sphere, and $\mu = O(\sqrt{\kappa})$.
\end{theorem}

As $n$ goes to infinity, $\kappa$ approaches 0, the integral in Eq.~(\ref{eq:aeodbound}) evaluates to $V/2$, so the second term vanishes.

Theorem~\ref{thm:fairnesG_bound} states that the unlearned and retrained models will be similar to each other in terms of fairness performance. By developing an unlearnable method that optimizes for fairness during training and retraining, our theory guarantees that unlearning will also be as fair. Furthermore, Thm.~\ref{thm:fairnesG_bound} only requires $||\thetaretrain-\thetaunlearn||_2 \leq \kappa$ in addition to standard assumptions in unlearning literature for linear models such as Lipschitzness and bounded data. For any other method that achieves the above bound on parameter distance, Thm.~\ref{thm:fairnesG_bound} bounds the change in fairness of the unlearned and retrained model. Proofs for both theorems can be found in the Appendix.

\section{EXPERIMENTAL RESULTS}

In this section, we evaluate the effectiveness of fair unlearning in terms of both fairness and accuracy in the scenario where data points are deleted at random. Next, we consider more extreme settings for unlearning where data points are deleted disproportionately from a specific subgroup.

\subsection{Experimental Setup}

We evaluate the proposed fair unlearning method on three real-world datasets using a logistic regression model. However, note that previous work has shown combining an unlearnable linear layer with a DP-SGD trained neural network preserves unlearning guarantees~\citep{guoCertifiedDataRemoval2020}. We simulate three types of data deletion requests and compare various unlearning methods with retraining. We also report the metrics of the original model trained on the full training set as a reference. For all results we repeat our experiments five times and report mean and standard deviation. Note that the effectiveness of erasing the influence of requested data points is guaranteed by our theory rather than evaluated empirically, as is common in prior unlearning literature~\citep{guoCertifiedDataRemoval2020,neelDescenttoDeleteGradientBasedMethods2020a}.

\paragraph{Datasets.} We consider three datasets commonly used in the machine learning fairness literature. 1) \emph{COMPAS}~\citep{angwin2016machine} is a classic fairness benchmark with the task of predicting a criminal defendant's likelihood of recidivism. 2) \emph{Adult} from the UCI Machine Learning Repository~\citep{asuncion2007uci} is a large scale dataset with the task of predicting income bracket. 3) High School Longitudinal Study (\emph{HSLS}) is a dataset containing survey responses from high school students and parents with the task of predicting academic year~\citep{ingels2011high,jeong2022fairness}. For all datasets, we use the individual's race as the sensitive group identity.

\paragraph{Data Deletion Settings.} We simulate data deletions in three settings: 1) \emph{random deletion}, where data deletion requests come randomly from the training set independent of subgroup membership; 2) \emph{minority deletion}, where data deletion requests come randomly from the minority group of the training set; 3) \emph{majority deletion}, where data deletion requests come randomly from the majority group of the training set. 
\paragraph{Evaluation Metrics.} We report experimental results on \emph{Absolute Equalized Odds Difference} (AEOD) as defined in Eq.~(\ref{eq:aeod}), with results for Demographic Parity, Equality of Opportunity, and Statistical Parity in the Appendix. In addition, we explore the tradeoffs between $\epsilon$ and $\delta$ which control different strengths of unlearning. Finally, we report runtime, test accuracy, and fairness-accuracy tradeoffs for $\gamma$ in the Appendix.

\begin{figure*}[h!]
    \centering
    \begin{subfigure}{0.3\linewidth}
         \centering         
         \includegraphics[width=\linewidth]{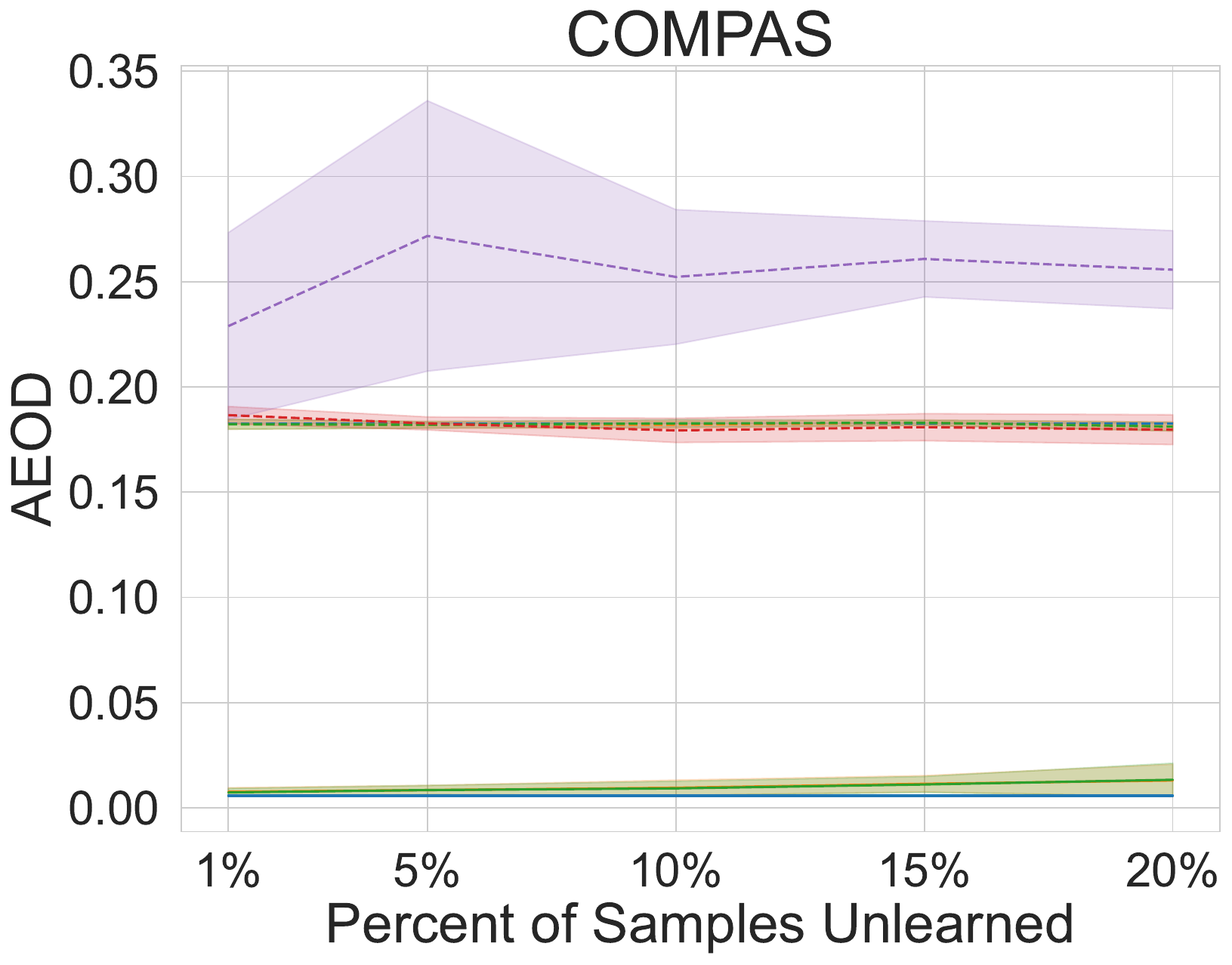}
     \end{subfigure}
     \begin{subfigure}{0.3\linewidth}
         \centering
         \includegraphics[width=\linewidth]{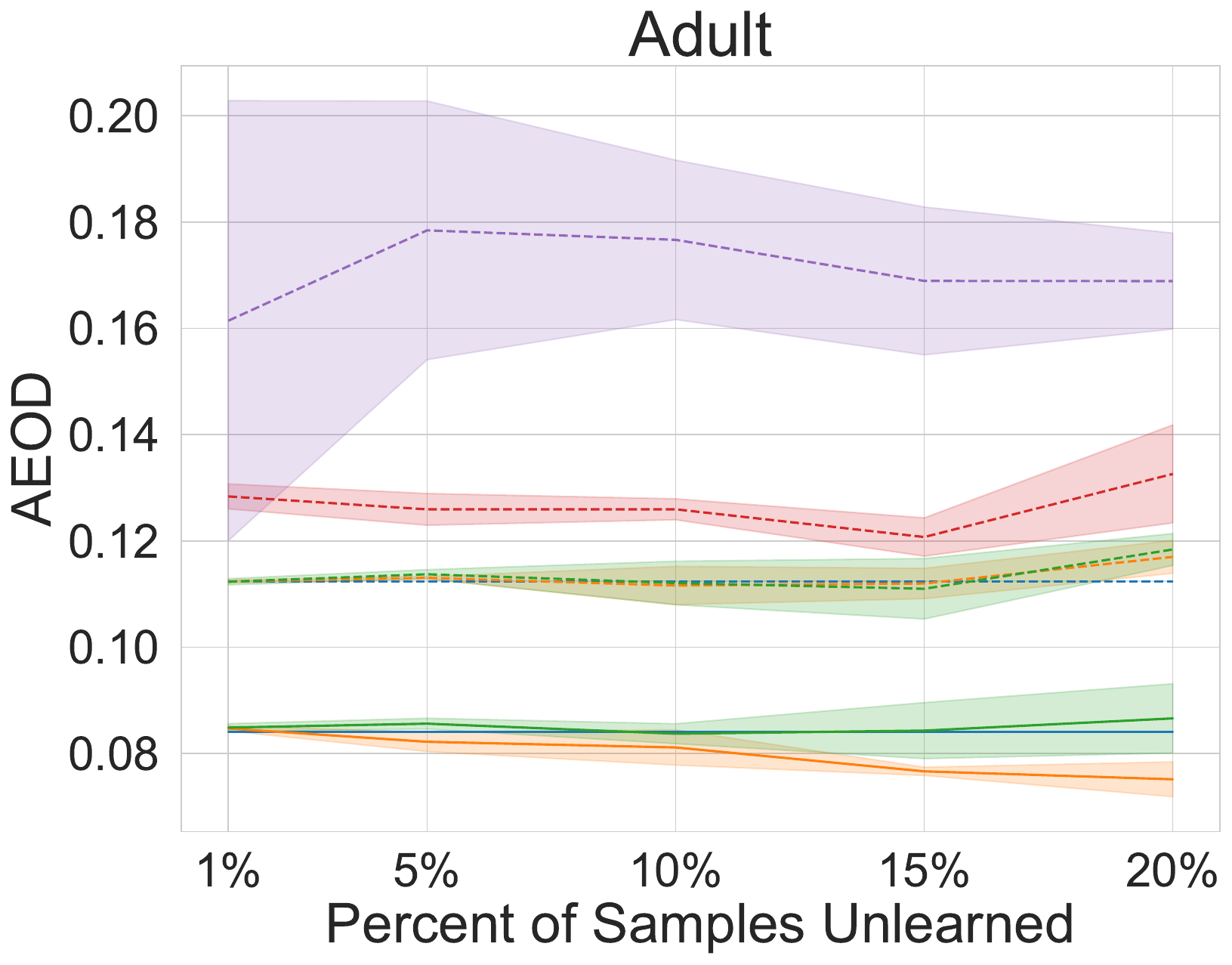}
     \end{subfigure}
     \begin{subfigure}{0.3\linewidth}
         \centering
         \includegraphics[width=\linewidth]{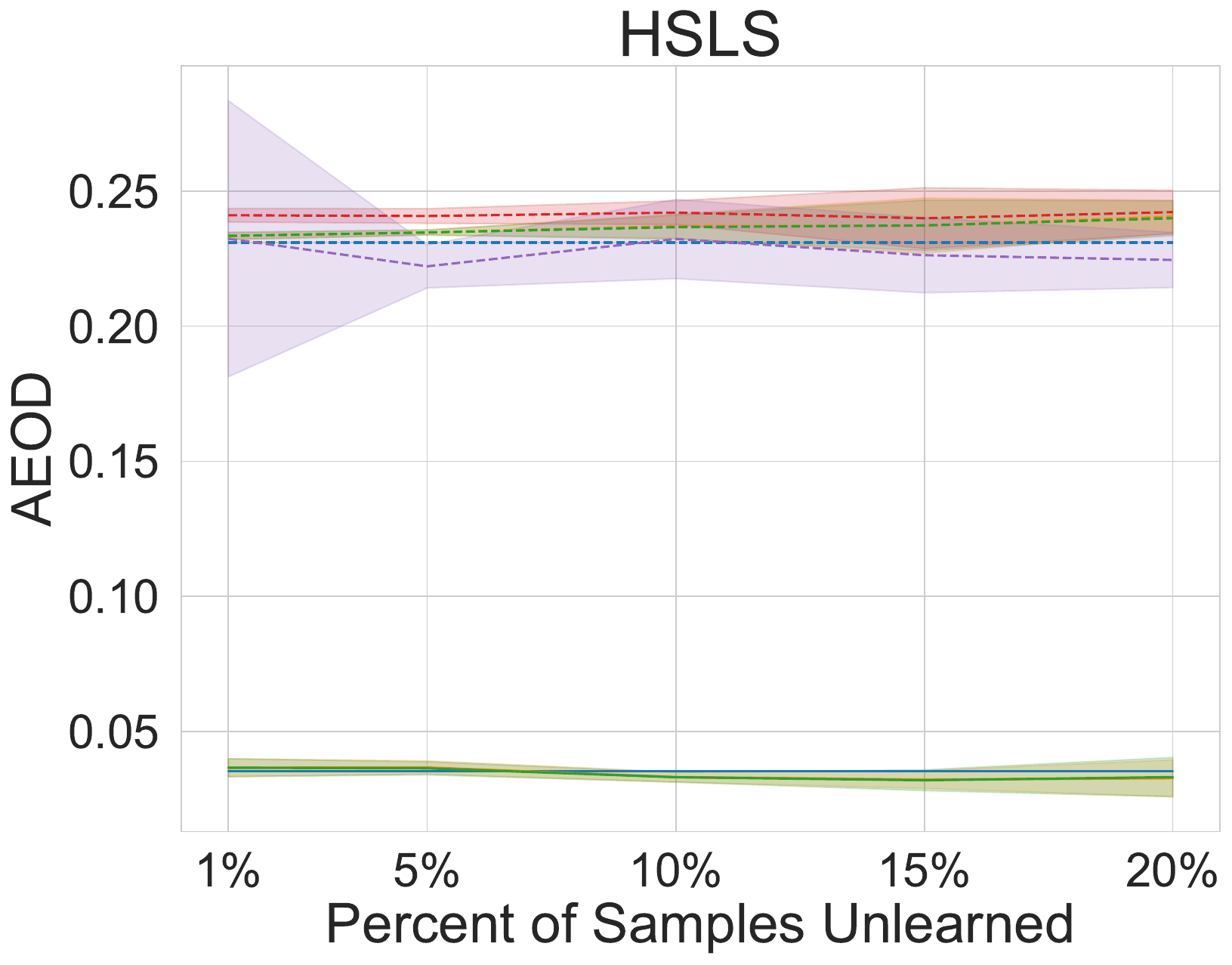}
     \end{subfigure}
     \\
     \includegraphics[width=0.7\linewidth]{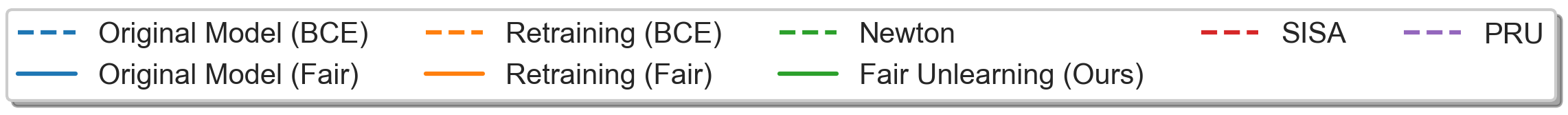}
        \caption{Absolute equalized odds difference (lower is better) for unlearning methods over random requests on COMPAS, Adult, and HSLS. Our method well-approximates fair retraining.}
        \label{fig:unlearning at random}
\end{figure*}
\vspace{-10pt}
\begin{figure*}[h!]
    \centering
    \begin{subfigure}{0.3\linewidth}
         \centering
         \includegraphics[width=\linewidth]{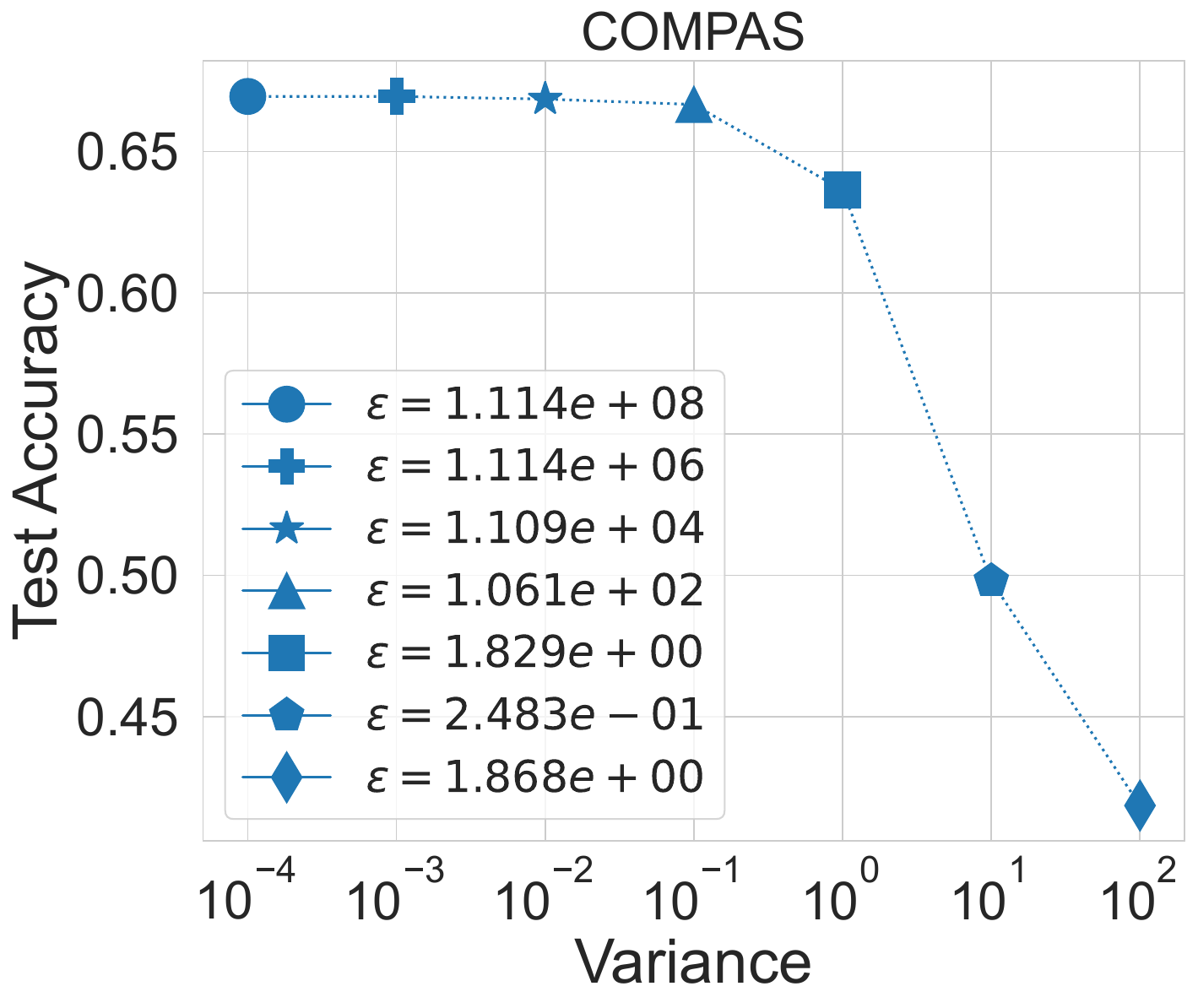}
     \end{subfigure}
     \begin{subfigure}{0.3\linewidth}
         \centering
         \includegraphics[width=\linewidth]{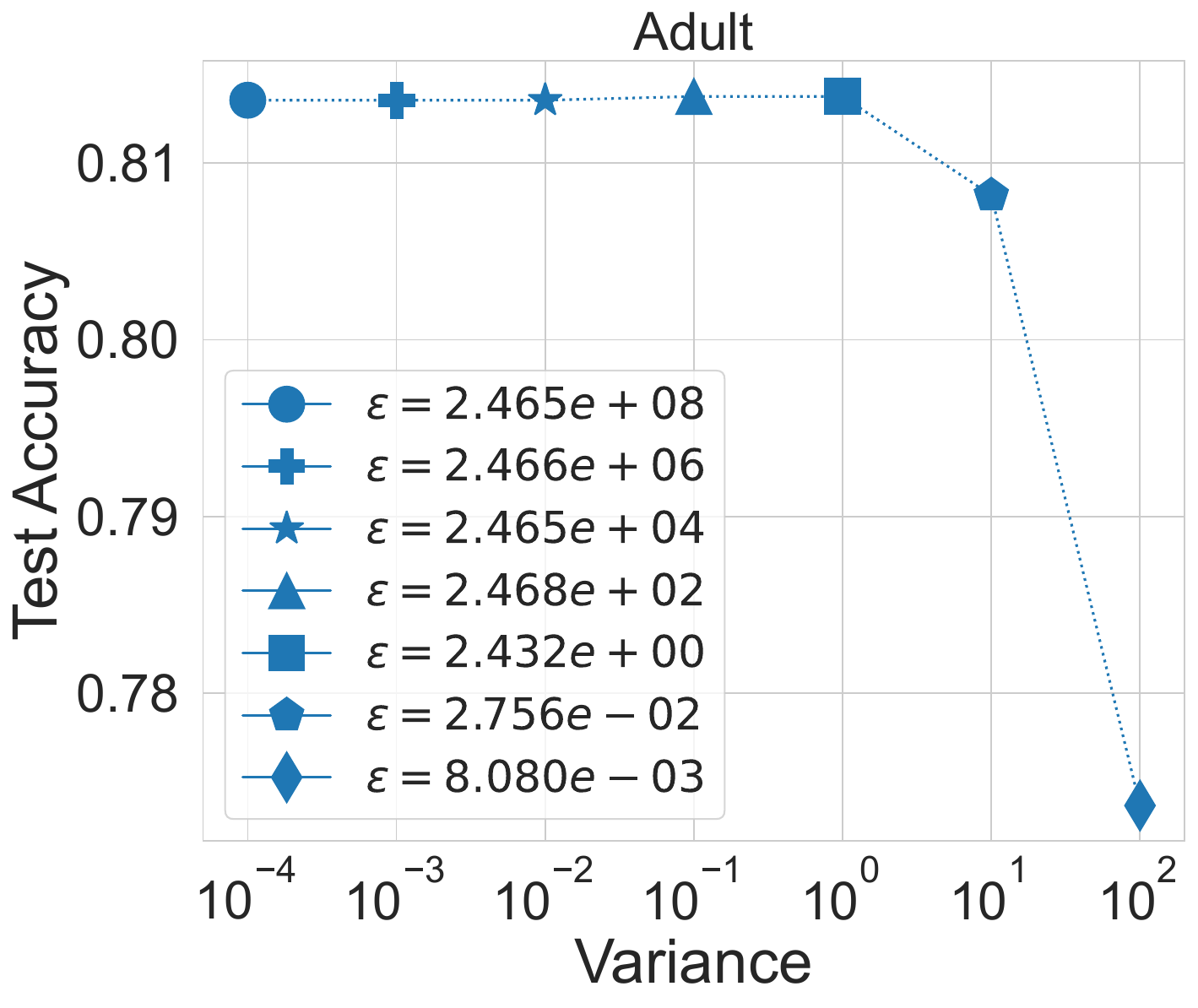}
     \end{subfigure}
     \begin{subfigure}{0.3\linewidth}
         \centering
         \includegraphics[width=\linewidth]{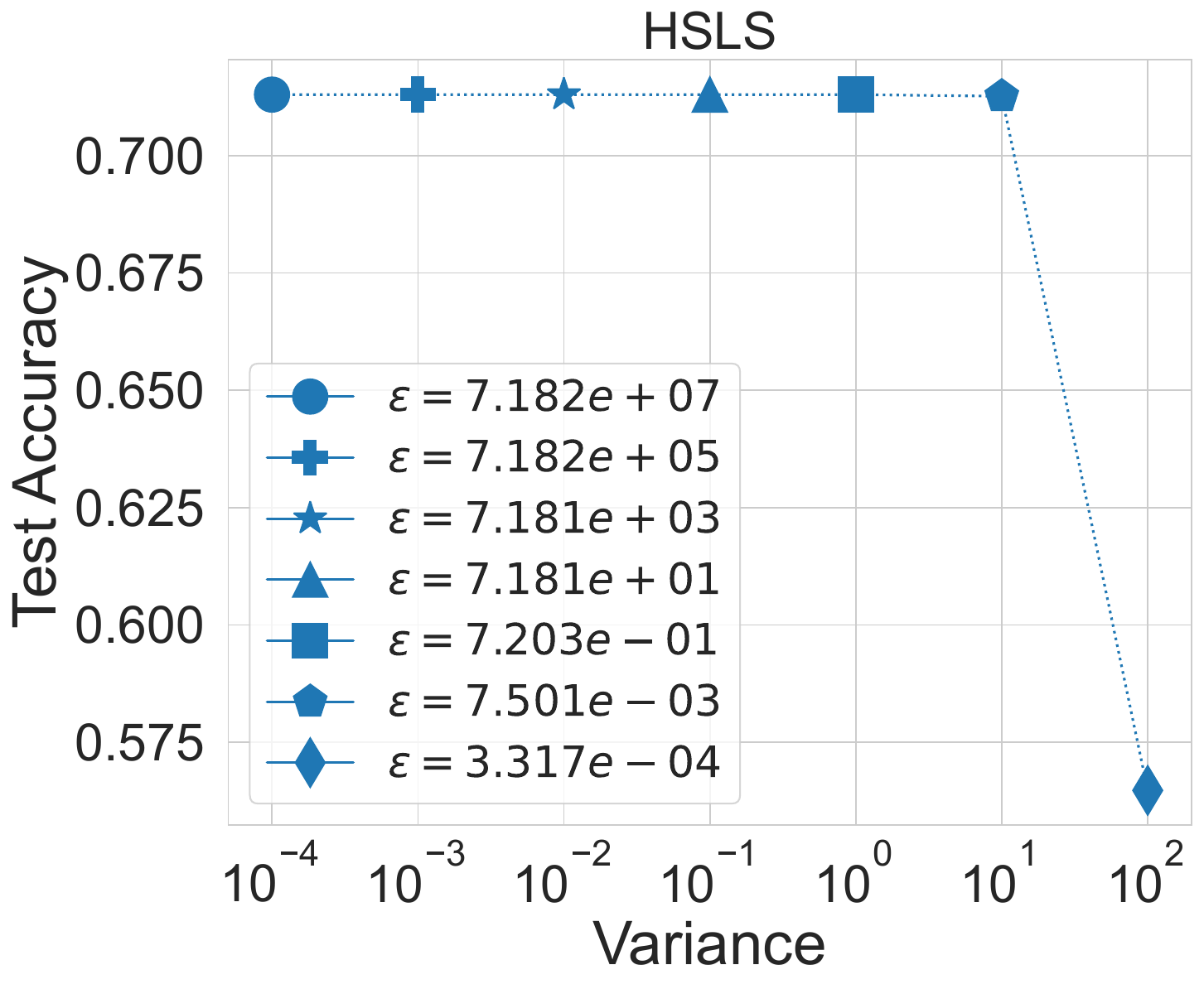}
     \end{subfigure}
     \\
        \caption{Unlearning performance and leakage for various levels of noise added while unlearning 100 samples according to Thm.~\ref{thm:eps_delta}. We set $\delta = 0.0001$ and report corresponding $\epsilon$ values in the legend.}
        \label{fig:unlearning leakage}
\vspace{-10pt}
\end{figure*}

\paragraph{Baseline Methods.} 
We compare our method with four baselines designed for unlearning with BCE loss (Eq.~(\ref{bceloss})). In addition to brute-force retraining with BCE loss (named as \textbf{Retraining (BCE)}), we consider three unlearning methods from existing literature. The three methods are 1) \textbf{Newton}~\citep{guoCertifiedDataRemoval2020} a second-order approach that takes a single Newton step to approximate retraining; 2) \textbf{PRU}~\citep{izzoApproximateDataDeletion} which uses a statistical technique called the residual method to efficiently approximate the Hessian used in the Newton step; and 3) \textbf{SISA}~\citep{bourtouleMachineUnlearning2020} a method that splits the dataset into shards and trains a model on each shard, while unlearning by retraining each model separately to speed up the process. In addition, we compare our \textbf{Fair Unlearning} method designed for unlearning with fair loss (Eq.~(\ref{fairloss})) to brute-force retraining with fair loss (named as \textbf{Retraining (Fair)}). Finally, as references, we also evaluate models trained on the full training set with BCE loss and fair loss, respectively named as \textbf{Full Training (BCE)} and \textbf{Full Training (Fair)}. Note that the goal of all machine unlearning methods is to approximate retraining.

\paragraph{Runtime Complexity.} We compare the runtimes of our unlearning method and retraining over various removal sizes and report results in the Appendix. We also report precomputation times for Hessian inversion. Overall we see a $2\times$ to $20\times$ speedup with our method over retraining.

\paragraph{Unlearning Leakage.} We report the unlearning leakage of our method for various levels of noise variance using the bound shown in Thm.~\ref{thm:eps_delta}. For small datasets, \citet{guoCertifiedDataRemoval2020} propose a tighter data dependent bound which we adapt to fair unlearning in Appendix \ref{sec:datadependentbound}. We pick the lower value of $\epsilon$ between the two bounds for our reported results. For various magnitudes of $\sigma$, we fix $\delta = 0.0001$ and compute the accuracy of the model unlearned with our method in addition to the leakage, $\epsilon$. Values are reported in Fig.~\ref{fig:unlearning leakage}. We can achieve realistic values of $\epsilon$ with reasonable amounts of noise while preserving performance.

\subsection{Experimental Results on Fairness}
\begin{figure*}[h!]
    \centering
    \begin{subfigure}{0.3\linewidth}
         \centering
         \includegraphics[width=\linewidth]{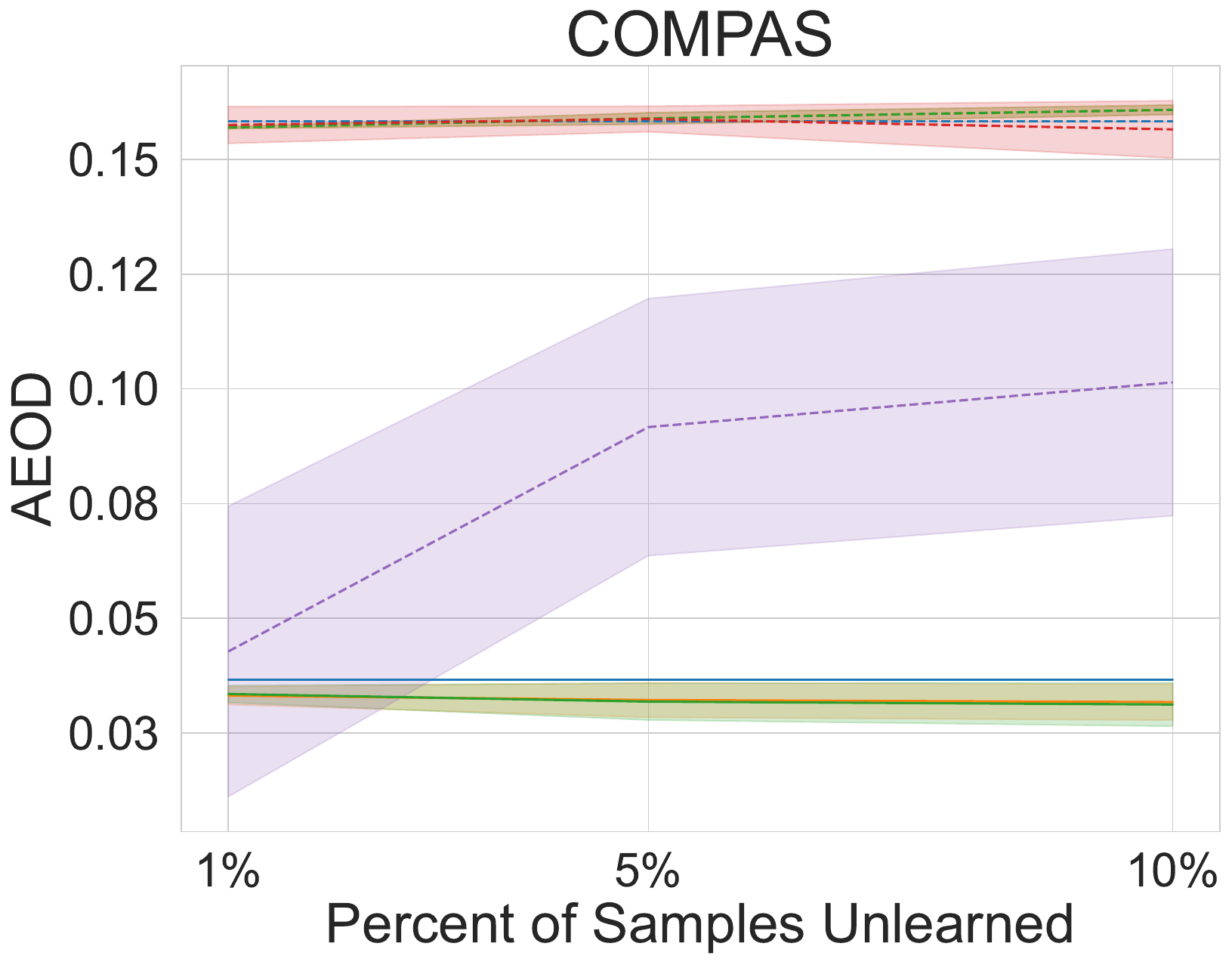}
     \end{subfigure}
     \begin{subfigure}{0.3\linewidth}
         \centering
         \includegraphics[width=\linewidth]{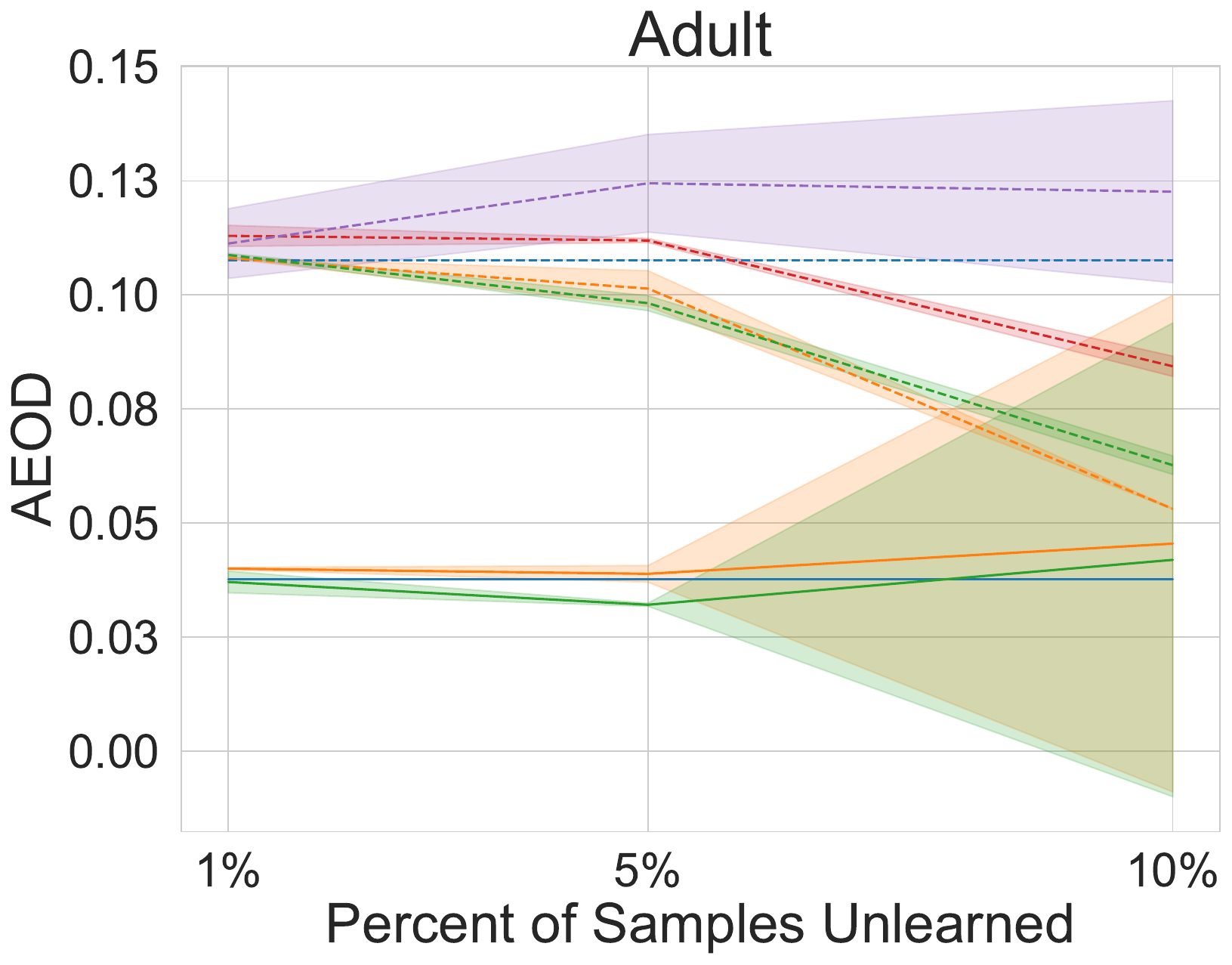}
     \end{subfigure}
     \begin{subfigure}{0.3\linewidth}
         \centering
         \includegraphics[width=\linewidth]{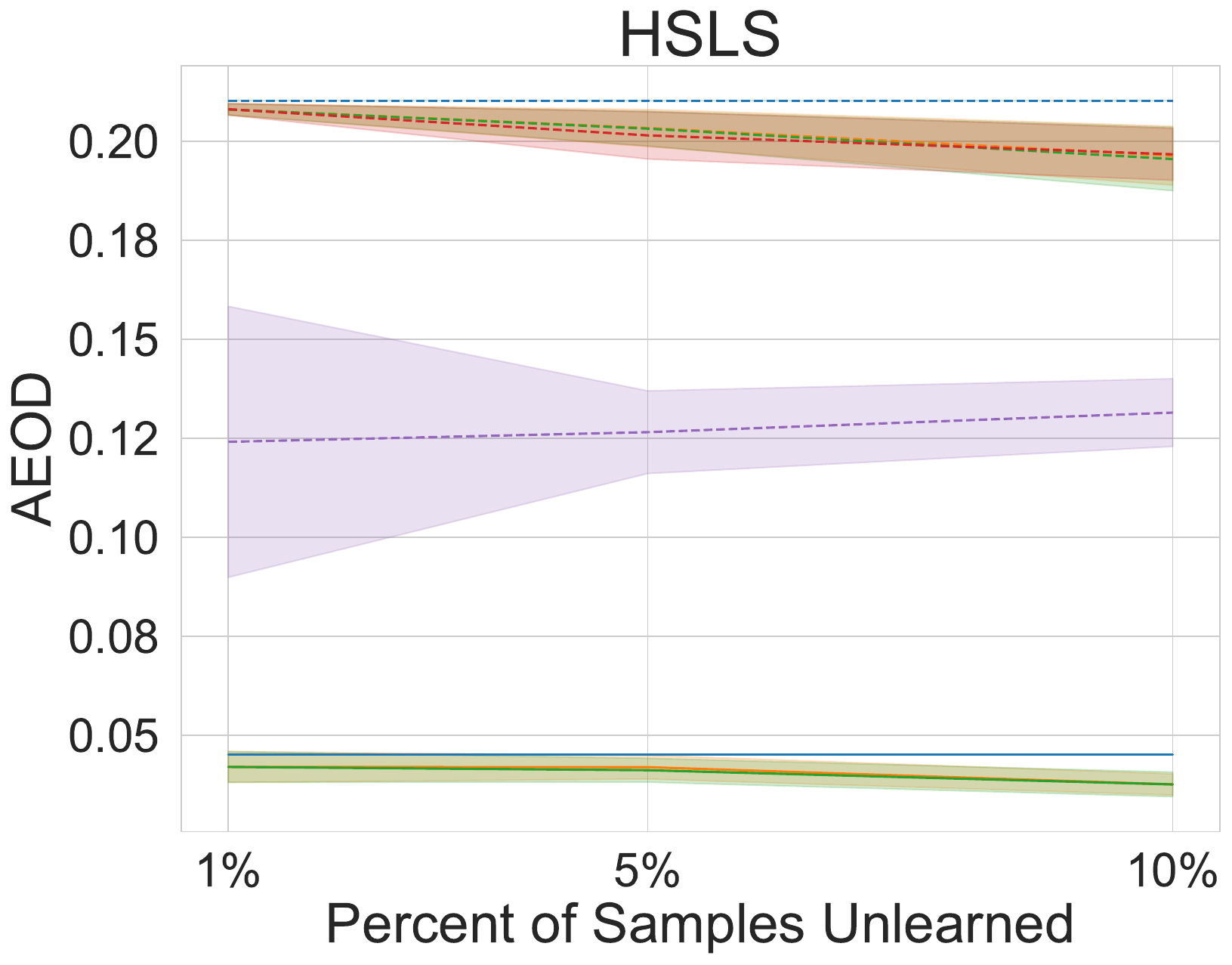}
     \end{subfigure}
     \\ \vspace{-5pt}
     \begin{subfigure}{0.3\linewidth}
         \centering
         \includegraphics[width=\linewidth]{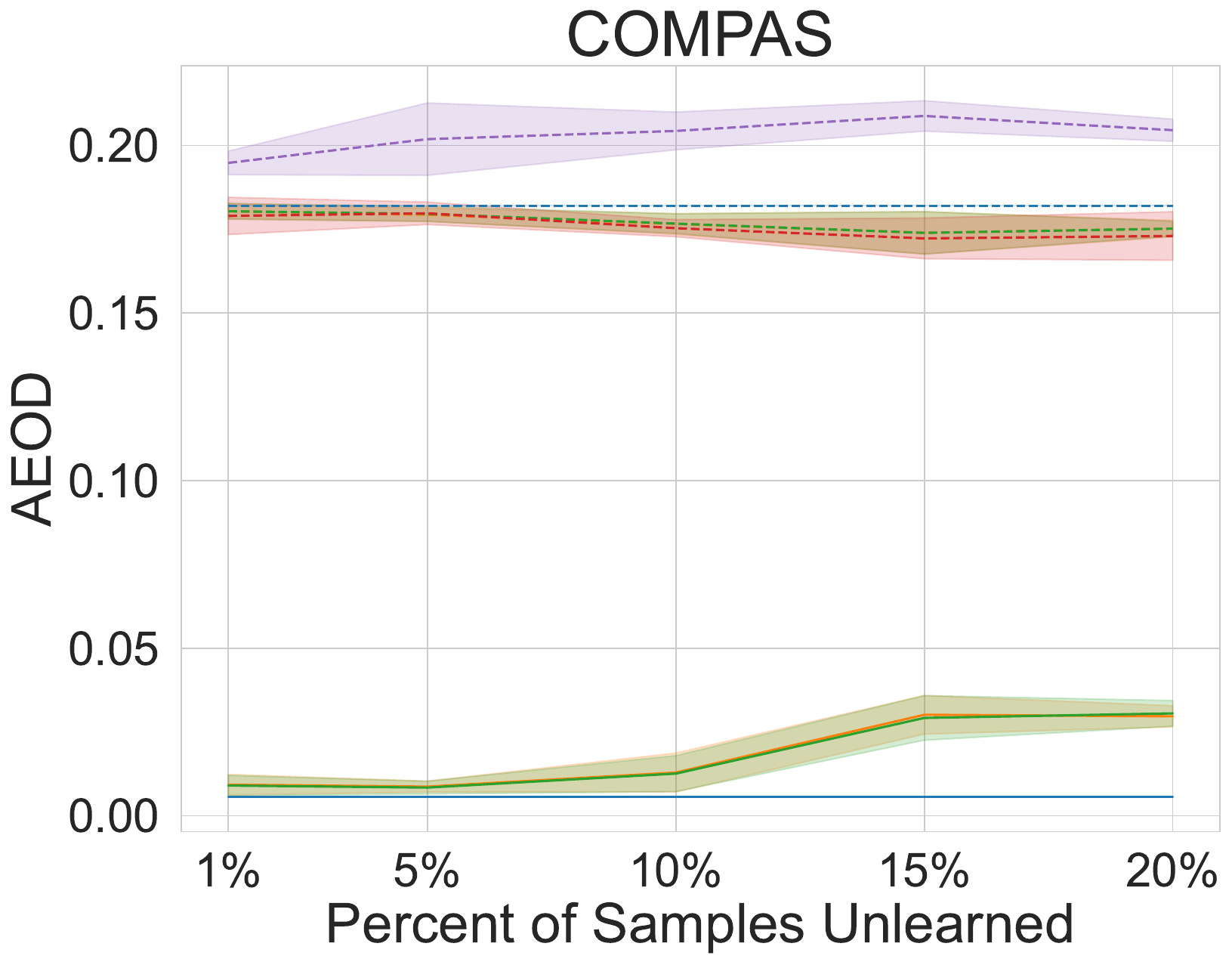}
     \end{subfigure}
     \begin{subfigure}{0.3\linewidth}
         \centering
         \includegraphics[width=\linewidth]{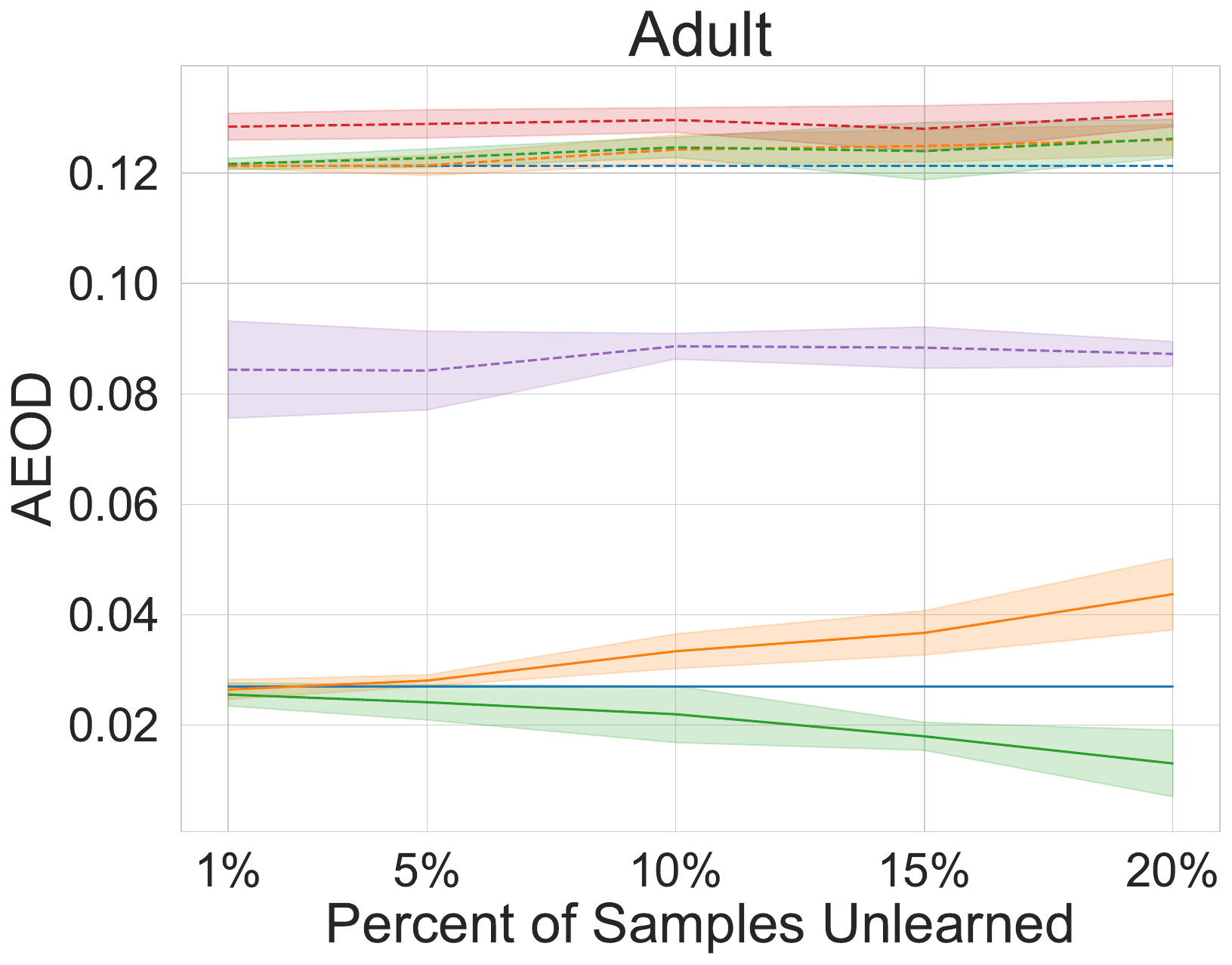}
     \end{subfigure}
     \begin{subfigure}{0.3\linewidth}
         \centering
         \includegraphics[width=\linewidth]{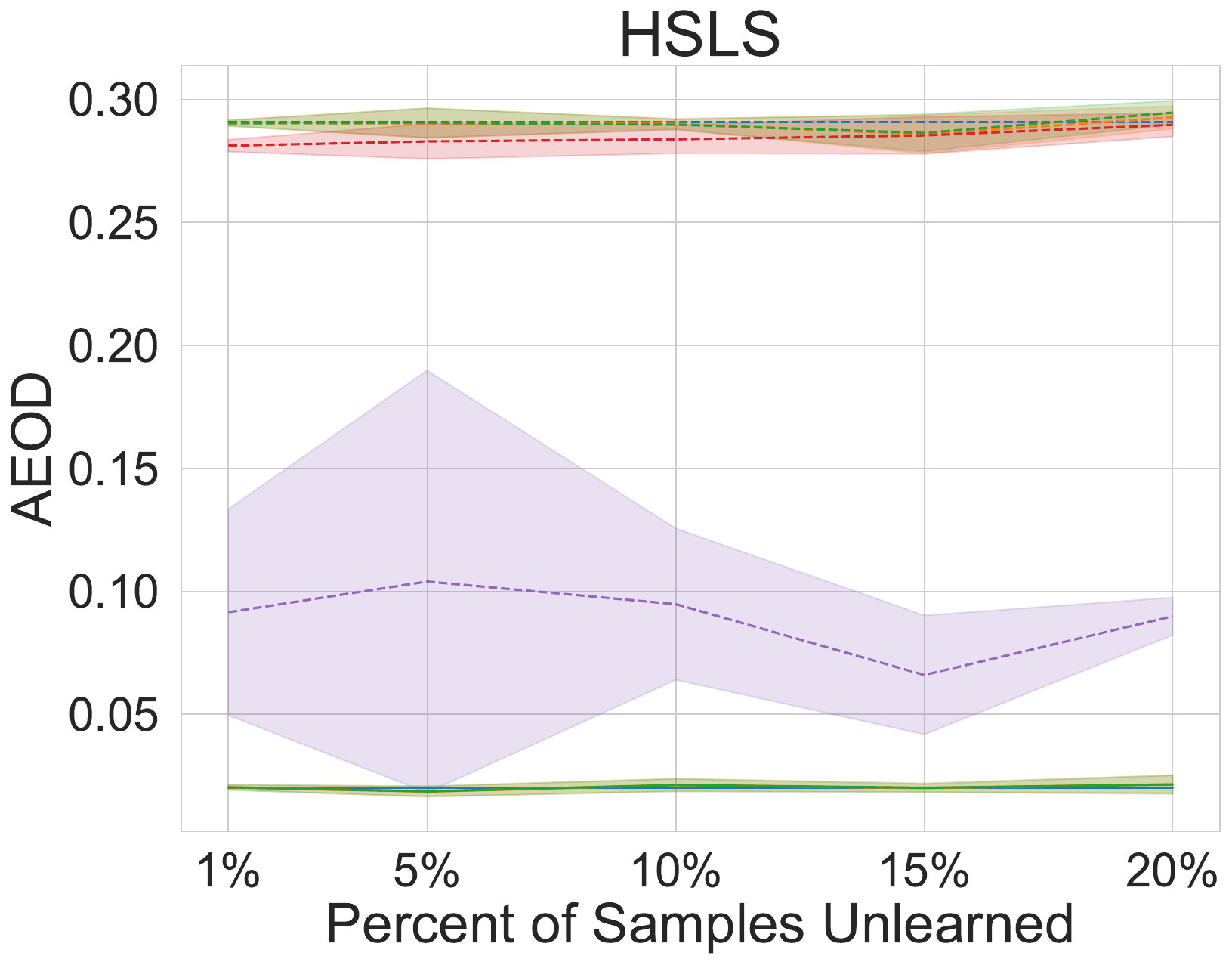}
     \end{subfigure}\\
     \includegraphics[width=0.7\linewidth]{figures/legend.png}
        \caption{Absolute equalized odds difference (lower is better) for unlearning methods when unlearning from the minority (top) and majority (bottom) subgroup on COMPAS, Adult, and HSLS.}
        \label{fig:unlearning from minority/majority subgroups}
\vspace{-10pt}
\end{figure*}
\paragraph{Deletion at Random.}
We evaluate our method in terms of fairness when unlearning at random from our training dataset. For each of our datasets, and for both BCE and fair loss, we train a base logistic regression model on the full training dataset. Then, we randomly select $1\%$, $5\%$, $10\%$, $15\%$, and $20\%$ of our training set to unlearn. We compare the AOED of our method to various baselines. Results are reported in Figure~\ref{fig:unlearning at random}. Note that methods computed with BCE loss are shown with dashed lines whereas methods computed with fair loss are shown with solid lines.

As predicted by our theoretical results, our \textbf{Fair Unlearning} method (green) well-approximates the fairness performance of retraining with fair loss. In addition, we see that across all levels of unlearning, our method outperforms all baselines by a significant margin in AEOD for COMPAS, with similar results for Adult and HSLS. In the full statement of Thm. \ref{thm:eps_delta} for unlearning $m$ samples (see Appendix \ref{sec:thm1_proof}), we show that the unlearning bound scales with $m$. We similarly observe that the precision of our method decreases as we unlearn more samples. Note that \textbf{Newton} is the best approximator for \textbf{Retraining (BCE)}, oftentimes exactly covering the retraining curve, and \textbf{Fair Unlearning} similarly replicates the performance of \textbf{Retraining (Fair)}. Further, we see that certain existing methods such as \textbf{PRU} and \textbf{SISA} are highly variable and can lead to greater disparity than retraining. Notably, across all datasets, our method outperforms baselines in preserving fairness.

\paragraph{Deletion from Subgroups.}

If unlearning at random preserves relative subgroup distribution in data, the worst case unlearning distribution for fairness is when unlearning requests are concentrated on specific subgroups. Subgroup deletion can also create a negative feedback loop where data deletion results in degradation in performance for that subgroup, motivating more members of said subgroup to request unlearning. To explore this, we unlearn only from minority groups and majority groups (Fig. \ref{fig:unlearning from minority/majority subgroups}). Note that when unlearning from minority groups, we only unlearn up to $10\%$ of the dataset as the minority subgroup has (by definition) less representation in the dataset. Too many deletions from the minority group would destroy the stability of the learned models. For example, in the Adult dataset, which has 90\% majority and 10\% minority data, 10\% deletion from the minority group is the maximum possible and already drastically reduces model performance.

We see that when unlearning from both minority and majority subgroups, our \textbf{Fair Unlearning} method approximates \textbf{Retraining (Fair)} and thus maintains the best fairness performance throughout unlearning. When unlearning from minority subgroups, for both COMPAS and HSLS, our method perfectly approximates fair retraining and largely outperforms other baselines in terms of AEOD. In Adult, note that vanilla unlearning methods improve in fairness despite unlearning from the minority subgroup. This is not unexpected as models can generalize better on subsets of a dataset, especially if the removed data are outliers, and the same can occur with fairness performance. However, we highlight that our method always achieves better AEOD than baselines. This indicates that even if removing a portion of data happens to help fairness, we can better improve fairness performance by explicitly accommodating fairness constraints. We also see that both fair retraining and unlearning have high standard deviation at $10\%$ unlearning due to its imbalanced subgroup distribution. We do not see a similar change for baseline methods trained with BCE loss because those methods do not consider subgroups in their loss, whereas in the fair setting this error can be attributed primarily to the fair loss term which requires well-represented subgroups for accurate estimation. This behavior highlights a problem in unlearning settings that even when unlearning a small proportion of the total dataset, unlearning requests may completely destroy accurate estimates of subgroup distributions, resulting in downstream disparity when not considering fairness. 

In majority subgroups, we see consistent performance by our method across all datasets, with significant improvements in terms of AEOD. Finally, similar to the results for random unlearning, \textbf{PRU} display high error bands and atypical AEOD scores when compared to other baselines, while \textbf{Newton} remains accurate at approximating \textbf{Retraining (BCE)}.

\subsection{Experimental Results on Accuracy}
Finally, we validate the accuracy of \textbf{Fair Unlearning} to ensure that increased fairness does not come at a cost to performance. Trivially, AEOD can be optimized by a random guessing classifier despite its poor performance. We report test accuracy at $5\%$ and $20\%$ of samples unlearned across our various baselines and datasets in the supplemental material. We see that while achieving significant improvements in fairness, our method and brute-force retraining maintain performance in terms of accuracy and can even improve in some cases when unlearning at random and from subgroups. 

\section{CONCLUSION}

In this work we show that existing methods fail to accommodate models with fairness considerations, and we resolve this issue by proposing the first fair unlearning algorithm. We prove our method is unlearnable and bound its fairness guarantees. Then, we evaluate our method on three real-world datasets and show that our method well approximates retraining with a fair loss function in terms of both accuracy and fairness. Further, we highlight worst case scenarios where unlearning requests disproportionately come from certain subgroups. Our fair unlearning method allows practitioners to achieve multiple data protection goals at the same time (fairness and unlearning) and satisfy regulatory principles. Furthermore, we hope this work sparks discussion surrounding other auxiliary goals that may need be compliant with unlearning such as robustness and interpretability, or other definitions of fairness such as multicalibration and multiaccuracy.

\subsubsection*{Acknowledgements}

This material is based upon work supported by the National Science Foundation Graduate Research Fellowship under Grant No. DGE-2140743. This work is also supported in part by the NSF awards IIS-2008461, IIS-2040989, IIS-2238714, FAI-2040880, and research awards from Google, JP Morgan, Amazon, Adobe, Harvard Data Science Initiative, and the Digital, Data, and Design (D$^3$) Institute at Harvard. The views expressed here are those of the authors and do not reflect the official policy or position of the funding agencies.

\bibliographystyle{unsrtnat}
\bibliography{main.bib}

\begin{thebibliography}{43}
\providecommand{\natexlab}[1]{#1}
\providecommand{\url}[1]{\texttt{#1}}
\expandafter\ifx\csname urlstyle\endcsname\relax
  \providecommand{\doi}[1]{doi: #1}\else
  \providecommand{\doi}{doi: \begingroup \urlstyle{rm}\Url}\fi

\bibitem[{European Commission}()]{GDPR}
{European Commission}.
\newblock Regulation (eu) 2016/679 of the european parliament and of the council of 27 april 2016 on the protection of natural persons with regard to the processing of personal data and on the free movement of such data, and repealing directive 95/46/ec (general data protection regulation).
\newblock URL \url{https://eur-lex.europa.eu/eli/reg/2016/679}.

\bibitem[CCPA(2018)]{CCPA}
CCPA.
\newblock California consumer privacy act (ccpa), 2018.
\newblock URL \url{https://oag.ca.gov/privacy/ccpa}.

\bibitem[Guo et~al.(2020)Guo, Goldstein, Hannun, and van~der Maaten]{guoCertifiedDataRemoval2020}
Chuan Guo, Tom Goldstein, Awni Hannun, and Laurens van~der Maaten.
\newblock Certified {Data} {Removal} from {Machine} {Learning} {Models}, August 2020.
\newblock URL \url{http://arxiv.org/abs/1911.03030}.
\newblock arXiv:1911.03030 [cs, stat].

\bibitem[Bourtoule et~al.(2020)Bourtoule, Chandrasekaran, Choquette-Choo, Jia, Travers, Zhang, Lie, and Papernot]{bourtouleMachineUnlearning2020}
Lucas Bourtoule, Varun Chandrasekaran, Christopher~A. Choquette-Choo, Hengrui Jia, Adelin Travers, Baiwu Zhang, David Lie, and Nicolas Papernot.
\newblock Machine {Unlearning}, December 2020.
\newblock URL \url{http://arxiv.org/abs/1912.03817}.
\newblock arXiv:1912.03817 [cs].

\bibitem[Izzo et~al.(2021)Izzo, Smart, Chaudhuri, and Zou]{izzoApproximateDataDeletion}
Zachary Izzo, Mary~Anne Smart, Kamalika Chaudhuri, and James Zou.
\newblock Approximate {Data} {Deletion} from {Machine} {Learning} {Models}.
\newblock page~10, 2021.

\bibitem[Neel et~al.(2020)Neel, Roth, and {Sharifi-Malvajerdi}]{neelDescenttoDeleteGradientBasedMethods2020a}
Seth Neel, Aaron Roth, and Saeed {Sharifi-Malvajerdi}.
\newblock Descent-to-{{Delete}}: {{Gradient-Based Methods}} for {{Machine Unlearning}}, July 2020.

\bibitem[Beutel et~al.(2019)Beutel, Chen, Doshi, Qian, Wei, Wu, Heldt, Zhao, Hong, Chi, et~al.]{beutel2019fairness}
Alex Beutel, Jilin Chen, Tulsee Doshi, Hai Qian, Li~Wei, Yi~Wu, Lukasz Heldt, Zhe Zhao, Lichan Hong, Ed~H Chi, et~al.
\newblock Fairness in recommendation ranking through pairwise comparisons.
\newblock In \emph{Proceedings of the 25th ACM SIGKDD international conference on knowledge discovery \& data mining}, pages 2212--2220, 2019.

\bibitem[Jiang et~al.(2019)Jiang, Chiappa, Lattimore, Gy{\"o}rgy, and Kohli]{jiang2019degenerate}
Ray Jiang, Silvia Chiappa, Tor Lattimore, Andr{\'a}s Gy{\"o}rgy, and Pushmeet Kohli.
\newblock Degenerate feedback loops in recommender systems.
\newblock In \emph{Proceedings of the 2019 AAAI/ACM Conference on AI, Ethics, and Society}, pages 383--390, 2019.

\bibitem[Corbett-Davies and Goel(2018)]{corbett2018measure}
Sam Corbett-Davies and Sharad Goel.
\newblock The measure and mismeasure of fairness: A critical review of fair machine learning.
\newblock \emph{arXiv preprint arXiv:1808.00023}, 2018.

\bibitem[Chen et~al.(2021)Chen, Chen, Lipkova, Wang, Williamson, Lu, Sahai, and Mahmood]{chen2021algorithm}
Richard~J Chen, Tiffany~Y Chen, Jana Lipkova, Judy~J Wang, Drew~FK Williamson, Ming~Y Lu, Sharifa Sahai, and Faisal Mahmood.
\newblock Algorithm fairness in ai for medicine and healthcare.
\newblock \emph{arXiv preprint arXiv:2110.00603}, 2021.

\bibitem[OSTP(2022)]{aibillofrights}
OSTP.
\newblock Blueprint for an {AI} {B}ill of {R}ights.
\newblock Technical report, The White House, Washington, DC, October 2022.

\bibitem[Lowy et~al.(2022)Lowy, Baharlouei, Pavan, Razaviyayn, and Beirami]{lowyStochasticOptimizationFramework2022}
Andrew Lowy, Sina Baharlouei, Rakesh Pavan, Meisam Razaviyayn, and Ahmad Beirami.
\newblock A {{Stochastic Optimization Framework}} for {{Fair Risk Minimization}}, September 2022.

\bibitem[Berk et~al.(2017)Berk, Heidari, Jabbari, Joseph, Kearns, Morgenstern, Neel, and Roth]{berk2017convex}
Richard Berk, Hoda Heidari, Shahin Jabbari, Matthew Joseph, Michael Kearns, Jamie Morgenstern, Seth Neel, and Aaron Roth.
\newblock A convex framework for fair regression.
\newblock \emph{arXiv preprint arXiv:1706.02409}, 2017.

\bibitem[Hardt et~al.(2016)Hardt, Price, and Srebro]{hardt2016equality}
Moritz Hardt, Eric Price, and Nati Srebro.
\newblock Equality of opportunity in supervised learning.
\newblock \emph{Advances in neural information processing systems}, 29, 2016.

\bibitem[Cao and Yang(2015)]{caoMakingSystemsForget2015}
Yinzhi Cao and Junfeng Yang.
\newblock Towards {{Making Systems Forget}} with {{Machine Unlearning}}.
\newblock In \emph{2015 {{IEEE Symposium}} on {{Security}} and {{Privacy}}}, pages 463--480, {San Jose, CA}, May 2015. {IEEE}.
\newblock ISBN 978-1-4673-6949-7.
\newblock \doi{10.1109/SP.2015.35}.

\bibitem[Ginart et~al.(2019)Ginart, Guan, Valiant, and Zou]{ginartMakingAIForget2019}
Antonio Ginart, Melody~Y. Guan, Gregory Valiant, and James Zou.
\newblock Making {{AI Forget You}}: {{Data Deletion}} in {{Machine Learning}}, November 2019.

\bibitem[Wu et~al.(2020)Wu, Dobriban, and Davidson]{wuDeltaGradRapidRetraining2020}
Yinjun Wu, Edgar Dobriban, and Susan~B. Davidson.
\newblock {{DeltaGrad}}: {{Rapid}} retraining of machine learning models, June 2020.

\bibitem[Wu et~al.(2022)Wu, Hashemi, and Srinivasa]{wuPUMAPerformanceUnchanged2022}
Ga~Wu, Masoud Hashemi, and Christopher Srinivasa.
\newblock {{PUMA}}: {{Performance Unchanged Model Augmentation}} for {{Training Data Removal}}.
\newblock \emph{Proceedings of the AAAI Conference on Artificial Intelligence}, 36\penalty0 (8):\penalty0 8675--8682, June 2022.
\newblock ISSN 2374-3468, 2159-5399.
\newblock \doi{10.1609/aaai.v36i8.20846}.

\bibitem[Sekhari et~al.()Sekhari, Acharya, Suresh, and Kamath]{sekhariRememberWhatYou}
Ayush Sekhari, Jayadev Acharya, Ananda~Theertha Suresh, and Gautam Kamath.
\newblock Remember {What} {You} {Want} to {Forget}: {Algorithms} for {Machine} {Unlearning}.
\newblock page~12.

\bibitem[Dwork et~al.(2012)Dwork, Hardt, Pitassi, Reingold, and Zemel]{dwork2012fairness}
Cynthia Dwork, Moritz Hardt, Toniann Pitassi, Omer Reingold, and Richard Zemel.
\newblock Fairness through awareness.
\newblock In \emph{Proceedings of the 3rd innovations in theoretical computer science conference}, pages 214--226, 2012.

\bibitem[H{\'e}bert-Johnson et~al.(2018)H{\'e}bert-Johnson, Kim, Reingold, and Rothblum]{hebert2018multicalibration}
Ursula H{\'e}bert-Johnson, Michael Kim, Omer Reingold, and Guy Rothblum.
\newblock Multicalibration: Calibration for the (computationally-identifiable) masses.
\newblock In \emph{International Conference on Machine Learning}, pages 1939--1948. PMLR, 2018.

\bibitem[Kearns et~al.(2018)Kearns, Neel, Roth, and Wu]{kearns2018preventing}
Michael Kearns, Seth Neel, Aaron Roth, and Zhiwei~Steven Wu.
\newblock Preventing fairness gerrymandering: Auditing and learning for subgroup fairness.
\newblock In \emph{International conference on machine learning}, pages 2564--2572. PMLR, 2018.

\bibitem[Deng et~al.(2023)Deng, Dwork, and Zhang]{deng2023happymap}
Zhun Deng, Cynthia Dwork, and Linjun Zhang.
\newblock Happymap: A generalized multi-calibration method.
\newblock \emph{arXiv preprint arXiv:2303.04379}, 2023.

\bibitem[Zafar et~al.(2017)Zafar, Valera, Gomez~Rodriguez, and Gummadi]{zafar2017fairness}
Muhammad~Bilal Zafar, Isabel Valera, Manuel Gomez~Rodriguez, and Krishna~P Gummadi.
\newblock Fairness beyond disparate treatment \& disparate impact: Learning classification without disparate mistreatment.
\newblock In \emph{Proceedings of the 26th international conference on world wide web}, pages 1171--1180, 2017.

\bibitem[Feldman et~al.(2015)Feldman, Friedler, Moeller, Scheidegger, and Venkatasubramanian]{feldman2015certifying}
Michael Feldman, Sorelle~A Friedler, John Moeller, Carlos Scheidegger, and Suresh Venkatasubramanian.
\newblock Certifying and removing disparate impact.
\newblock In \emph{proceedings of the 21th ACM SIGKDD international conference on knowledge discovery and data mining}, pages 259--268, 2015.

\bibitem[Zliobaite(2015)]{zliobaite2015relation}
Indre Zliobaite.
\newblock On the relation between accuracy and fairness in binary classification.
\newblock \emph{arXiv preprint arXiv:1505.05723}, 2015.

\bibitem[Calders et~al.(2009)Calders, Kamiran, and Pechenizkiy]{calders2009building}
Toon Calders, Faisal Kamiran, and Mykola Pechenizkiy.
\newblock Building classifiers with independency constraints.
\newblock In \emph{2009 IEEE international conference on data mining workshops}, pages 13--18. IEEE, 2009.

\bibitem[Calmon et~al.(2017)Calmon, Wei, Vinzamuri, Natesan~Ramamurthy, and Varshney]{calmon2017optimized}
Flavio Calmon, Dennis Wei, Bhanukiran Vinzamuri, Karthikeyan Natesan~Ramamurthy, and Kush~R Varshney.
\newblock Optimized pre-processing for discrimination prevention.
\newblock \emph{Advances in neural information processing systems}, 30, 2017.

\bibitem[Agarwal et~al.(2018)Agarwal, Beygelzimer, Dud{\'\i}k, Langford, and Wallach]{agarwal2018reductions}
Alekh Agarwal, Alina Beygelzimer, Miroslav Dud{\'\i}k, John Langford, and Hanna Wallach.
\newblock A reductions approach to fair classification.
\newblock In \emph{International Conference on Machine Learning}, pages 60--69. PMLR, 2018.

\bibitem[Martinez et~al.(2020)Martinez, Bertran, and Sapiro]{martinez2020minimax}
Natalia Martinez, Martin Bertran, and Guillermo Sapiro.
\newblock Minimax pareto fairness: A multi objective perspective.
\newblock In \emph{International Conference on Machine Learning}, pages 6755--6764. PMLR, 2020.

\bibitem[Alghamdi et~al.(2022)Alghamdi, Hsu, Jeong, Wang, Michalak, Asoodeh, and Calmon]{alghamdi2022beyond}
Wael Alghamdi, Hsiang Hsu, Haewon Jeong, Hao Wang, Peter Michalak, Shahab Asoodeh, and Flavio Calmon.
\newblock Beyond adult and compas: Fair multi-class prediction via information projection.
\newblock \emph{Advances in Neural Information Processing Systems}, 35:\penalty0 38747--38760, 2022.

\bibitem[Krishna et~al.(2023)Krishna, Ma, and Lakkaraju]{krishna2023towards}
Satyapriya Krishna, Jiaqi Ma, and Himabindu Lakkaraju.
\newblock Towards bridging the gaps between the right to explanation and the right to be forgotten.
\newblock \emph{arXiv preprint arXiv:2302.04288}, 2023.

\bibitem[Esipova et~al.(2022)Esipova, Ghomi, Luo, and Cresswell]{esipova2022disparate}
Maria~S Esipova, Atiyeh~Ashari Ghomi, Yaqiao Luo, and Jesse~C Cresswell.
\newblock Disparate impact in differential privacy from gradient misalignment.
\newblock \emph{arXiv preprint arXiv:2206.07737}, 2022.

\bibitem[Bagdasaryan et~al.(2019)Bagdasaryan, Poursaeed, and Shmatikov]{bagdasaryan2019differential}
Eugene Bagdasaryan, Omid Poursaeed, and Vitaly Shmatikov.
\newblock Differential privacy has disparate impact on model accuracy.
\newblock \emph{Advances in neural information processing systems}, 32, 2019.

\bibitem[Cummings et~al.(2019)Cummings, Gupta, Kimpara, and Morgenstern]{cummings2019compatibility}
Rachel Cummings, Varun Gupta, Dhamma Kimpara, and Jamie Morgenstern.
\newblock On the compatibility of privacy and fairness.
\newblock In \emph{Adjunct Publication of the 27th Conference on User Modeling, Adaptation and Personalization}, pages 309--315, 2019.

\bibitem[Wang et~al.(2022)Wang, Wang, and Liu]{wang2022understanding}
Jialu Wang, Xin~Eric Wang, and Yang Liu.
\newblock Understanding instance-level impact of fairness constraints.
\newblock In \emph{International Conference on Machine Learning}, pages 23114--23130. PMLR, 2022.

\bibitem[Zhang et~al.(2023)Zhang, Pan, Hoang, Xing, Staples, Xu, Yao, Lu, and Zhu]{zhang2023forgotten}
Dawen Zhang, Shidong Pan, Thong Hoang, Zhenchang Xing, Mark Staples, Xiwei Xu, Lina Yao, Qinghua Lu, and Liming Zhu.
\newblock To be forgotten or to be fair: Unveiling fairness implications of machine unlearning methods.
\newblock \emph{arXiv preprint arXiv:2302.03350}, 2023.

\bibitem[Koch and Soll(2023)]{kochno}
Korbinian Koch and Marcus Soll.
\newblock No matter how you slice it: Machine unlearning with sisa comes at the expense of minority classes.
\newblock In \emph{First IEEE Conference on Secure and Trustworthy Machine Learning}, 2023.

\bibitem[Wang et~al.(2023)Wang, Huai, and Wang]{wang2023inductive}
Cheng-Long Wang, Mengdi Huai, and Di~Wang.
\newblock Inductive graph unlearning.
\newblock \emph{arXiv preprint arXiv:2304.03093}, 2023.

\bibitem[Angwin et~al.(2016)Angwin, Larson, Mattu, and Kirchner]{angwin2016machine}
Julia Angwin, Jeff Larson, Surya Mattu, and Lauren Kirchner.
\newblock Machine bias.
\newblock In \emph{Ethics of data and analytics}, pages 254--264. Auerbach Publications, 2016.

\bibitem[Asuncion and Newman(2007)]{asuncion2007uci}
Arthur Asuncion and David Newman.
\newblock Uci machine learning repository, 2007.

\bibitem[Ingels et~al.(2011)Ingels, Pratt, Herget, Burns, Dever, Ottem, Rogers, Jin, and Leinwand]{ingels2011high}
Steven~J Ingels, Daniel~J Pratt, Deborah~R Herget, Laura~J Burns, Jill~A Dever, Randolph Ottem, James~E Rogers, Ying Jin, and Steve Leinwand.
\newblock High school longitudinal study of 2009 (hsls: 09): Base-year data file documentation. nces 2011-328.
\newblock \emph{National Center for Education Statistics}, 2011.

\bibitem[Jeong et~al.(2022)Jeong, Wang, and Calmon]{jeong2022fairness}
Haewon Jeong, Hao Wang, and Flavio~P Calmon.
\newblock Fairness without imputation: A decision tree approach for fair prediction with missing values.
\newblock In \emph{Proceedings of the AAAI Conference on Artificial Intelligence}, volume~36, pages 9558--9566, 2022.

\end{thebibliography}

\newpage
\appendix
\onecolumn
\section{APPENDIX}
\subsection{Proof of Theorem \ref{thm:eps_delta}}\label{sec:thm1_proof}

We begin by introducing two lemmas.

Recall the definition of our full loss function,
\begin{align}
    \mathcal{L}(\theta, D) = \frac{1}{n}\sum_{i=1}^n \ell(\theta, x_i, y_i) + \frac{\lambda}{2n}||\theta||_2^2& + \gamma\frac{1}{|N|}\sum_{i,j,k,l \in N}\ell_\textrm{fair}(\theta, \{(x_f, y_f)\}_{f \in \{i,j,k,l\}}), \nonumber \\
    \ell_\textrm{fair}(\theta, \{(x_f, y_f)\}_{f \in \{i,j,k,l\}}) &= \mathbbm{1}[y_i = y_j](\langle x_i, \theta\rangle  - \langle x_j, \theta\rangle)\mathbbm{1}[y_k = y_l](\langle x_k, \theta\rangle  - \langle x_l, \theta\rangle) \nonumber \\
    &=\mathbbm{1}[y_i = y_j]\mathbbm{1}[y_k = y_l]\left(\theta^T (x_i^T x_k - x_i^T x_l - x_j^T x_k + x_j^T x_l)\theta\right). \nonumber
\end{align}
For our proof of Lemma \ref{lem:bounded_gradient}, we multiply by the entire objective by $n$:
\begin{align}
    \mathcal{L}(\theta, D) = \sum_{i=1}^n \ell(\theta, x_i, y_i) + \frac{\lambda}{2}||\theta||_2^2 + \gamma\frac{n}{|N|}\sum_{i,j,k,l \in N}\ell_\textrm{fair}(\theta, \{(x_f, y_f)\}_{f \in \{i,j,k,l\}}).
\end{align}
Let $\thetaunlearn := \thetafull+H_{\thetafull}^{-1}\Delta$ be the update step of our unlearing algorithm, where $\thetafull$ is our optimum when training on the entire dataset, $H_{\thetafull}^{-1} := (\nabla^2 \mathcal{L}(\thetafull; D'))^{-1}$ is the inverse hessian of the loss at $\thetafull$ evaluated over the remaining dataset defined as $D' = D \setminus R$. 
Let $R$ be our unlearning requests, and without loss of generality assume these are the last $m$ elements of $D$, $\{x_{n-m}, ... , x_n\}$. Let $G_a := \{i : s_i = a\}$, $G_b := \{i : s_i = b\}$ be sets of indices indicating subgroup membership for each sample. Define $r_a := \{i \in G_a : x_i \in R\}$, $r_b := \{i \in G_b : x_i \in R\}$, and $d'_a, d'_b$ similarly for $x_i \in D'$. Recall $n = |D|$, and let $n_a = |G_a|, n_b = |G_b|$ and similarly let $m = |R|, m_a = |r_a|, m_b = |r_b|$. 

Let $N = G_a \times G_b \times G_a \times G_b$, and let $C_{D'} = d'_a \times d'_b \times d'_a \times d'_b.$ Then,

Then, define our unlearning step $\Delta$ as the following:
\begin{align}
    &\Delta := \sum_{i=m}^{n}\ell'(\thetafull, x_i, y_i) + m\lambda \thetafull+\gamma\frac{n}{|N|} \sum_{(i,j,k,l) \in N}\ell_\textrm{fair}'(\thetafull, \{(x_f, y_f)\}_{f \in \{i,j,k,l\}})  \nonumber\\
    &-\gamma\frac{n-m}{|C_{D'}|}\sum_{(i,j,k,l)\in C_{D'}}\ell_\textrm{fair}'(\thetafull, \{(x_f, y_f)\}_{f \in \{i,j,k,l\}}). \label{eqn:delta}
\end{align}
\begin{lemma} \label{lem:bounded_gradient}
Assume the ERM loss $\ell$ is $\psi$-Lipschitz in its second derivative ($\ell''$ is $\psi$-Lipschitz), and bounded in its first derivative by a constant $||\ell'(\theta, x, y)||_2 \leq g$. Suppose the data is bounded such that for all $i \in n$,  $||x_i||_2 \leq 1$. Then the gradient of our unlearned model on the remaining dataset is bounded by:
\begin{align}
    ||\nabla \mathcal{L}(\thetaunlearn; D')||_2 \leq  \frac{\psi}{\lambda^2(n-m)}\left(2mg + ||\thetafull||_2\left|\left|\frac{8(m_a^2n_b^2 + m_b^2n_a^2 - m_a^2m_b^2)(n-m)}{(n_a^2n_b^2)} \right|\right|_2\right)^2.
\end{align}
\end{lemma}
\textit{Proof:} This proof largely follows the structure of the proof of Theorem 1 in~\citet{guoCertifiedDataRemoval2020}, but differs in the addition of our fairness loss term which introduces additional terms into the bound.

Let $G(\theta) = \nabla L(\theta; D')$ be the gradient at the \textit{remaining} dataset $D' = D \setminus R$. We want to bound the norm of the gradient at $\theta = \thetaunlearn$, as we know for convex loss, the optimum when retraining over the \textit{remainder} dataset $D'$ is zero. Keeping the norm of the unlearned weights close to zero ensures we have achieved unlearning.

We substitute the expression for the unlearned parameter and do a first-order Taylor approximation, which states there exists an $\eta \in [0,1]$ such that the following holds:
\begin{align*}
    G(\thetaunlearn) &= G(\thetafull + H_{\thetafull}^{-1}\Delta), \\
    &= G(\thetafull) + \nabla G(\thetafull + \eta H_{\thetafull}^{-1} \Delta)H_{\thetafull}^{-1}\Delta. 
\end{align*}
Recall that G is the gradient of the loss, so here $\nabla G$ is the Hessian evaluated at this $\eta$ perturbed value. Let $H_{\theta_\eta}$ be the Hessian evaluated at the point $\theta_\eta = \thetafull + \eta H_{\thetafull}^{-1}\Delta$. Then we have
\begin{align}
    &= G(\thetafull) + H_{\theta_\eta}H_{\thetafull}^{-1}\Delta, \nonumber \\
    &= G(\thetafull) + \Delta + H_{\theta_\eta}H_{\thetafull}^{-1}\Delta - \Delta.
\end{align}
By our choice of $\Delta$, $G(\thetafull) + \Delta = 0$.
\begin{align}
    G(\thetafull) &= \nabla \mathcal{L}(\thetafull; D'), \nonumber \\
    &= \sum_{i=1}^{n-m} \ell'(\thetafull, x_i, y_i) + (n-m)\lambda\thetafull \nonumber \\
    &+\gamma\frac{n-m}{|C_{D'}|}\sum_{(i,j,k,l)\in C_{D'}}\ell_\textrm{fair}'(\theta, \{(x_f, y_f)\}_{f \in \{i,j,k,l\}}). \label{eqn:G}
\end{align}
Combining \eqref{eqn:delta} and \eqref{eqn:G}, we see $G(\thetafull) + \Delta$ equals 0. 
\begin{align}
    G(\thetafull) + \Delta &= \sum_{i=1}^{n} \ell'(\thetafull, x_i, y_i) + n\lambda\thetafull + \gamma\frac{n}{|N|}\sum_{i,j,k,l \in N}\ell_\textrm{fair}'(\theta, \{(x_f, y_f)\}_{f \in \{i,j,k,l\}}), \label{eqn:G_plus_delta} \\
    &= \nabla \mathcal{L}(\thetafull; D),  \nonumber \\
    &= 0. \nonumber
\end{align}
We see in~\eqref{eqn:G_plus_delta} that $G(\thetafull) + \Delta = \nabla \mathcal{L}(\thetafull; D)$. Then, $\nabla \mathcal{L}(\thetafull; D) = 0$ because our loss function is convex and our optimum occurs at $\thetafull$. 

Returning to the proof of Lemma \ref{lem:bounded_gradient}, we have:
\begin{align}
    G(\thetaunlearn) &= G(\thetafull) + \Delta + H_{\theta_\eta}H_{\thetafull}^{-1}\Delta - \Delta, \nonumber \\
    &= 0 + H_{\theta_\eta}H_{\thetafull}^{-1}\Delta - \Delta, \nonumber \\
    &= H_{\theta_\eta}H_{\thetafull}^{-1}\Delta - H_{\thetafull}H_{\thetafull}^{-1}\Delta, \nonumber\\
    &= (H_{\theta_\eta} - H_{\thetafull})H_{\thetafull}^{-1}\Delta. \label{eqn:simplified_unlearned_gradient}
\end{align}
We want to bound the norm of~\eqref{eqn:simplified_unlearned_gradient}.
\begin{align*}
    ||G(\thetaunlearn)||_2 &= ||(H_{\theta_\eta} - H_{\thetafull})H_{\thetafull}^{-1}\Delta||_2 \\
    &\leq ||H_{\theta_\eta} - H_{\thetafull}||_2||H_{\thetafull}^{-1}\Delta||_2
\end{align*}
First, we bound the left term. Recall we define $H$ as the Hessian evaluated over the remainder dataset $D'$. Also note that $\ell''_\textrm{fair}$ and $\frac{\lambda}{2}||\theta||_2^2$ are constant in $\theta$ so $\ell_\textrm{fair}''(\theta_\eta, \{(x_f, y_f)\}_{f \in \{i,j,k,l\}}) - \ell_\textrm{fair}''(\thetafull, \{(x_f, y_f)\}_{f \in \{i,j,k,l\}}) = 0$. Then we have
\begin{align}
    ||H_{\theta_\eta} - H_{\thetafull}||_2 &= ||\sum_{i=1}^{n-m} \nabla^2\ell(\theta_\eta, x_i, y_i) + \lambda(n-m) + \gamma\frac{n-m}{|C_{D'}|}\sum_{i,j,k,l\in C_{D'}}\ell_\textrm{fair}''(\theta_\eta, \{(x_f, y_f)\}_{f \in \{i,j,k,l\}}) \nonumber\\ 
     - \sum_{i=1}^{n-m}& \nabla^2\ell(\thetafull, x_i, y_i) - \lambda(n-m) - \gamma\frac{n-m}{|C_{D'}|}\sum_{i,j,k,l\in C_{D'}}\ell_\textrm{fair}''(\thetafull, \{(x_f, y_f)\}_{f \in \{i,j,k,l\}})||_2, \nonumber \\
    &\leq \sum_{i=1}^{n-m} || \nabla^2 \ell (\theta_\eta, x_i, y_i)-\nabla^2 \ell(\thetafull,x_i,y_i)||_2, \nonumber \\
    &= \sum_{i=1}^{n-m}|| x_i^T(\ell'' (\theta_\eta, x_i, y_i)-\ell''(\thetafull,x_i,y_i)) x_i||_2, \label{eqn:chainrule}  \\
    &\leq \sum_{i=1}^{n-m}||\ell'' (\theta_\eta, x_i, y_i)-\ell''(\thetafull, x_i,y_i)||_2||x_i||_2^2, \nonumber \\
    &\leq \sum_{i=1}^{n-m}\psi ||\theta_\eta - \thetafull ||_2||x_i||_2^2, \label{eqn:lipschitz} \\ 
    &\leq \sum_{i=1}^{n-m}\psi ||\theta_\eta - \thetafull ||_2, \label{eqn:x_simplify} \\
    &= (n-m)\psi ||\theta_\eta - \thetafull ||_2, \nonumber \\
    &= (n-m)\psi ||\thetafull + \eta H_{\thetafull}^{-1}\Delta - \thetafull||_2, \nonumber \\
    &= (n-m)\psi ||\eta H_{\thetafull}^{-1}\Delta||_2, \nonumber \\
    &\leq  (n-m)\psi ||H_{\thetafull}^{-1}\Delta||_2,
\end{align}
where \eqref{eqn:chainrule} comes from chain rule, \eqref{eqn:lipschitz} comes from our assumption $\ell''$ is $\psi$-Lipschitz, and \eqref{eqn:x_simplify} comes from our assumption that $||x_i||_2 \leq 1$.

Thus, we have:
\begin{align}
    ||G(\thetaunlearn)||_2 &\leq ||H_{\theta_\eta} - H_{\thetafull}||_2||H_{\thetafull}^{-1}\Delta||_2, \nonumber \\
    &\leq (n-m)\psi ||H_{\thetafull}^{-1}\Delta||_2^2
\end{align}
We bound the quantity $||H_{\thetafull}^{-1}\Delta||_2 $. Remember we are evaluating our Hessians and gradients over the remainder dataset, $D'$. With a convex $\ell$ and $\ell_\textrm{fair}$ where $\nabla^2\ell, \nabla^2\ell_\textrm{fair}$ are positive semi-definite and regularization of strength $\frac{n\lambda}{2}$, our loss function is $n\lambda$-strongly convex. Thus, our Hessian over the remainder dataset (of size $n-m$) is bounded:
\begin{align}
    ||H_{\theta}||_2 \geq (n-m)\lambda \nonumber
\end{align}
And so is its inverse 
\begin{align}
    ||H_{\theta}^{-1}||_2 \leq \frac{1}{(n-m)\lambda} \label{hessinv}
\end{align}
Finally, we bound $\Delta$. Recall its definition:
\begin{align}
    &\Delta := \sum_{i=m}^{n}\ell'(\thetafull, x_i, y_i) + m\lambda \thetafull+\gamma\frac{n}{|N|} \sum_{(i,j,k,l) \in N}\ell_\textrm{fair}'(\thetafull, \{(x_f, y_f)\}_{f \in \{i,j,k,l\}})  \nonumber\\
    &-\gamma\frac{n-m}{|C_{D'}|}\sum_{(i,j,k,l)\in C_{D'}}\ell_\textrm{fair}'(\thetafull, \{(x_f, y_f)\}_{f \in \{i,j,k,l\}}).
\end{align}
We bound $\thetafull$ to bound our second term. Recall that the gradient over the full dataset at $\thetafull$ equals 0. 
\begin{align}
    0 &= \nabla \mathcal{L}(\thetafull, D) = \sum_{i=1}^{n} \ell'(\thetafull, x_i, y_i) + n\lambda\thetafull + \gamma\frac{n}{|N|} \sum_{i,j,k,l \in N}\ell_\textrm{fair}'(\thetafull, \{(x_f, y_f)\}_{f \in \{i,j,k,l\}}) \label{eqn:stationary_condition}
\end{align}
We expand $\ell'_\textrm{fair}$ and see it is linear in $\theta$. Denote this coefficient $F(i,j,k,l)$.
\begin{align}
    \ell_\textrm{fair}'(\thetafull, \{(x_f, y_f)\}_{f \in \{i,j,k,l\}})  &= 2\cdot\mathbbm{1}[y_i = y_j]\mathbbm{1}[y_k = y_l] (x_i^T x_k - x_i^T x_l - x_j^T x_k + x_j^T x_l)\thetafull \nonumber \\
    &= F(i,j,k,l)\thetafull
\end{align}
We can bound $\thetafull$ by taking the stationary point condition~\eqref{eqn:stationary_condition} and solving for $\thetafull$. Using the assumption the ERM loss is bounded $||\ell'(\theta, x_i, y_i)||_2 \leq g$, we see:
\begin{align}
    \thetafull &= \frac{-\sum_{i=1}^{n} \ell'(\thetafull, x_i, y_i)}{n\lambda + \gamma\frac{n}{|N|} \sum_{i,j,k,l \in N}F(i,j,k,l)}, \nonumber\\
    ||\thetafull||_2 &= \left|\left| \frac{\sum_{i=1}^{n} \ell'(\thetafull, x_i, y_i)}{n\lambda + \gamma\frac{n}{|N|} \sum_{i,j,k,l \in N}F(i,j,k,l)} \right|\right|_2, \nonumber\\
    ||\thetafull||_2 &\leq ng \left|\left|\frac{1}{n\lambda + \gamma\frac{n}{|N|} \sum_{i,j,k,l \in N}F(i,j,k,l)}\right|\right|_2, \nonumber \\
    ||\thetafull||_2 &\leq g \left|\left|\frac{1}{\lambda + \gamma\frac{1}{|N|} \sum_{i,j,k,l \in N}F(i,j,k,l)}\right|\right|_2  \label{thetanorm}
\end{align}
Now, we can bound the norm of $\Delta$. We use the triangle inequality as well as the fact that $|C_{D'}| < |N|$, and the terms $n > m > 0$ to simplify.
\begin{align}
    ||\Delta||_2 &= \left|\left|\sum_{i=m}^{n}\ell'(\thetafull, x_i, y_i) + \thetafull\left[m\lambda + \gamma\left( \frac{n}{|N|} \sum_{i,j,k,l \in N}F(i,j,k,l) - \frac{n-m}{|C_{D'}|}\sum_{i,j,k,l\in C_{D'}}F(i,j,k,l) \right)\right]\right|\right|_2, \nonumber \\
    &\leq mg + ||\thetafull||_2\left|\left|m\lambda + \gamma\left(\frac{n}{|N|} \sum_{i,j,k,l \in N}F(i,j,k,l) - \frac{n-m}{|C_{D'}|}\sum_{i,j,k,l\in C_{D'}}F(i,j,k,l) \right)\right|\right|_2, \nonumber \\
    &\leq mg + ||\thetafull||_2\left|\left|m\lambda + \gamma\left(\frac{n}{|N|} \sum_{i,j,k,l \in N}F(i,j,k,l) - \frac{n-m}{|N|}\sum_{i,j,k,l\in C_{D'}}F(i,j,k,l)\right)\right|\right|_2, \nonumber \\
    &= mg + ||\thetafull||_2\left|\left|m\lambda + \gamma\left(\frac{n}{|N|}\sum_{i,j,k,l\in N \setminus C_{D'}}F(i,j,k,l)+\frac{m}{|N|} \sum_{i,j,k,l \in C_{D'}}F(i,j,k,l) \right)\right|\right|_2, \nonumber \\
    &= gm + ||\thetafull||_2\left|\left|m\lambda + \gamma\left(\frac{m}{|N|} \sum_{i,j,k,l \in N}F(i,j,k,l) + \frac{n-m}{|N|}\sum_{i,j,k,l\in N \setminus C_{D'}}F(i,j,k,l)\right)\right|\right|_2, \nonumber \\
    &\leq mg + ||\thetafull||_2\left|\left|m\lambda + \gamma\frac{m}{|N|} \sum_{i,j,k,l \in N}F(i,j,k,l)\right|\right|_2+||\thetafull||_2\left|\left|\frac{n-m}{|N|}\sum_{i,j,k,l\in N \setminus C_{D'}}F(i,j,k,l)\right|\right|_2
\end{align}
We can plug in our solution for $||\thetafull||_2$ from \eqref{thetanorm} and we see the second term equals $g$.
\begin{align}
    ||\Delta||_2 &\leq 2mg + ||\thetafull||_2\left|\left|\frac{n-m}{|N|}\sum_{i,j,k,l\in N \setminus C_{D'}}F(i,j,k,l)\right|\right|_2 \label{eqn:deltabound}
\end{align}
Next we analyze the final term. Let, $m_a, m_b$ be the number of removals from subgroups a and b respectively and recall that $n_a, n_b$ is the total number of elements in subgroups a and b respectively. Recall our assumption $||x_i||_2 \leq 1$. Note that $N \setminus C_{D'}$ is the union of two sets; first, the set of removed points from subgroup a ($r_a$) and their interactions with all points in subgroup b ($G_b$), and second the set of removed points from subgroup b ($r_b$) and their interactions with all points in subgroup a ($G_a$). Then, we can compute the cardinality $|N \setminus C_{D'}| = m_a^2n_b^2 + m_b^2n_a^2 - m_a^2m_b^2.$ Note that $|N| = n_a^2n_b^2$. 
\begin{align}
    &||\thetafull||_2\left|\left|\frac{n-m}{|N|}\sum_{i,j,k,l\in N \setminus C_{D'}}F(i,j,k,l) \right|\right|_2 \nonumber\\
    &= ||\thetafull||_2\left|\left|\frac{n-m}{|N|} \sum_{i,j,k,l \in N \setminus C_{D'}}2\cdot\mathbbm{1}[y_i = y_j]\mathbbm{1}[y_k = y_l] (x_i^T x_k - x_i^T x_l - x_j^T x_k + x_j^T x_l) \right|\right|_2, \nonumber \\
    &\leq ||\thetafull||_2\left|\left|\frac{8|N \setminus C_{D'}|(n-m)}{|N|} \right|\right|_2, \nonumber \\
    &= ||\thetafull||_2\left|\left|\frac{8(m_a^2n_b^2 + m_b^2n_a^2 - m_a^2m_b^2)(n-m)}{(n_a^2n_b^2)} \right|\right|_2. \label{eqn:delta_rhs}
\end{align}
Combining \eqref{eqn:deltabound} and \eqref{eqn:delta_rhs}, 
\begin{align}
    ||\Delta||_2 &\leq 2mg + ||\thetafull||_2\left|\left|\frac{8(m_a^2n_b^2 + m_b^2n_a^2 - m_a^2m_b^2)(n-m)}{(n_a^2n_b^2)} \right|\right|_2.
\end{align}
The final form of our gradient norm $||\mathcal{L}(\thetaunlearn, D')||_2 = ||G(\thetaunlearn)||_2$ is the following.
\begin{align}
    ||G(\thetaunlearn)||_2 &\leq (n-m)\psi ||H_{\thetafull}^{-1}\Delta||_2^2, \nonumber \\ 
    &\leq (n-m)\psi ||H_{\thetafull}^{-1}||^2_2||\Delta||_2^2, \nonumber \\
    &\leq (n-m)\psi \frac{1}{(n-m)^2\lambda^2}(2mg + ||\thetafull||_2\left|\left|\frac{8(m_a^2n_b^2 + m_b^2n_a^2 - m_a^2m_b^2)(n-m)}{(n_a^2n_b^2)} \right|\right|_2)^2, \nonumber \\
    &= \frac{\psi}{\lambda^2(n-m)}\left(2mg + ||\thetafull||_2\left|\left|\frac{8(m_a^2n_b^2 + m_b^2n_a^2 - m_a^2m_b^2)(n-m)}{(n_a^2n_b^2)} \right|\right|_2\right)^2. \label{eqn:final_gradient} && \square
\end{align}
Assume the number of samples in subgroups a and b $(n_a, n_b)$ are relatively balanced ($O(n)$), and the number of unlearned samples $(m_a, m_b)$ are sufficiently low, such that $m_a = O(\sqrt{n_a}), m_b = O(\sqrt{n_b})$. Under these assumptions, we see that the term in the parenthesis is $O(1)$, so~\eqref{eqn:final_gradient} is $O(1/n).$ When we let $m=1$, and without loss of generality $m_a = 1, m_b = 0$, we exactly recover Thm. \ref{thm:eps_delta} in the main paper.
\begin{lemma}[\citet{guoCertifiedDataRemoval2020}] \label{guo_noise_bound}
    Let A be the learning algorithm that returns the unique optimum of the perturbed loss $\mathcal{L}(\theta, D) + \textbf{b}^T\theta$, and let M be the an unlearning mechanism that outputs $\thetaunlearn$. Suppose that $||\nabla\mathcal{L}(\thetaunlearn, D')||_2 \leq \epsilon'$ for some $\epsilon' > 0$. Then, we have the following guarantees for M:

    \begin{enumerate}
        \item If \textbf{b} is drawn from a distribution with density $p(\textbf{b}) \propto \exp{-\frac{\epsilon}{\epsilon'}||\textbf{b}||_2}$, then M is $\epsilon$-unlearnable for A;
        \item If $\textbf{b} \sim N(0, k\epsilon'/\epsilon)^d$ with $k > 0$, then M is $(\epsilon, \delta)$-unlearnable for A with $\delta = 1.5\exp(-k^2/2)$
    \end{enumerate}
\end{lemma}
\subsubsection{Proof of Theorem~\ref{thm:eps_delta}:}
By combining Lemmas~\ref{lem:bounded_gradient} and \ref{guo_noise_bound}, we show that the loss gradient over the remaining dataset, evaluated at the unlearned parameter $\thetaunlearn$ is bounded, and we can directly plug this into Lemma~\ref{guo_noise_bound} to achieve $(\delta, \epsilon)$-unlearning for some $\delta, \epsilon > 0$. \hfill $\square$
\subsection{Data-Dependent Bound}\label{sec:datadependentbound} From Eqn.~\eqref{eqn:simplified_unlearned_gradient}, we are computing a bound on the norm of the following:
\begin{align}
    (H_{\theta_\eta} - H_{\thetafull})H_{\thetafull}^{-1}\Delta. \label{eqn:datadependentbound}
\end{align}
The Hessian for $\mathcal{L}(\theta, D')$ is 
\begin{align*}
    X^{T}W_\theta X + \lambda(n-m) + \gamma\frac{n-m}{|C_{D'}|}\sum_{i,j,k,l\in C_{D'}}\ell_\textrm{fair}''(\theta, \{(x_f, y_f)\}_{f \in \{i,j,k,l\}})
\end{align*}
Where $X$ is the data matrix of samples in $D'$, and $W_\theta$ is a diagonal matrix where $W_{ii} = \ell''(\theta, x_i, y_i)$. Then, Eqn.~\eqref{eqn:datadependentbound} becomes
\begin{align}
    (X^T(W_{\theta_\eta} - W_{\thetafull})X)H_{\thetafull}^{-1}\Delta,
\end{align}
which by \citep{guoCertifiedDataRemoval2020}, Corollary 1, shows that 
\begin{align}
    ||\nabla \mathcal{L}(\thetafull; D')||_2 = ||(X^T(W_{\theta_\eta} - W_{\thetafull})X)H_{\thetafull}^{-1}\Delta||_2 \leq \gamma ||X||_2||H^{-1}_{\thetafull}\Delta||_2||XH^{-1}_{\thetafull}\Delta||_2.
\end{align}

\subsection{Proof of Theorem~\ref{thm:fairnesG_bound}}

We begin with a useful corollary of Lemma \ref{lem:bounded_gradient}:

\begin{corollary}[$||\thetaunlearn - \thetaretrain||_2$ is bounded.] \label{corr1}
Recall $\mathcal{L}(\theta, D')$ is $(n-m)\lambda$-strongly convex. Then, by strong convexity,
\begin{align}
    (n-m)\lambda||\thetaunlearn - \thetaretrain||_2^2 &\leq (\nabla \mathcal{L}(\thetaunlearn, \mathcal{D'}) - \nabla\mathcal{L}(\thetaretrain, \mathcal{D'})^T(\thetaunlearn - \thetaretrain), \nonumber \\
    &= (\nabla \mathcal{L}(\thetaunlearn, \mathcal{D'}))^T(\thetaunlearn - \thetaretrain), \nonumber \\
    &\leq ||\nabla\mathcal{L}(\thetaunlearn, \mathcal{D'})||_2 ||\thetaunlearn - \thetaretrain||_2, \nonumber \\
    (n-m)\lambda||\thetaunlearn - \thetaretrain||_2 &\leq  \frac{\psi}{\lambda^2(n-m)}\left(2mg + ||\thetafull||_2\left|\left|\frac{8(m_a^2n_b^2 + m_b^2n_a^2 - m_a^2m_b^2)(n-m)}{(n_a^2n_b^2)} \right|\right|_2\right)^2, \nonumber \\
    ||\thetaunlearn - \thetaretrain||_2 &\leq \frac{\psi}{\lambda^3(n-m)^2}\left(2mg + ||\thetafull||_2\left|\left|\frac{8(m_a^2n_b^2 + m_b^2n_a^2 - m_a^2m_b^2)(n-m)}{(n_a^2n_b^2)} \right|\right|_2\right)^2
\end{align}
\end{corollary}
Next, we introduce the lemmas we will use in this proof.
\begin{lemma}[The interior angle between $\thetaretrain$ and $\thetaunlearn$ is bounded.]\label{boundedangle}
    Let $||\thetaretrain - \thetaunlearn||_2 \leq \kappa$. Define the angle between $\thetaretrain$ and $\thetaunlearn$ as $\phi$. Then,
    \begin{align}
        \phi &\leq \arccos\left(\frac{||\thetaretrain||_2^2 + ||\thetaunlearn||_2^2 - \kappa^2}{2||\thetaretrain||_2||\thetaunlearn||_2}\right), \nonumber \\
        &\leq \arccos\left(\frac{||\thetaretrain||_2^2 + (||\thetaretrain||_2 - \kappa)^2 - \kappa^2}{2||\thetaretrain||_2(||\thetaretrain||_2+\kappa)}\right)\nonumber\\
        &= \arccos\left(\frac{||\thetaretrain||_2 - \kappa}{||\thetaretrain||_2+\kappa}\right)
    \end{align}
\end{lemma}

The first inequality follows directly from applying the cosine angle formula to the triangle described by $\thetaretrain$, $\thetaunlearn$, and $\thetaretrain-\thetaunlearn$ and the fact that $||\thetaretrain-\thetaunlearn||_2 \leq \kappa$. Then, we use the triangle inequality to show 
\begin{align}
    ||\thetaunlearn||_2 \leq ||\thetaretrain||_2 + ||\thetaunlearn-\thetaretrain||_2 \leq ||\thetaretrain||_2 + \kappa, \nonumber \\
    ||\thetaunlearn||_2 \geq ||\thetaretrain||_2 - ||\thetaretrain-\thetaunlearn||_2 \geq ||\thetaretrain||_2 - \kappa. \nonumber
\end{align}
This lemma implies that as $\kappa$ approaches $0$, the angle $\phi$ approaches zero as well.

\begin{lemma}[Volume of internal segment of a d-sphere]\label{boundedvolume}
    Let $S$ be a $d$-sphere with radius 1 and volume $V = \frac{\pi^{d/2}}{\Gamma(\frac{d}{2}+1)}$. Then, let $\theta_1$, $\theta_2$ be normal vectors defining two hyperplanes such that their intersection lies tangent to the surface of $S$ with internal angle $\phi$ such that the angle bisector to $\phi$ also bisects $S$. Then, the volume of the region of intersection $B$ between the half spaces defined by the hyperplanes at $\phi$ within $S$ equals
    \begin{align}
        B &= V - 2\int_{\mu}^{1}\frac{\pi^{\frac{d-1}{2}}}{\Gamma(\frac{d+1}{2})}(1-y^2)^{d-1} dy, \nonumber \\
        \mu &= \frac{1}{2}\sin{\frac{\phi}{2}}. \nonumber
    \end{align}
\end{lemma}

We take the full volume $V$ and subtract off the volume of the spherical segment defined as the integral of the volume for a $d-1$-sphere over various radius values along the hyperplane to the surface of the sphere. Note that as $\phi \rightarrow 0, \mu \rightarrow 0$, and therefore $B \rightarrow 0$ as the right term approaches $V$.
\subsubsection{Proof of Theorem~\ref{thm:fairnesG_bound}:}
Let $\thetaretrain$ have a mean absolute equalized odds upper bounded by $\alpha$, or
\begin{align}
    AEOD(\thetaretrain) \leq \alpha.
\end{align}
This implies
\begin{align}
    |P(h_{\thetaretrain}(X) = 1 \mid S=a,Y=1)-P(h_{\thetaretrain}(X) = 1 \mid S=b,Y=1)| \leq \alpha, \\
    |P(h_{\thetaretrain}(X) = 1 \mid S=a,Y=0)-P(h_{\thetaretrain}(X) = 1 \mid S=b,Y=0)| \leq \alpha. 
\end{align}
Define $A_{\thetaretrain} = \{x : h_{\thetaretrain}(x) = 1\}, A_{\thetaunlearn} = \{x : h_{\thetaunlearn}(x) = 1\}$. Then,
\begin{align}
    |P(A_{\thetaretrain} \mid S=a,Y=1)-P(A_{\thetaretrain} \mid S=b, Y=1)| \leq \alpha, \\
    |P(A_{\thetaretrain} \mid S=a,Y=0)-P(A_{\thetaretrain} \mid S=b,Y=0)| \leq \alpha. 
\end{align}
Now, define the regions of agreement as $A = A_{\thetaretrain} \cap A_{\thetaunlearn}, B = A_{\thetaretrain}^C \cap A_{\thetaunlearn}^C$ and the regions of disagreement as $C = A_{\thetaretrain}^C \cap A_{\thetaunlearn}, D = A_{\thetaretrain} \cap A_{\thetaunlearn}^C$. These events are disjoint, so we can decompose our conditional probability measures in to sums over these events and express $P(A_{\thetaunlearn}\mid S,Y)$ in terms of $P(A_{\thetaretrain}\mid S,Y)$
\begin{align}
    P(A_{\thetaretrain}\mid S,Y) = P(A\mid S,Y) + P(D\mid S,Y), \nonumber\\
    P(A_{\thetaunlearn}\mid S,Y) = P(A\mid S,Y) + P(C\mid S,Y), \nonumber\\
    P(A_{\thetaunlearn}\mid S,Y) = P(A_{\thetaretrain}\mid S,Y) + P(C\mid S,Y) - P(D\mid S,Y).
\end{align}
Then, we can bound the absolute difference:
\begin{align}
    |P(A_{\thetaunlearn}&\mid S=a,Y=1) - P(A_{\thetaunlearn}\mid S=b,Y=1)| \nonumber\\
    =& |P(A_{\thetaretrain}\mid S=a,Y=1)-P(A_{\thetaretrain}\mid S=b,Y=1) \nonumber\\
    &+P(C \mid, S=a, Y=1) - P(C \mid, S=b, Y=1) \nonumber\\
    &- P(D \mid, S=a, Y=1) + P(D \mid, S=b, Y=1)|, \nonumber\\
    \leq& |P(A_{\thetaretrain}\mid S=a,Y=1)-P(A_{\thetaretrain}\mid S=b,Y=1) \nonumber\\
    &+ |P(C \mid, S=a, Y=1) - P(C \mid, S=b, Y=1)| \nonumber\\
    &+ |P(D \mid, S=b, Y=1) - P(D \mid, S=a, Y=1)|, \nonumber\\
    \leq& \alpha + |P(C \mid, S=a, Y=1) - P(C \mid, S=b, Y=1)| \nonumber\\
    &+ |P(D \mid, S=b, Y=1) - P(D \mid, S=a, Y=1)|. \label{basebound}
\end{align}
Next, we relate the probability measures $P(\cdot \mid S, Y)$ to the size of their event spaces to establish a bound on the second and third terms in Equation~\ref{basebound}. We assume for all events $x \subseteq \mathcal{X}$, $P(x | S, Y) \leq c \frac{|x|}{|\mathcal{X}|}$, or in other words, that our probability mass is relatively well distributed up to a constant. Then, computing $|C|$ and $|D|$ will provide upper bounds on their conditional likelihoods.

Recall our assumption that $||x_i||_2 \leq 1$. This means all of our data lies within a unit $d$-sphere we call $S$, which defines our event space $\mathcal{X}$ and has volume $V = |\mathcal{X}| = \frac{\pi^{d/2}}{\Gamma(\frac{d}{2}+1)}$. We can define events $A_{\thetaretrain}$ and $A_{\thetaunlearn}$ the halfspaces defined by $\thetaretrain$ and $\thetaunlearn$ intersected with $S$. Then, $A, B, C, \text{ and } D$ partition $S$ into four disjoint volumes. We seek to bound the volume of $C$ and $D$. First observe that the largest volume for $C + D$ occurs when $\thetaretrain, \thetaunlearn$ intersect tangent to $S$ and create a region where the angle between $\thetaretrain$ and $\thetaunlearn$ is bisected by a hyperplane that also bisects $S$. This creates a region where one of $C,D$ is empty and the other draws a large slice through $S$. Define the angle between $\thetaretrain$ and $\thetaunlearn$ as $\phi$. Without loss of generality, assume $|D| = 0$. Then, by Lemmas~\ref{boundedangle} and~\ref{boundedvolume}, if $||\thetaretrain - \thetaunlearn||_2 \leq d$, then $\phi$ is bounded and thus $|C|$ is bounded.
\begin{align}
    |C| &= V - 2\int_{\mu}^{1}\frac{\pi^{\frac{d-1}{2}}}{\Gamma(\frac{d+1}{2})}(1-y^2)^{d-1} dy,
    \\
    \phi &\leq \arccos\left(\frac{||\thetaretrain||_2 - \kappa}{||\thetaretrain||_2+\kappa}\right),
\end{align}
where
\begin{align}
    \mu &= \frac{1}{2}\sin{\frac{\phi}{2}}, \nonumber \\
    &= \frac{1}{2}\sqrt{\frac{1-\cos{\phi}}{2}}, \nonumber \\
    &\leq \frac{\sqrt{\kappa}}{2}.
\end{align}
by the double angle formula for cosine.

Returning to Equation~\ref{basebound}, we have
\begin{align}
    |P(A_{\thetaunlearn}&\mid S=a,Y=1) - P(A_{\thetaunlearn}\mid S=b,Y=1)| \nonumber \\
    &\leq \alpha + |P(C \mid, S=a, Y=1) - P(C \mid, S=b, Y=1)| \nonumber\\
    &+ |P(D \mid, S=b, Y=1) - P(D \mid, S=a, Y=1)|, \nonumber \\
    &\leq \alpha + \max(P(C \mid, S=a, Y=1), P(C \mid, S=b, Y=1)) \nonumber \\
    &+ \max(P(D \mid, S=b, Y=1), P(D \mid, S=a, Y=1)), \nonumber \\
    &\leq \alpha + c\frac{|C|}{|\mathcal{X}|} + c\frac{|D|}{|\mathcal{X}|}, \nonumber \\
    &= \alpha + c\frac{V- 2\int_{\mu}^{1}\frac{\pi^{\frac{d-1}{2}}}{\Gamma(\frac{d+1}{2})}(1-y^2)^{d-1} dy}{V}, \nonumber \\
    &= \alpha + c\left(1- \frac{2\int_{\mu}^{1}\frac{\pi^{\frac{d-1}{2}}}{\Gamma(\frac{d+1}{2})}(1-y^2)^{d-1} dy}{V}\right),
\end{align}
where 
\begin{align}
    \mu \leq \frac{\sqrt{\kappa}}{2}. \nonumber
\end{align}
By a symmetric argument, we also see
\begin{align}
    |P(A_{\thetaunlearn}&\mid S=a,Y=0) - P(A_{\thetaunlearn}\mid S=b,Y=0)| \leq \alpha + c\left(1- \frac{2\int_{\mu}^{1}\frac{\pi^{\frac{d-1}{2}}}{\Gamma(\frac{d+1}{2})}(1-y^2)^{d-1} dy}{V}\right).
\end{align}
And thus we have that the absolute mean equalized odds of $\thetaunlearn$ is bounded:
\begin{align}
    \frac{1}{2}\big[|P(A_{\thetaunlearn}&\mid S=a,Y=1) - P(A_{\thetaunlearn}\mid S=b,Y=1)| \nonumber \\
    &+ |P(A_{\thetaunlearn}\mid S=a,Y=1) - P(A_{\thetaunlearn}\mid S=b,Y=1)|\big], \nonumber \\
    &\leq \alpha + c\left(1- \frac{2\int_{\mu}^{1}\frac{\pi^{\frac{d-1}{2}}}{\Gamma(\frac{d+1}{2})}(1-y^2)^{d-1} dy}{V}\right), 
\end{align}
where
\begin{align}
    h \leq \frac{\sqrt{\kappa}}{2}. \tag*{$\square$} \nonumber
\end{align}
\subsection{Incompatibility of existing unlearning methods.}
\label{sec:incompatible}

Existing unlearning methods require a decomposition of the loss term as $\mathcal{L}(\theta) = \frac{1}{n} \sum_{i=1}^n \ell(\theta)$. For example, the most similar method to Fair Unlearning comes from~\citet{guoCertifiedDataRemoval2020}, and their unlearning step involves a second order approximation over $\Delta = \lambda \thetafull + \nabla\ell(\thetafull, x_n, y_n)$ to unlearn datapoint $(x_n, y_n)$. Similarly to our proof of Lemma~\ref{lem:bounded_gradient}, adding $\Delta$ with the gradient of $\thetaunlearn$ over the remaining dataset $\mathcal{L}(\thetaunlearn, D')$ results in a cancellation that enables the rest of their proof to bound the gradient. However, in their case $\Delta$ is directly corresponds to the gradient over sample $(x_n, y_n)$ only because of the decomposability of their loss term. This behavior makes all unlearning over decomposable loss functions the same: compute a Newton step $H^{-1}\Delta$ where $H$ is the hessian over $D'$ and $\Delta$ is the gradient over $R$. In our case our $\Delta$ is more complex because the fairness loss involves the interactions between each sample and samples in the opposite subgroup. Thus, removing one point requires removing a set of terms from the fairness loss, and thus $\Delta$ must be a more complex term to account for this rather than simply the gradient over $(x_n, y_n)$.

\subsection{Extension of fairness penalty to other definitions of group fairness.} Recall the definition of Equalized Odds:
\begin{align}
    \Pr(h_\theta(X) = 1 \mid S = a, Y = y) = \Pr(h_\theta(X) = 1 \mid S = b, Y = y).
\end{align}
In contrast, Demographic Parity does not condition on the true label, $Y$.
\begin{align}
    \Pr(h_\theta(X) = 1 \mid S = a) = \Pr(h_\theta(X) = 1 \mid S = b).
\end{align}
Thus, to achieve this, we remove the indicator variable in our loss function.
\begin{align}
    &\mathcal{L}_\textrm{Demo. Par.}(\theta, D) := \left(\frac{1}{n_an_b} \sum_{i \in G_a}\sum_{j \in G_b}(\langle x_i,\theta\rangle - \langle x_j, \theta\rangle) \right)^2.
\end{align}
Equality of opportunity conditions only on true positive labels, $Y = 1$. 
\begin{align}
    \Pr(h_\theta(X) = 1 \mid S = a, Y=1) = \Pr(h_\theta(X) = 1 \mid S = b, Y=1).
\end{align}
Thus, to achieve this, we modify the indicator variable in our loss function to only consider true positives.
\begin{align}
    &\mathcal{L}_\textrm{Eq. Opp.}(\theta, D) := \left(\frac{1}{n_an_b} \sum_{i \in G_a}\sum_{j \in G_b} \mathbbm{1}[y_i = y_j = 1](\langle x_i,\theta\rangle - \langle x_j, \theta\rangle) \right)^2.
\end{align}
\subsection{Experiment details.} \label{sec:hyperparameters} We run all of our experiments on a 32GB Tesla V100 GPU, but using tabular data and small models, these results can be replicated with weaker hardware. When training, we use $\sigma = 1$ for our objective perturbation and compute $\epsilon, \delta$ bounds with $\delta = 0.0001$. We report the following hyperparameters for our final results in the main paper:
\begin{table}[h!]
    \centering
    \begin{tabular}{ccc}
    \toprule
         & $\ell_2$ Penalty ($\lambda$) & Fairness Penalty ($\gamma$) \\
    \midrule
       COMPAS  & $1\mathrm{e}-4$ & $1\mathrm{e}1$\\
       Adult  & $1\mathrm{e}-4$ & $1\mathrm{e}0$\\
       HSLS  & $1\mathrm{e}-2$ & $1\mathrm{e}1$\\
    \bottomrule
    \end{tabular}
    \caption{Hyperparameters for experiments in paper.}
    \label{tab:hyperparameters}
\end{table}
\subsection{Runtimes.} \label{sec:runtimes}
Below we include runtime comparisons (Table \ref{tab:runtimes}) for our method with naive retraining as well as precomputation times for COMPAS, Adult, and HSLS (Table \ref{tab:precomputation_times}). Our method achieves significant speedup over retraining a model.
\begin{table*}[h!]
\centering
\caption{Runtime Comparisons for Proposed Method vs Retraining.}
\label{tab:runtimes}
\resizebox{\linewidth}{!}{
\begin{tabular}{lcccccc} 
\toprule
& \multicolumn{2}{c}{COMPAS}             & \multicolumn{2}{c}{Adult}             & \multicolumn{2}{c}{HSLS}               \\
Method                                & 5\% Unlearned       & 20\% Unlearned       & 5\% Unlearned       & 20\% Unlearned      & 5\% Unlearned       & 20\% Unlearned       \\ 
\midrule
Retraining               & $3.500 \pm 0.340$ s & $3.180 \pm 0.142$ s & $36.796 \pm 1.090$ s & $31.175 \pm 1.040$ s & $14.963 \pm 0.396$ s& $12.362 \pm 0.795$ s \\
Fair Unlearning (Ours)                            & $0.170 \pm 0.010$ s & $0.164 \pm 0.003$ s& $8.800 \pm 0.511$ s & $8.483 \pm 0.938$ s & $6.014 \pm 0.249$ s & $5.704 \pm 0.015$ s  \\
\bottomrule
\end{tabular}
}
\end{table*}

\begin{table*}[h!]
\centering
\caption{Precomputation Runtime Comparisons for Hessian Inversion.}
\label{tab:precomputation_times}
\begin{tabular}{lccc} 
\toprule
& COMPAS & Adult & HSLS \\
Precomputation Time: & $0.182 \pm 0.006$s & $6.210 \pm 0.041$s & $4.964 \pm 0.117$s\\
\bottomrule
\end{tabular}
\end{table*}

\newpage

\subsection{Test accuracy when unlearning.}\label{appendix:accuracy}
In Tables \ref{tab:unlearning at random}, \ref{tab:unlearning from minority}, and \ref{tab:unlearning from majority}, we report test accuracy at various amounts of unlearning with standard deviation.
\begin{table*}[h!]
\centering
\caption{Test accuracy when unlearning at random.}
\label{tab:unlearning at random}
\resizebox{\linewidth}{!}{%
\begin{tabular}{lcccccc} 
\toprule
                                      & \multicolumn{2}{c}{COMPAS}             & \multicolumn{2}{c}{Adult}             & \multicolumn{2}{c}{HSLS}               \\
Method                                & 5\% Unlearned       & 20\% Unlearned       & 5\% Unlearned       & 20\% Unlearned      & 5\% Unlearned       & 20\% Unlearned       \\ 
\midrule
Full Training (BCE)                & $0.652 \pm 0.000$ & $0.652 \pm 0.000 $ & $0.817 \pm 0.000$ & $0.816 \pm 0.000$ & $0.724 \pm 0.000$ & $0.724 \pm 0.000$  \\
Retraining (BCE)            & $0.652 \pm 0.001$ & $0.654 \pm 0.003$  & $0.817 \pm 0.000$ & $0.817 \pm 0.001$ & $0.724 \pm 0.002$ & $0.724 \pm 0.002$  \\
Newton (\citet{guoCertifiedDataRemoval2020})  & $0.652 \pm 0.001$ & $0.654 \pm 0.003$  & $0.817 \pm 0.000$ & $0.817 \pm 0.001$ & $0.724 \pm 0.002$ & $0.724 \pm 0.002$  \\
SISA (\citet{bourtouleMachineUnlearning2020}) & $0.653 \pm 0.001$ & $0.656 \pm 0.004$  & $0.817 \pm 0.000$ & $0.817 \pm 0.001$ & $0.725 \pm 0.001$ & $0.724 \pm 0.003$  \\
PRU (\citet{izzoApproximateDataDeletion})     & $0.688 \pm 0.005$ & $0.691 \pm 0.001$  & $0.810 \pm 0.003$ & $0.810 \pm 0.001$ & $0.726 \pm 0.004$ & $0.729 \pm 0.000$  \\ 
\midrule
Full Training (Fair)                   & $0.647 \pm 0.000$ & $0.647 \pm 0.000$  & $0.819 \pm 0.000$ & $0.819 \pm 0.000$ & $0.713 \pm 0.000$ & $0.713 \pm 0.000$  \\
Retraining (Fair)               & $0.646 \pm 0.002$ & $0.642 \pm 0.003$  & $0.819 \pm 0.000$ & $0.819 \pm 0.001$ & $0.713 \pm 0.002$ & $0.713 \pm 0.002$  \\
Fair Unlearning (Ours)                            & $0.646 \pm 0.002$ & $0.642 \pm 0.003$  & $0.819 \pm 0.000$ & $0.819 \pm 0.001$ & $0.713 \pm 0.002$ & $0.713 \pm 0.002$  \\
\bottomrule
\end{tabular}
}
\end{table*}

\begin{table}[h!]
\centering
\caption{Test accuracy when unlearning from minority subgroups.}
\label{tab:unlearning from minority}
\resizebox{\linewidth}{!}{%
\begin{tabular}{lcccccc} 
\toprule
                                   & \multicolumn{2}{c}{COMPAS}            & \multicolumn{2}{c}{Adult}             & \multicolumn{2}{c}{HSLS}               \\
Method                             & 5\% Unlearned       & 10\% Unlearned      & 5\% Unlearned       & 10\% Unlearned      & 5\% Unlearned       & 10\% Unlearned       \\ 
\midrule
Full Training (BCE)                & $0.652 \pm 0.000$ & $0.652 \pm 0.000$ & $0.821 \pm 0.000$ & $0.821 \pm 0.000$ & $0.729 \pm 0.000$ & $0.729 \pm 0.000$  \\
Retraining (BCE)                   & $0.651 \pm 0.002$ & $0.652 \pm 0.001$ & $0.821 \pm 0.000$ & $0.822 \pm 0.000$ & $0.729 \pm 0.001$ & $0.730 \pm 0.001$  \\
Newton (\citet{guoCertifiedDataRemoval2020}) & $0.651 \pm 0.002$ & $0.652 \pm 0.001$ & $0.821 \pm 0.000$ & $0.823 \pm 0.000$ & $0.729 \pm 0.001$ & $0.730 \pm 0.002$  \\
SISA (\citet{bourtouleMachineUnlearning2020})   & $0.653 \pm 0.001$ & $0.654 \pm 0.002$ & $0.822 \pm 0.000$ & $0.823 \pm 0.000$ & $0.728 \pm 0.001$ & $0.728 \pm 0.001$  \\
PRU (\citet{izzoApproximateDataDeletion})    & $0.661 \pm 0.007$ & $0.666 \pm 0.007$ & $0.819 \pm 0.004$ & $0.815 \pm 0.035$ & $0.720 \pm 0.004$ & $0.724 \pm 0.003$  \\ 
\midrule
Full Training (Fair)               & $0.653 \pm 0.000$ & $0.653 \pm 0.000$ & $0.812 \pm 0.000$ & $0.812 \pm 0.000$ & $0.723 \pm 0.000$ & $0.723 \pm 0.000$  \\
Retraining (Fair)                  & $0.653 \pm 0.001$ & $0.661 \pm 0.005$ & $0.813 \pm 0.000$ & $0.816 \pm 0.001$ & $0.713 \pm 0.002$ & $0.713 \pm 0.002$  \\
Fair Unlearning (Ours)             & $0.652 \pm 0.001$ & $0.661 \pm 0.004$ & $0.814 \pm 0.000$ & $0.816 \pm 0.002$ & $0.724 \pm 0.001$ & $0.721 \pm 0.001$  \\
\bottomrule
\end{tabular}
}
\end{table}

\begin{table}[h!]
\centering
\caption{Test accuracy when unlearning from majority subgroups.}
\label{tab:unlearning from majority}
\resizebox{\linewidth}{!}{%
\begin{tabular}{lcccccc} 
\toprule
                                      & \multicolumn{2}{c}{COMPAS}            & \multicolumn{2}{c}{Adult}             & \multicolumn{2}{c}{HSLS}               \\
Method                                & 5\% Unlearned       & 20\% Unlearned      & 5\% Unlearned       & 20\% Unlearned      & 5\% Unlearned       & 20\% Unlearned       \\ 
\midrule
Full Training (BCE)                & $0.654 \pm 0.000$ & $0.654 \pm 0.000$ & $0.814 \pm 0.000$ & $0.814 \pm 0.000$ & $0.738 \pm 0.000$ & $0.738 \pm 0.000$  \\
Retraining (BCE)            & $0.654 \pm 0.001$ & $0.652 \pm 0.002$ & $0.814 \pm 0.000$ & $0.814 \pm 0.000$ & $0.736 \pm 0.000$ & $0.736 \pm 0.002$  \\
Newton (\citet{guoCertifiedDataRemoval2020})  & $0.654 \pm 0.001$ & $0.652 \pm 0.002$ & $0.814 \pm 0.000$ & $0.814 \pm 0.001$ & $0.736 \pm 0.000$ & $0.735 \pm 0.002$  \\
SISA (\citet{bourtouleMachineUnlearning2020}) & $0.654 \pm 0.002$ & $0.653 \pm 0.004$ & $0.814 \pm 0.000$ & $0.815 \pm 0.001$ & $0.734 \pm 0.001$ & $0.733 \pm 0.002$  \\
PRU (\citet{izzoApproximateDataDeletion})     & $0.696 \pm 0.006$ & $0.688 \pm 0.003$ & $0.812 \pm 0.001$ & $0.811 \pm 0.002$ & $0.723 \pm 0.012$ & $0.731 \pm 0.004$  \\ 
\midrule
Full Training (Fair)                   & $0.666 \pm 0.000$ & $0.666 \pm 0.000$ & $0.815 \pm 0.000$ & $0.815 \pm 0.000$ & $0.725 \pm 0.000$ & $0.725 \pm 0.000$  \\
Retraining (Fair)               & $0.663 \pm 0.004$ & $0.653 \pm 0.002$ & $0.815 \pm 0.000$ & $0.813 \pm 0.000$ & $0.724 \pm 0.001$ & $0.725 \pm 0.002$  \\
Fair Unlearning (Ours)                            & $0.663 \pm 0.004$ & $0.653 \pm 0.003$ & $0.815 \pm 0.000$ & $0.814 \pm 0.001$ & $0.724 \pm 0.001$ & $0.725 \pm 0.002$  \\
\bottomrule
\end{tabular}
}
\end{table}

\newpage 

\subsection{Fairness-Accuracy Tradeoffs.}

Below we report the impact of the fairness regularizer on the fairness and accuracy performance of the model without unlearning. We see that the fairness regularizer has a tangible impact and controls a tradeoff between the fairness and accuracy of the model. This motivates the selection of our fairness penalty, $\gamma$, as detailed in \ref{sec:hyperparameters} where we see a good balance between fairness and accuracy performance.

\begin{figure}[h]
    \centering
    \begin{subfigure}{0.47\linewidth}
         \includegraphics[width=\linewidth]{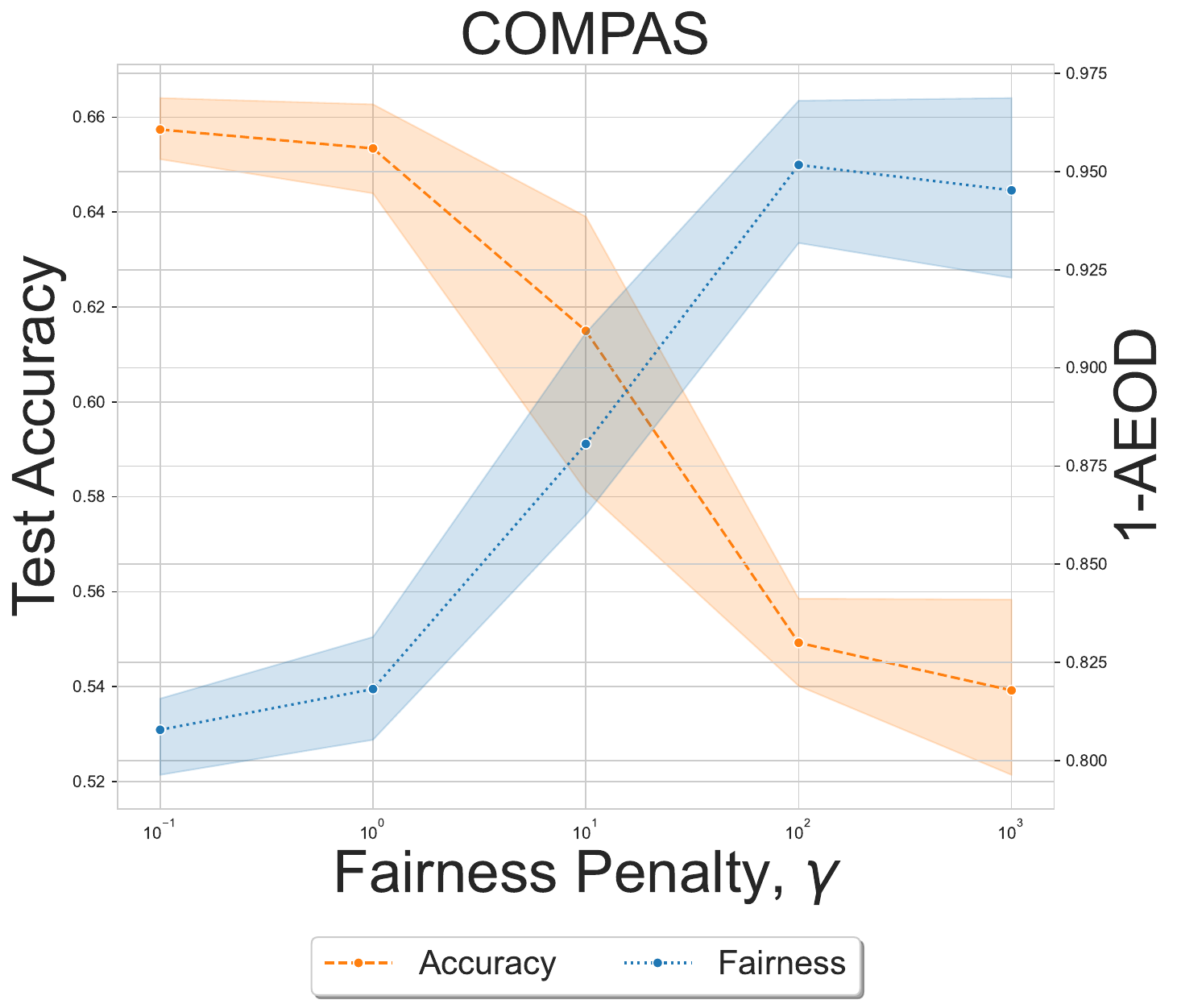}
     \end{subfigure}
     \begin{subfigure}{0.47\linewidth}
         \includegraphics[width=\linewidth]{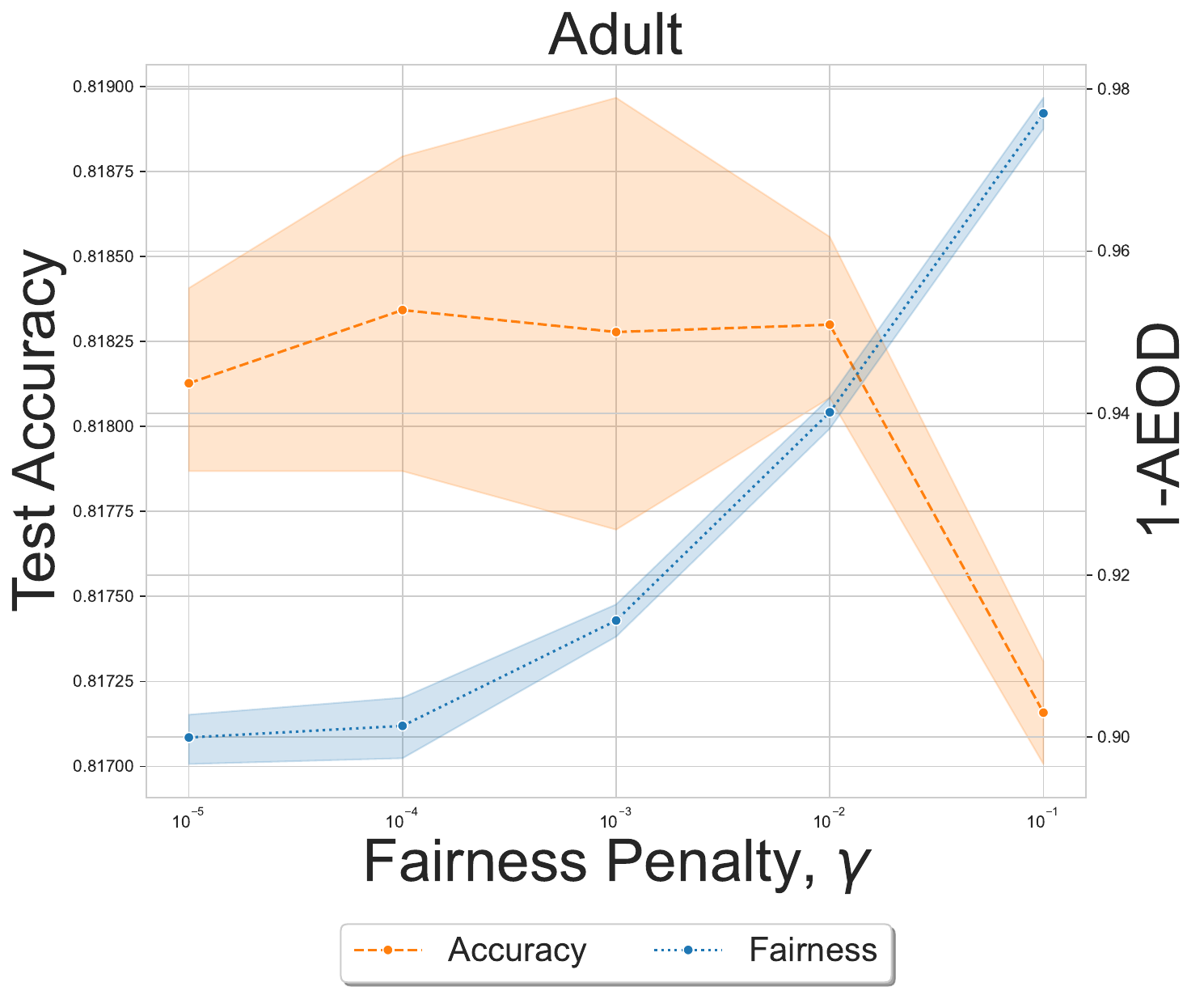}
     \end{subfigure}
     \begin{subfigure}{0.47\linewidth}
         \includegraphics[width=\linewidth]{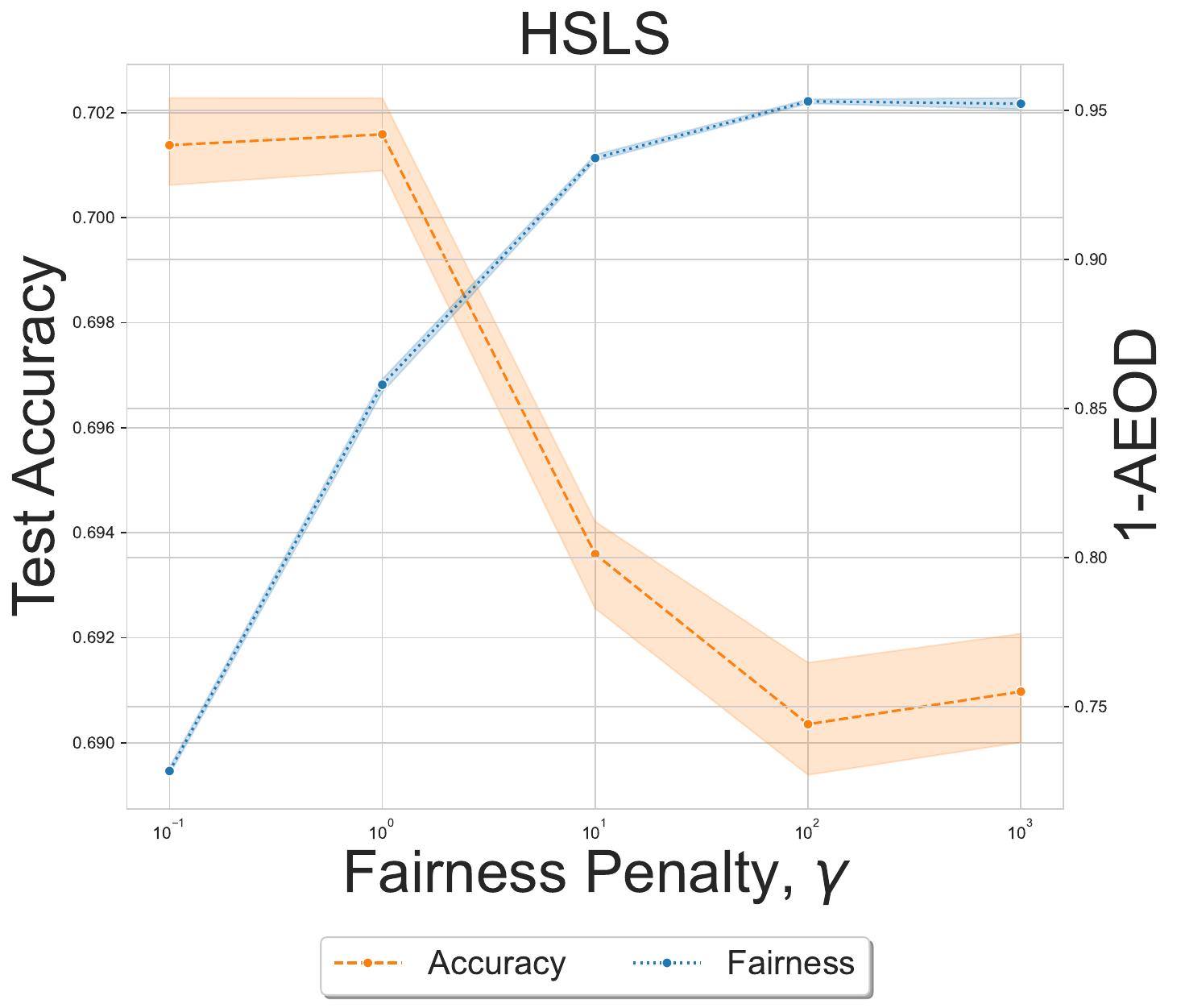}
     \end{subfigure}
\end{figure}

\subsection{Figures for various fairness metrics.}

Below we report figures for Demographic Parity, Equality of Opportunity, and Subgroup Accuracy. These results reflect the results shown in the main body of the paper, with the exception of Subgroup Accuracy which has more variable results, and sometimes worse results for fair methods. However, this is because our fairness regularizer optimizes for group fairness metrics such as Equalized Odds, Equality of Opportunity, and Demographic Parity which are metrics based on individual true and false positive rates rather than the raw test accuracy of each subgroup. 

\begin{figure}[h]
    \centering
    \begin{subfigure}{0.3\linewidth}
         \centering
         \includegraphics[width=\linewidth]{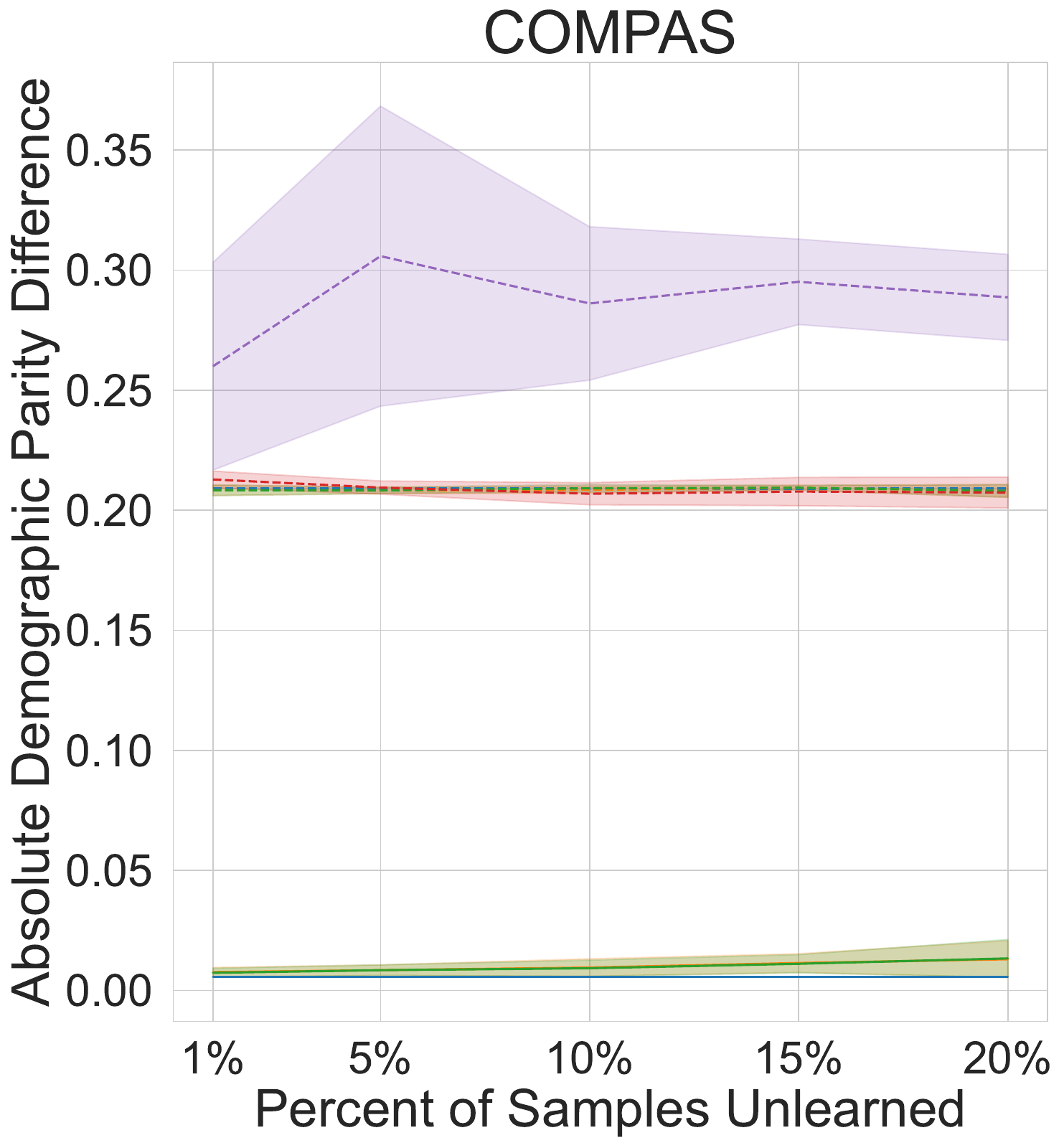}
     \end{subfigure}
     \hfill
     \begin{subfigure}{0.3\linewidth}
         \centering
         \includegraphics[width=\linewidth]{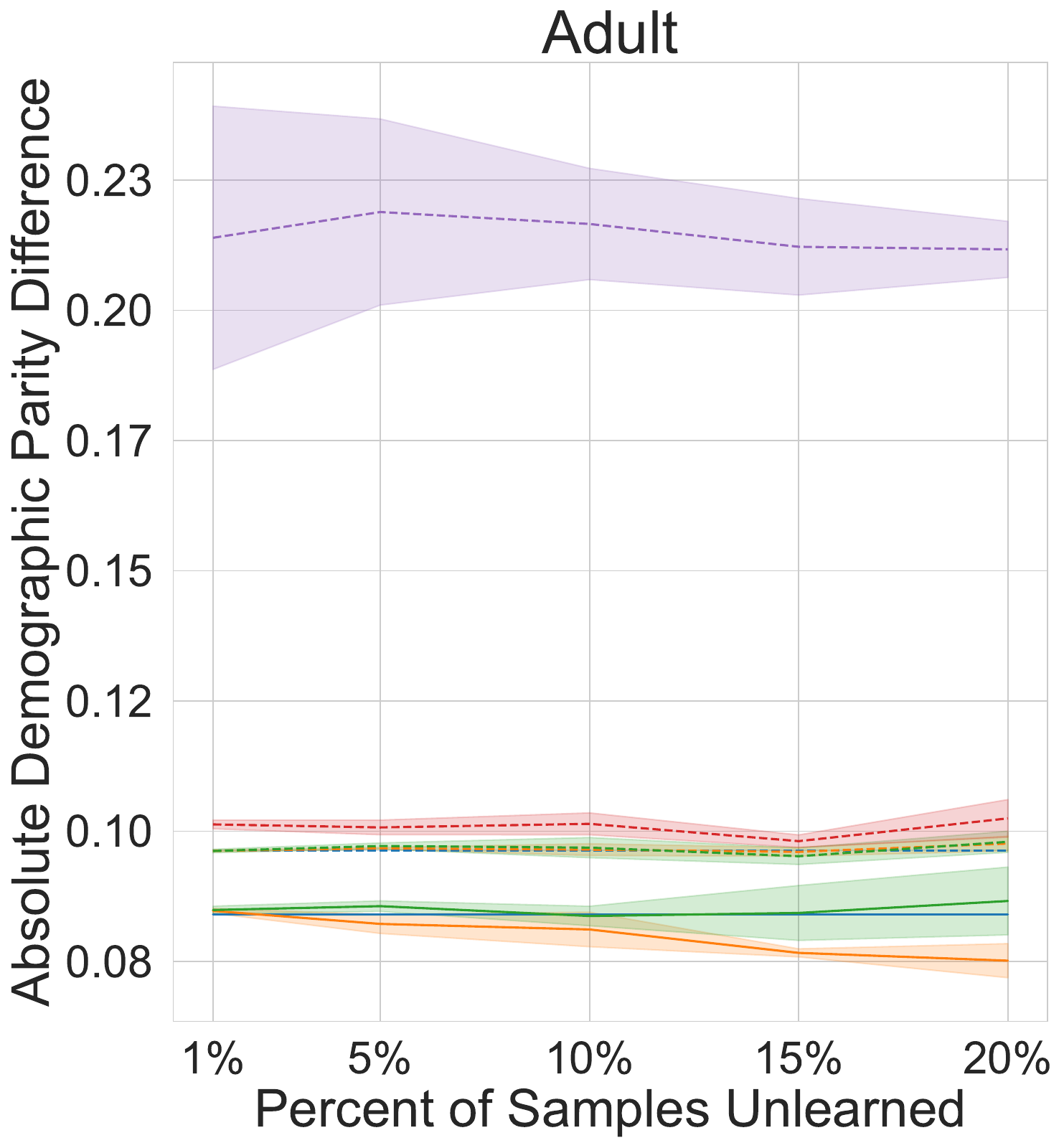}
     \end{subfigure}
     \hfill
     \begin{subfigure}{0.3\linewidth}
         \centering
         \includegraphics[width=\linewidth]{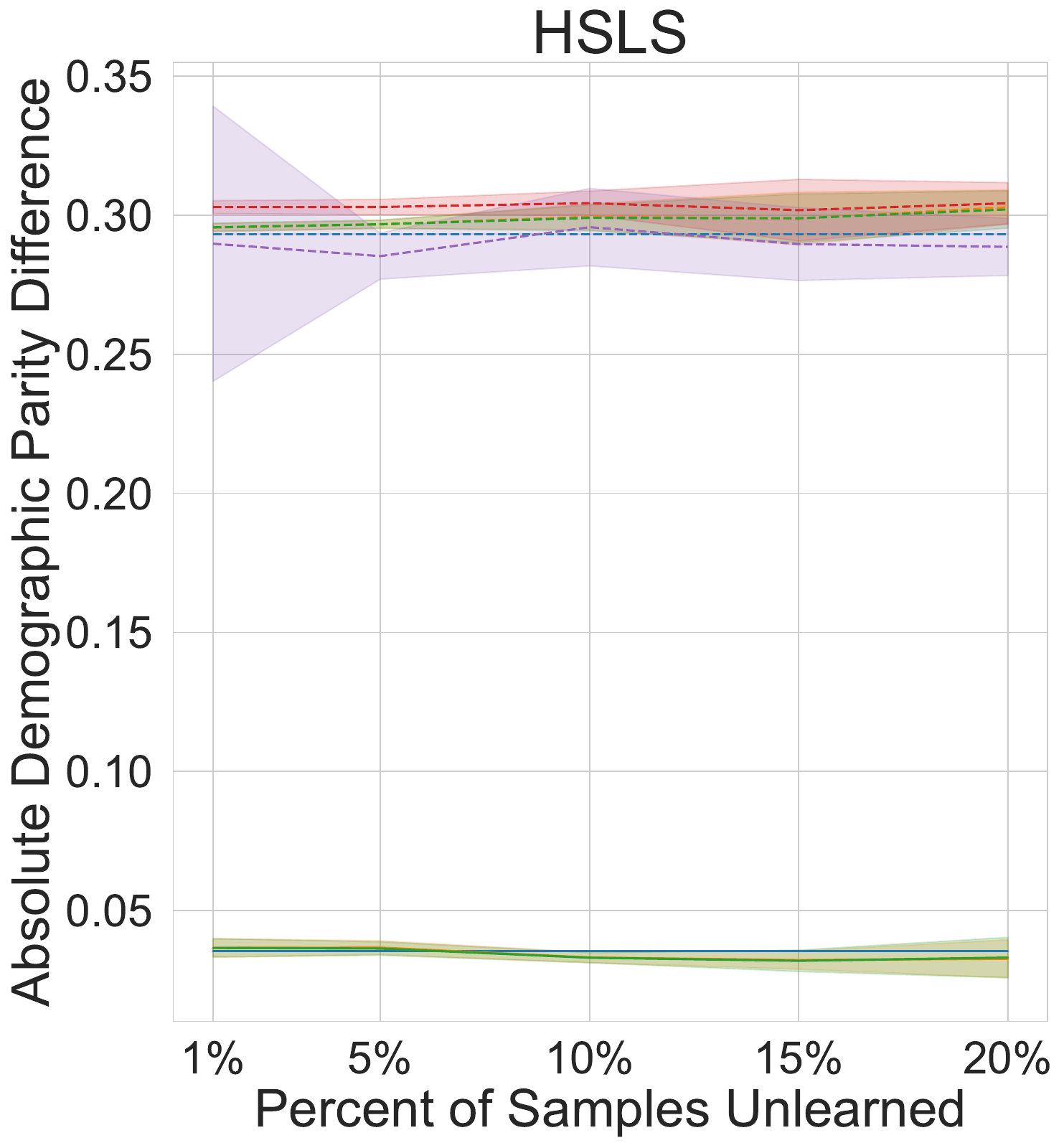}
     \end{subfigure}
     \\
     \begin{subfigure}{0.3\linewidth}
         \centering
         \includegraphics[width=\linewidth]{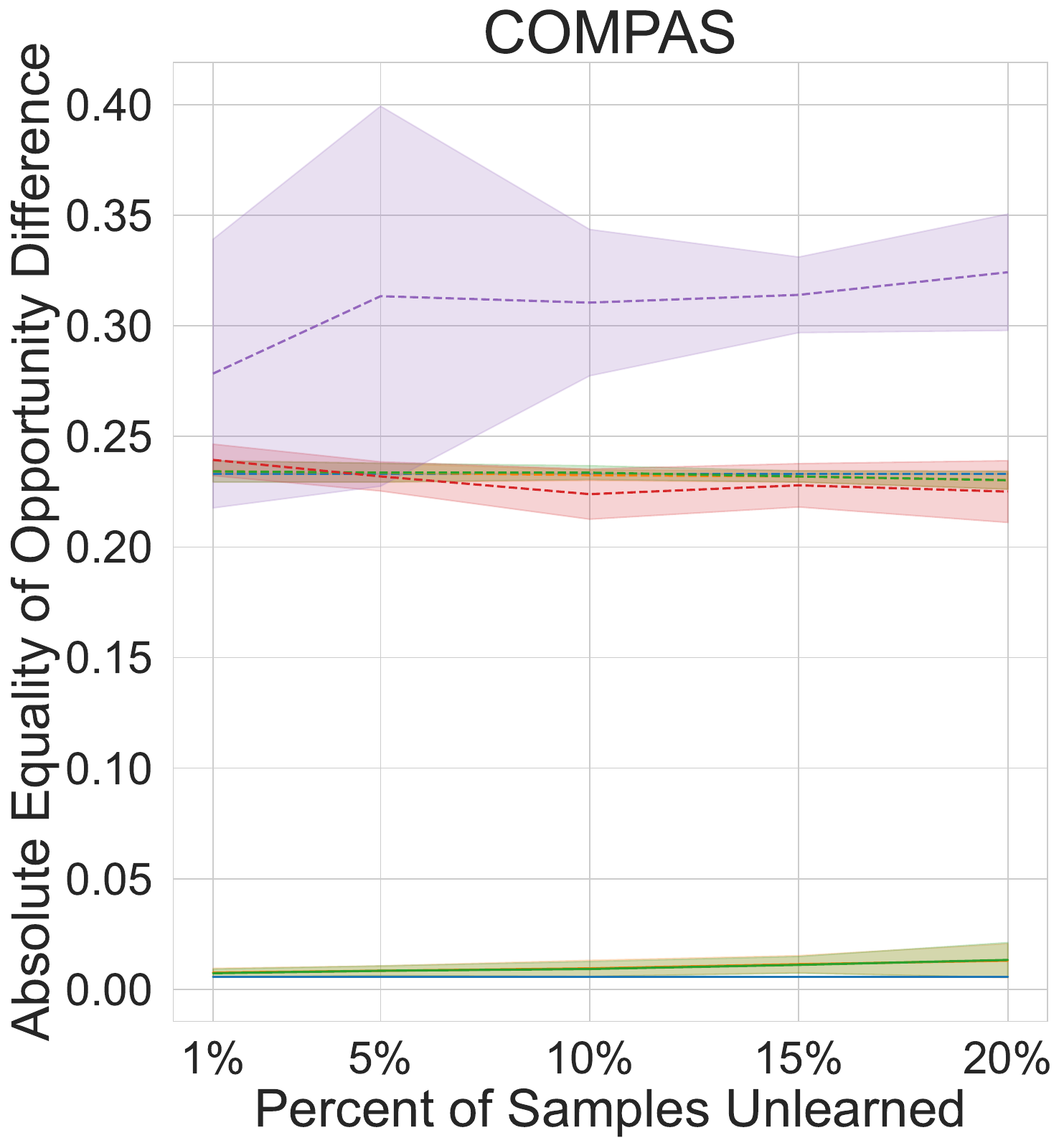}
     \end{subfigure}
     \hfill
     \begin{subfigure}{0.3\linewidth}
         \centering
         \includegraphics[width=\linewidth]{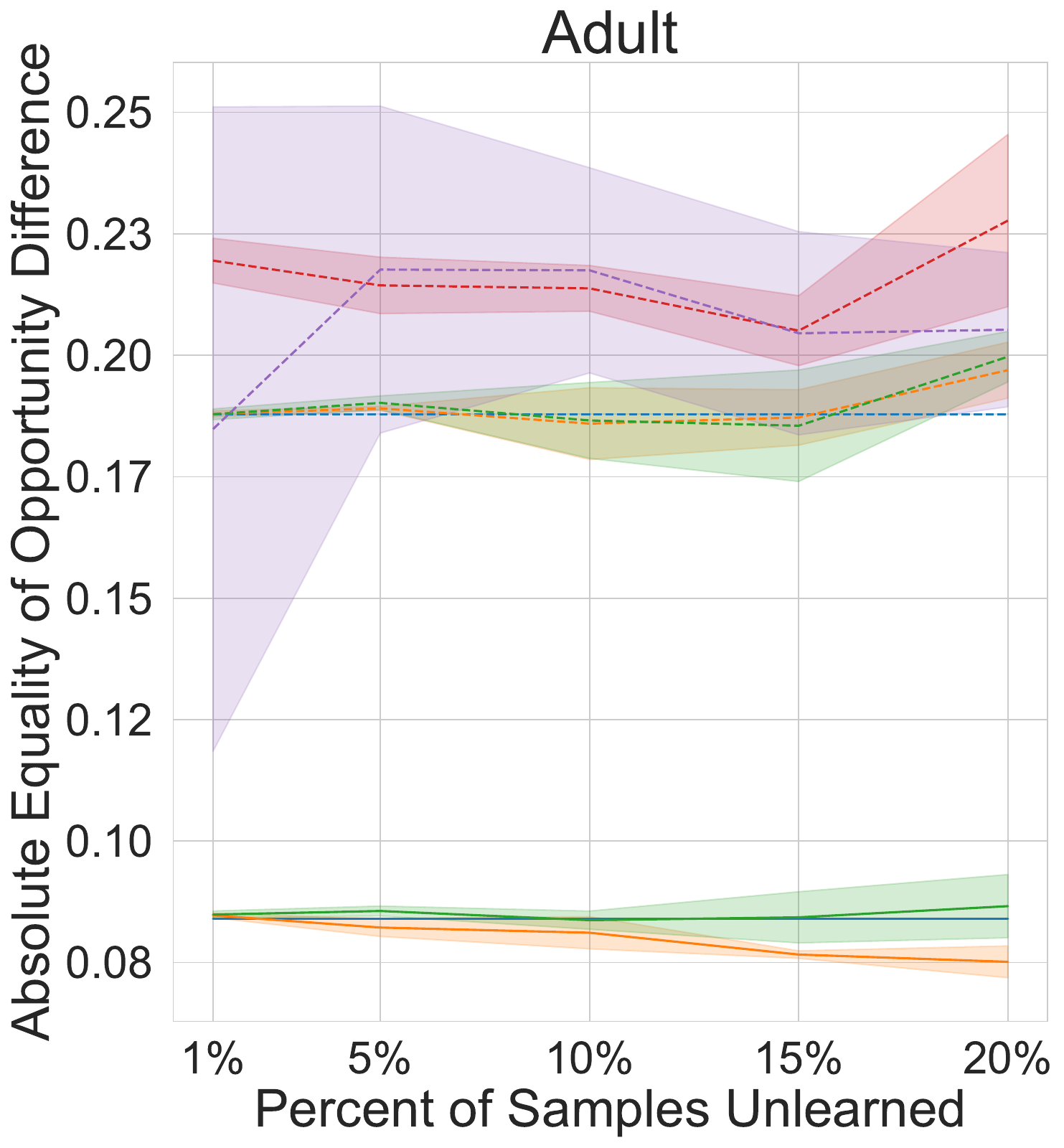}
     \end{subfigure}
     \hfill
     \begin{subfigure}{0.3\linewidth}
         \centering
         \includegraphics[width=\linewidth]{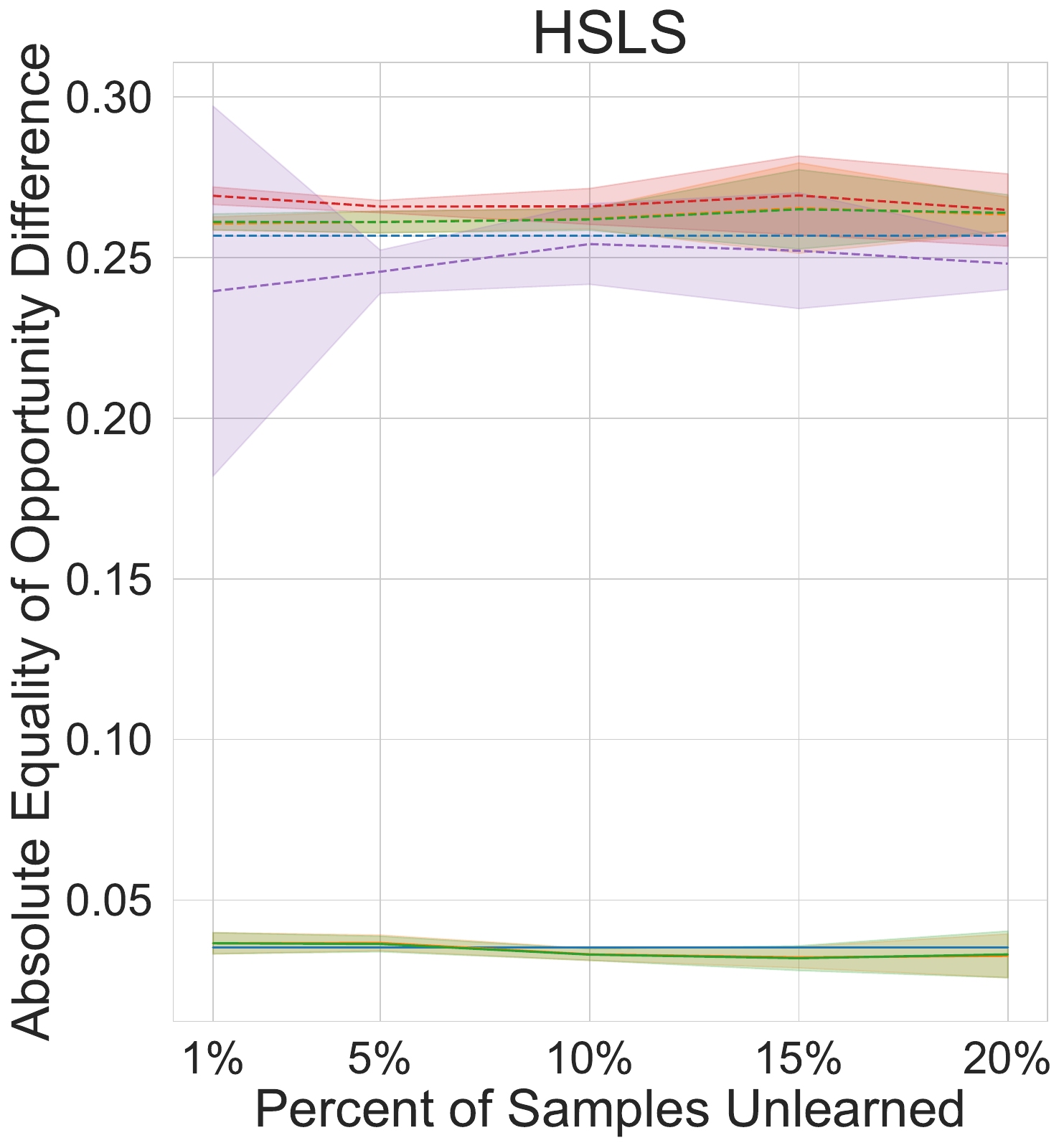}
     \end{subfigure}
     \\
     \begin{subfigure}{0.3\linewidth}
         \centering
         \includegraphics[width=\linewidth]{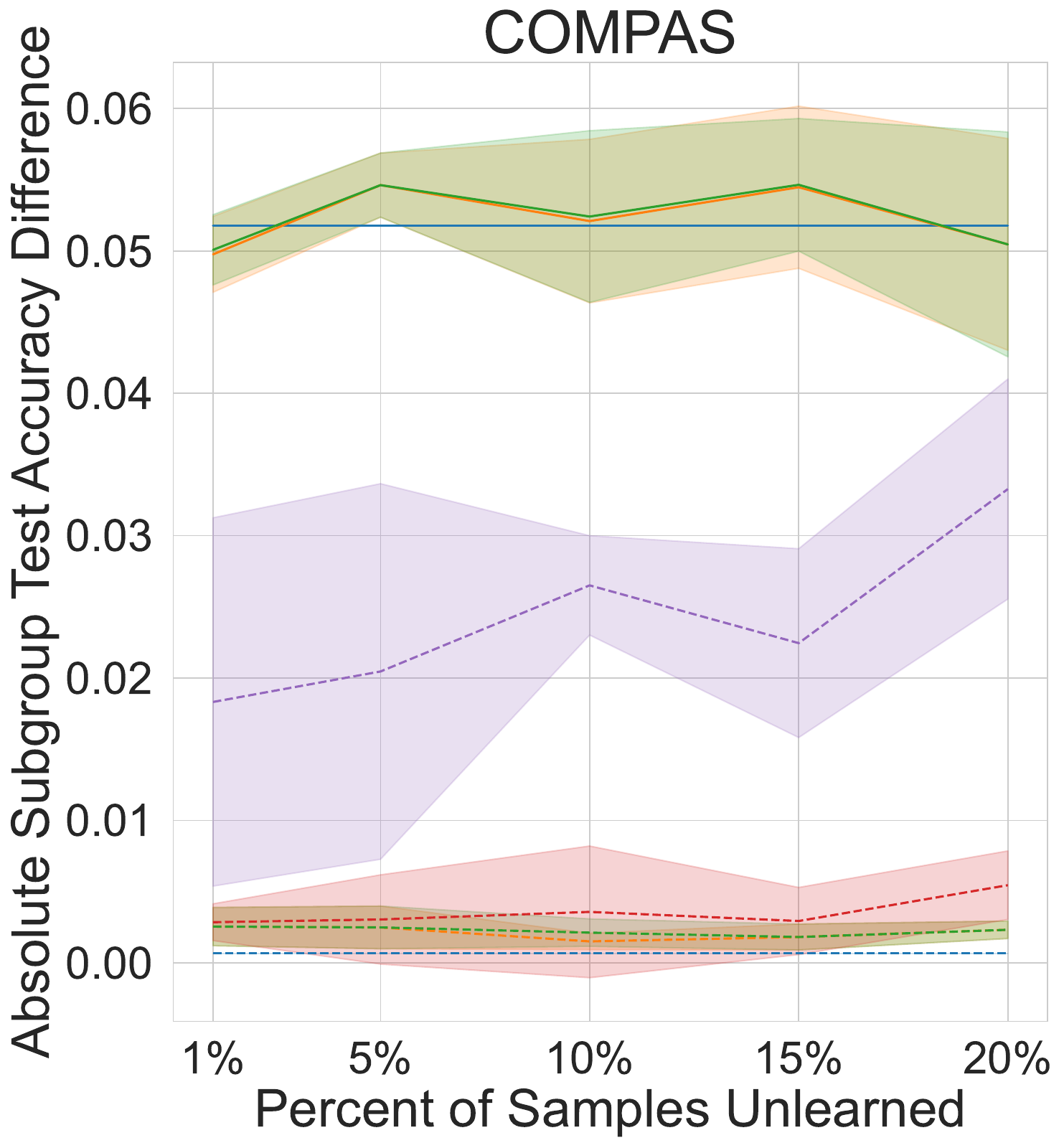}
     \end{subfigure}
     \hfill
     \begin{subfigure}{0.3\linewidth}
         \centering
         \includegraphics[width=\linewidth]{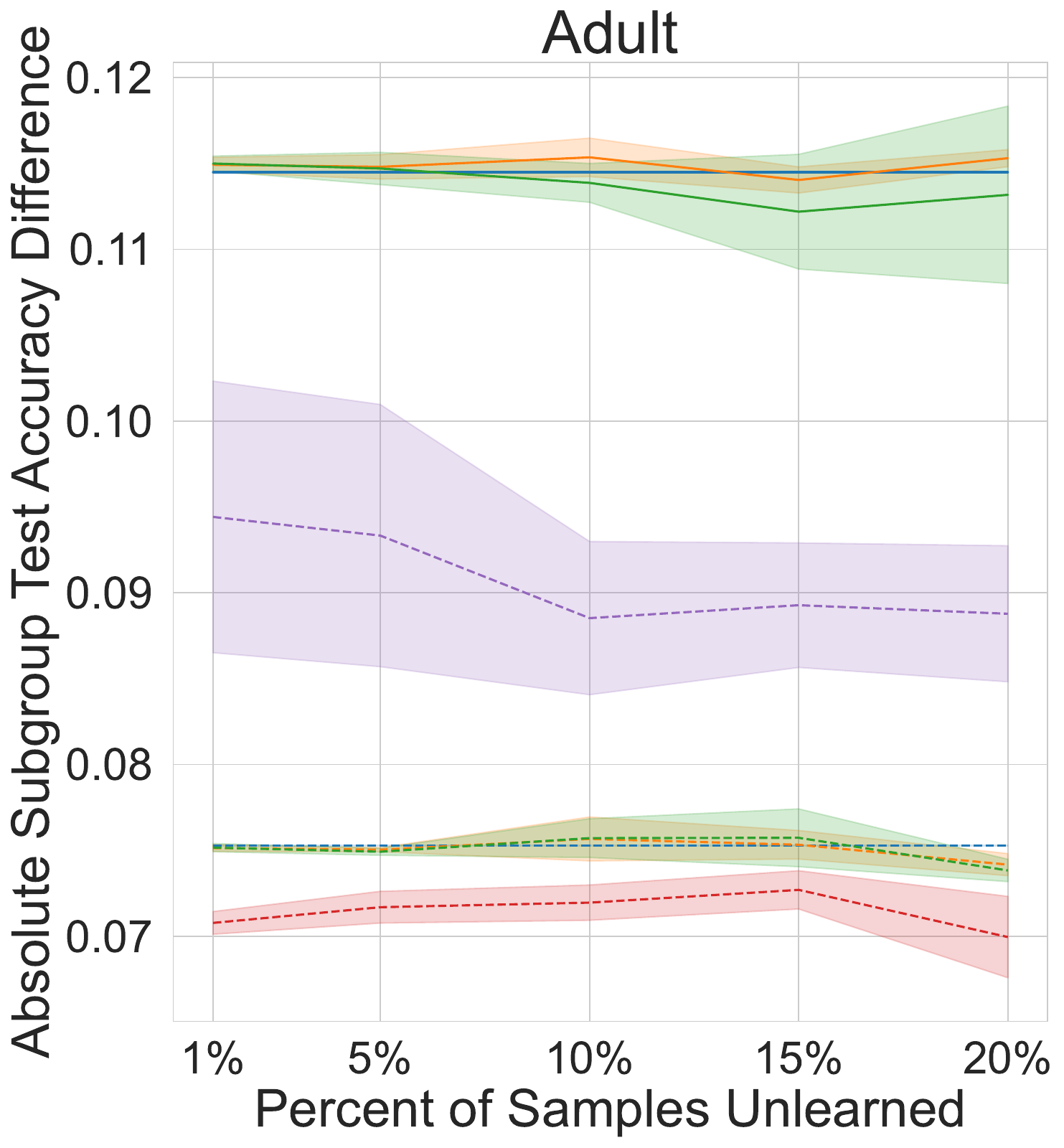}
     \end{subfigure}
     \hfill
     \begin{subfigure}{0.3\linewidth}
         \centering
         \includegraphics[width=\linewidth]{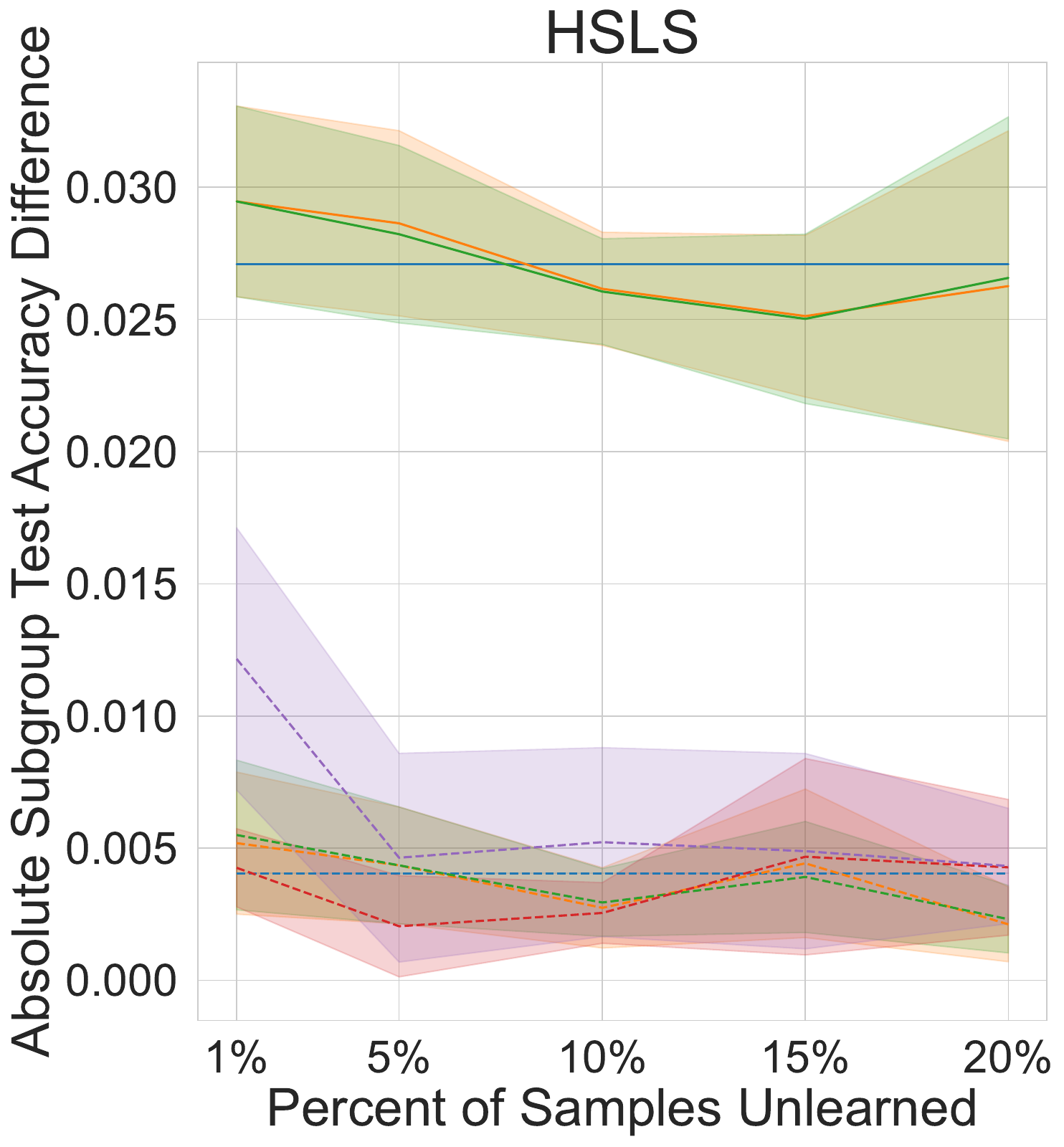}
     \end{subfigure}
     \includegraphics[width=0.8\linewidth]{figures/legend.png}
        \caption{Absolute demographic parity (top), equality of opportunity (middle), and subgroup test accuracy (bottom) differences (lower is better) for unlearning methods over random requests on COMPAS, Adult, and HSLS.}
        \label{fig:appdx unlearning at random}
\end{figure}

\begin{figure}[h]
    \centering
    \begin{subfigure}{0.3\linewidth}
         \centering
         \includegraphics[width=\linewidth]{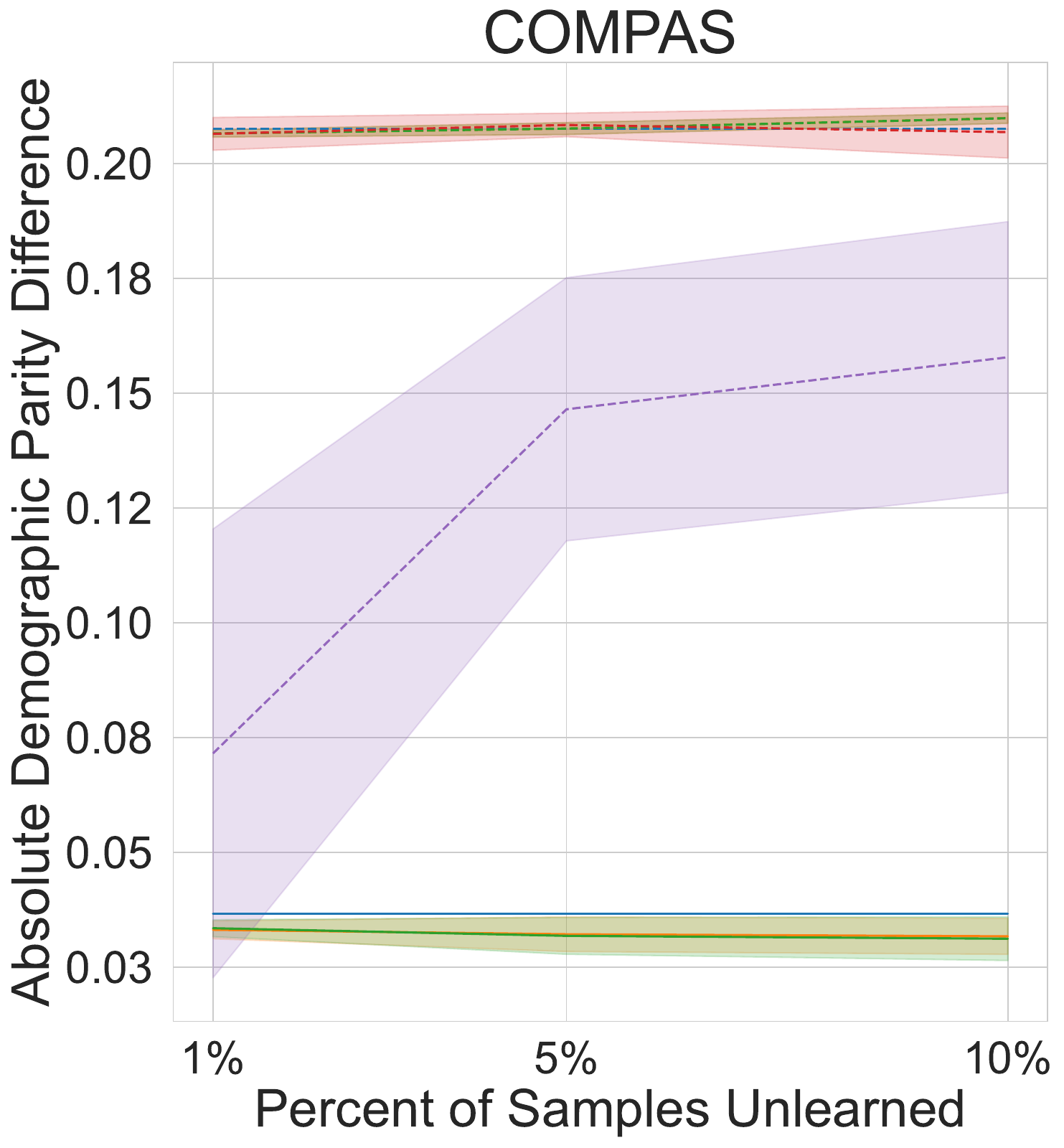}
     \end{subfigure}
     \hfill
     \begin{subfigure}{0.3\linewidth}
         \centering
         \includegraphics[width=\linewidth]{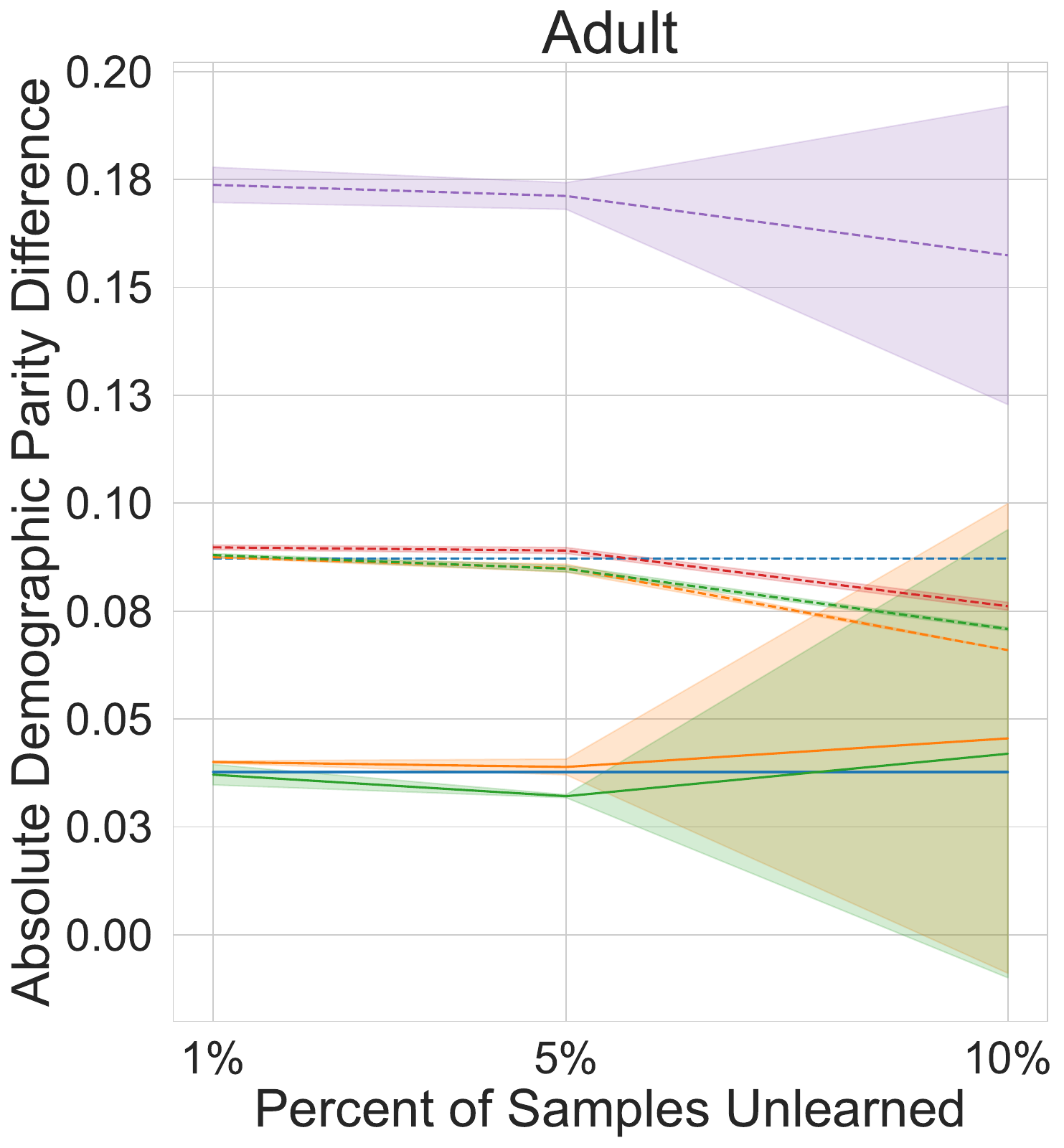}
     \end{subfigure}
     \hfill
     \begin{subfigure}{0.3\linewidth}
         \centering
         \includegraphics[width=\linewidth]{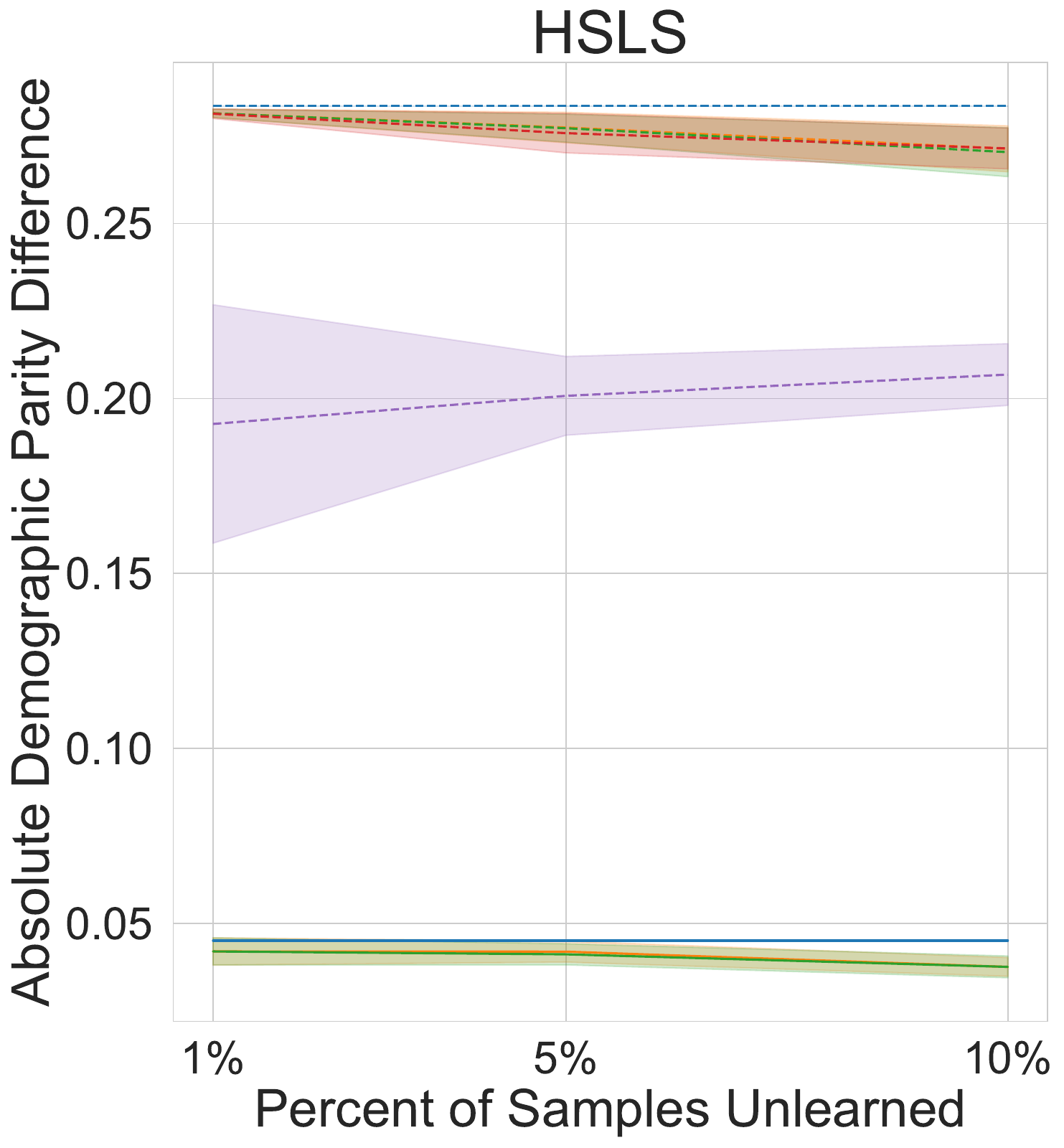}
     \end{subfigure}
     \\
     \begin{subfigure}{0.3\linewidth}
         \centering
         \includegraphics[width=\linewidth]{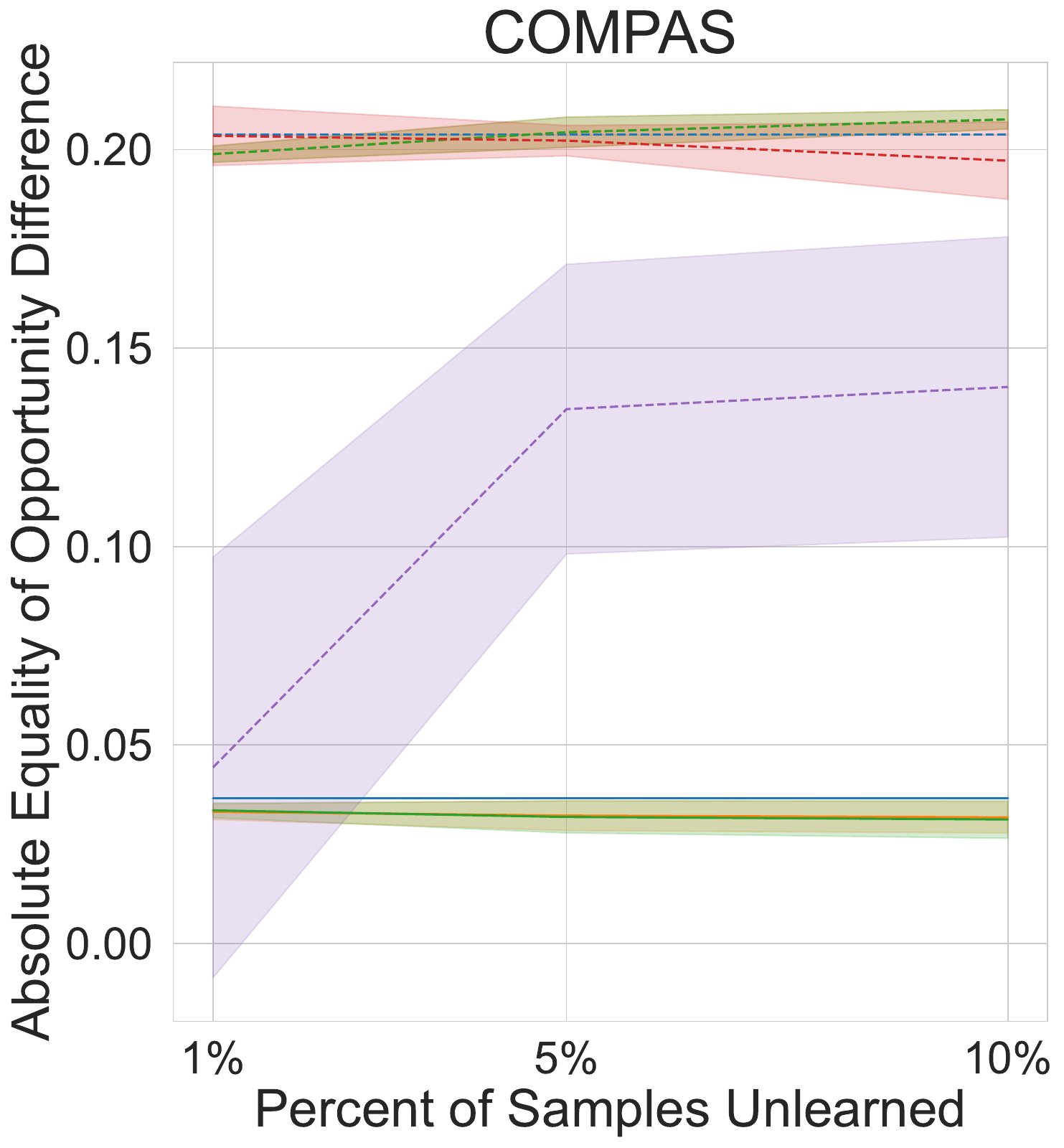}
     \end{subfigure}
     \hfill
     \begin{subfigure}{0.3\linewidth}
         \centering
         \includegraphics[width=\linewidth]{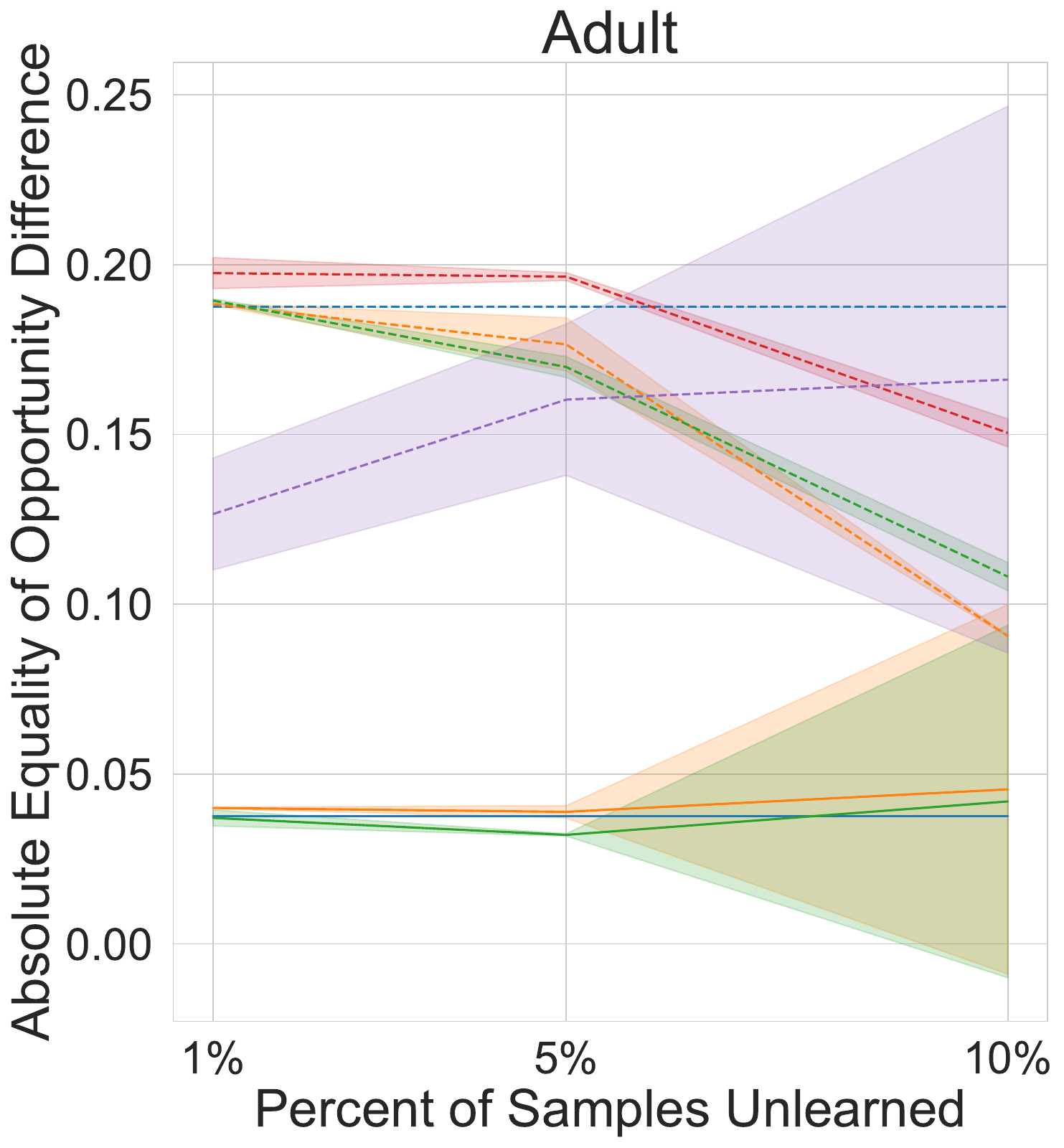}
     \end{subfigure}
     \hfill
     \begin{subfigure}{0.3\linewidth}
         \centering
         \includegraphics[width=\linewidth]{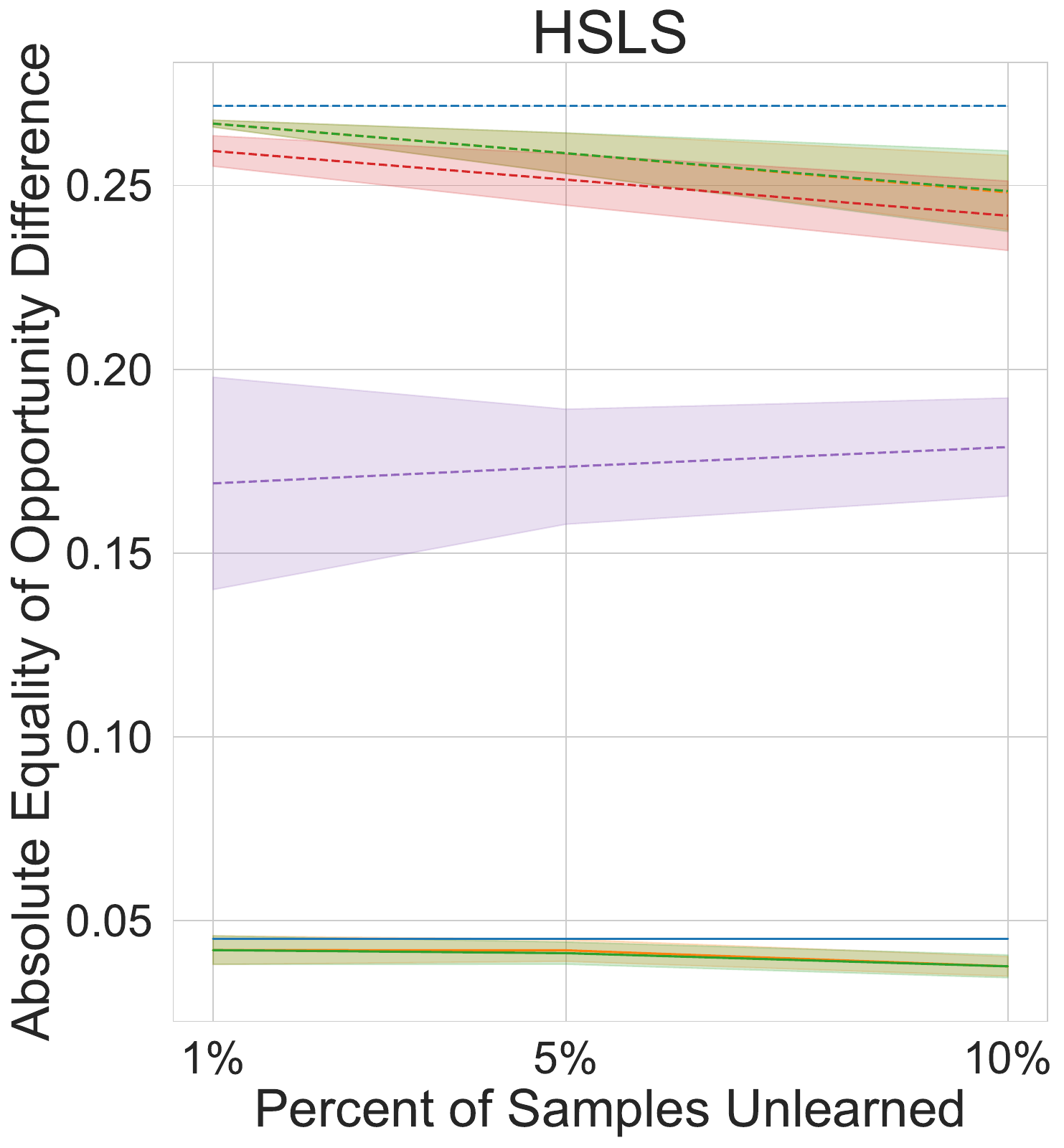}
     \end{subfigure}
     \\
     \begin{subfigure}{0.3\linewidth}
         \centering
         \includegraphics[width=\linewidth]{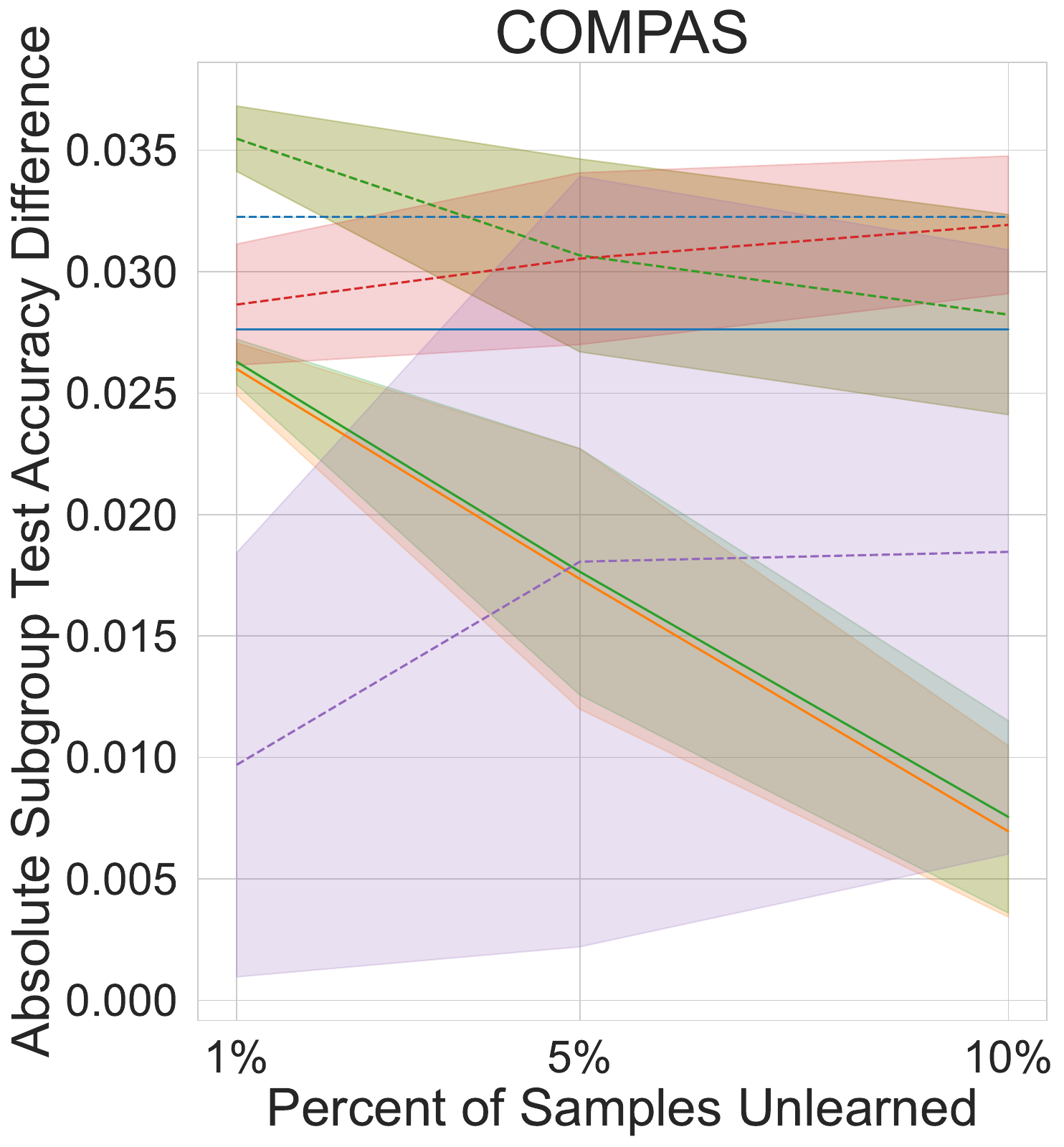}
     \end{subfigure}
     \hfill
     \begin{subfigure}{0.3\linewidth}
         \centering
         \includegraphics[width=\linewidth]{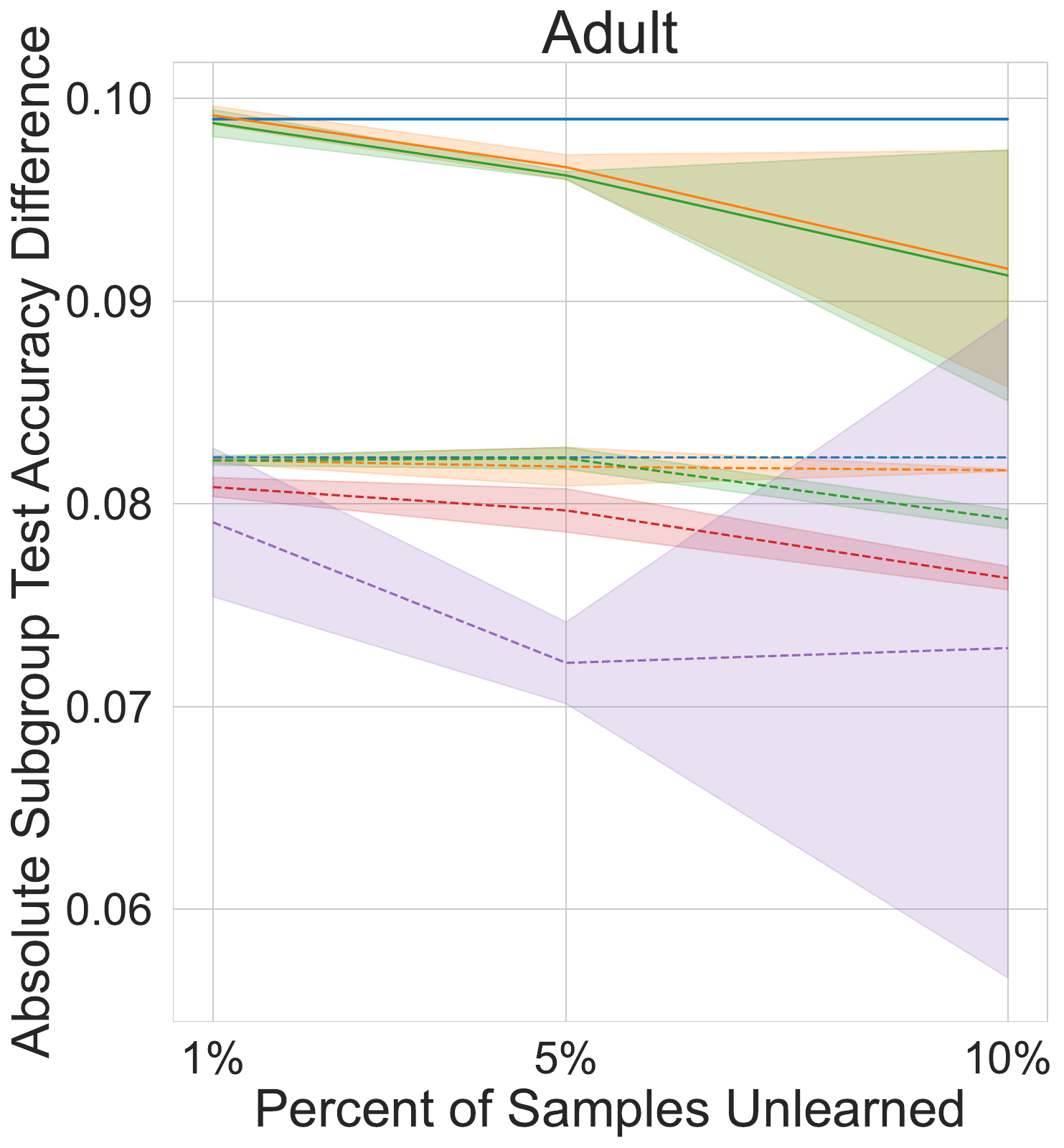}
     \end{subfigure}
     \hfill
     \begin{subfigure}{0.3\linewidth}
         \centering
         \includegraphics[width=\linewidth]{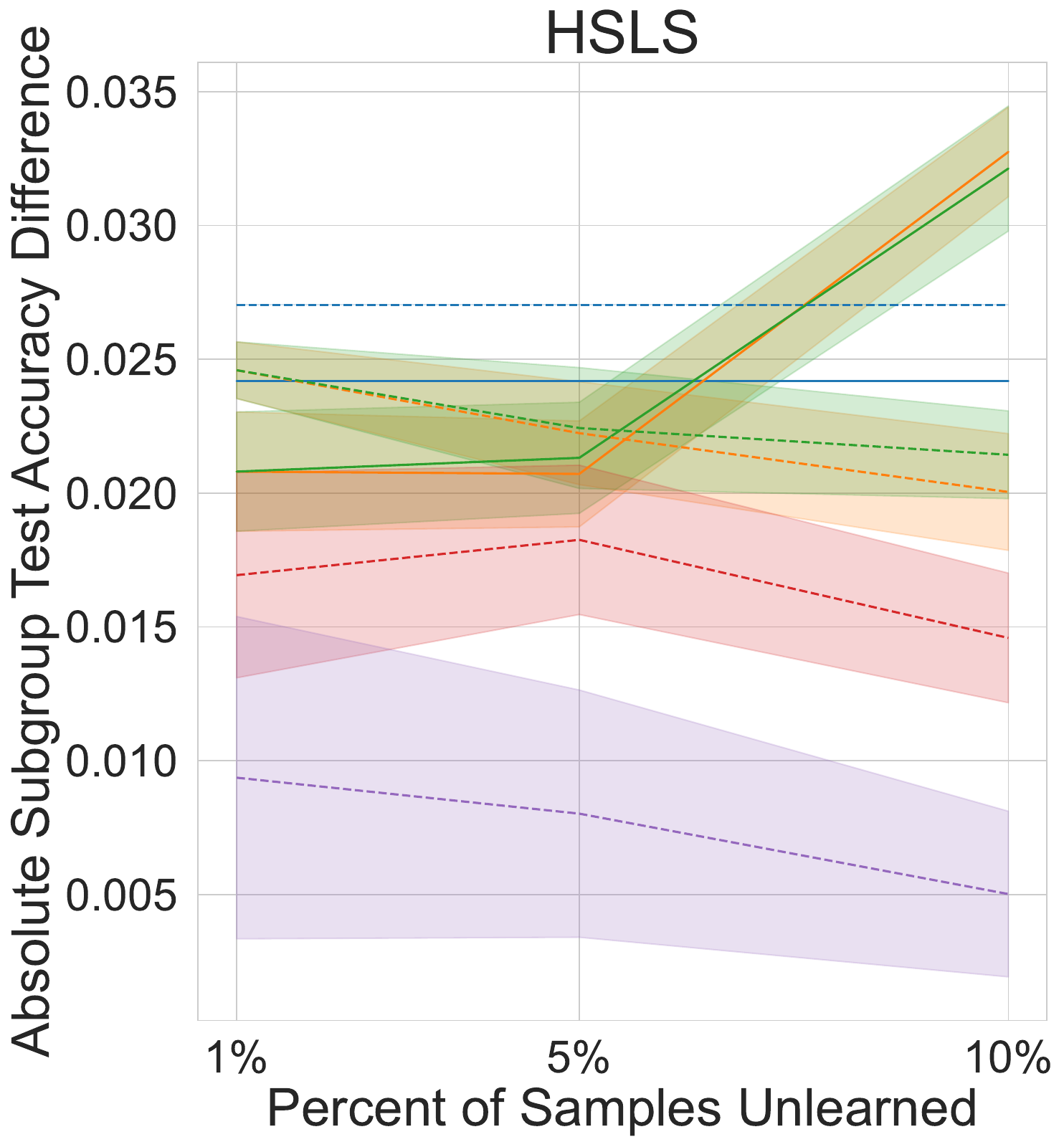}
     \end{subfigure}
     \includegraphics[width=0.8\linewidth]{figures/legend.png}
        \caption{Absolute demographic parity (top), equality of opportunity (middle), and subgroup test accuracy (bottom) differences (lower is better) for unlearning methods when unlearning from the minority subgroup on COMPAS, Adult, and HSLS.}
        \label{fig:appdx unlearning from minorities}
\end{figure}
\newpage

\begin{figure}[h]
    \centering
    \begin{subfigure}{0.3\linewidth}
         \centering
         \includegraphics[width=\linewidth]{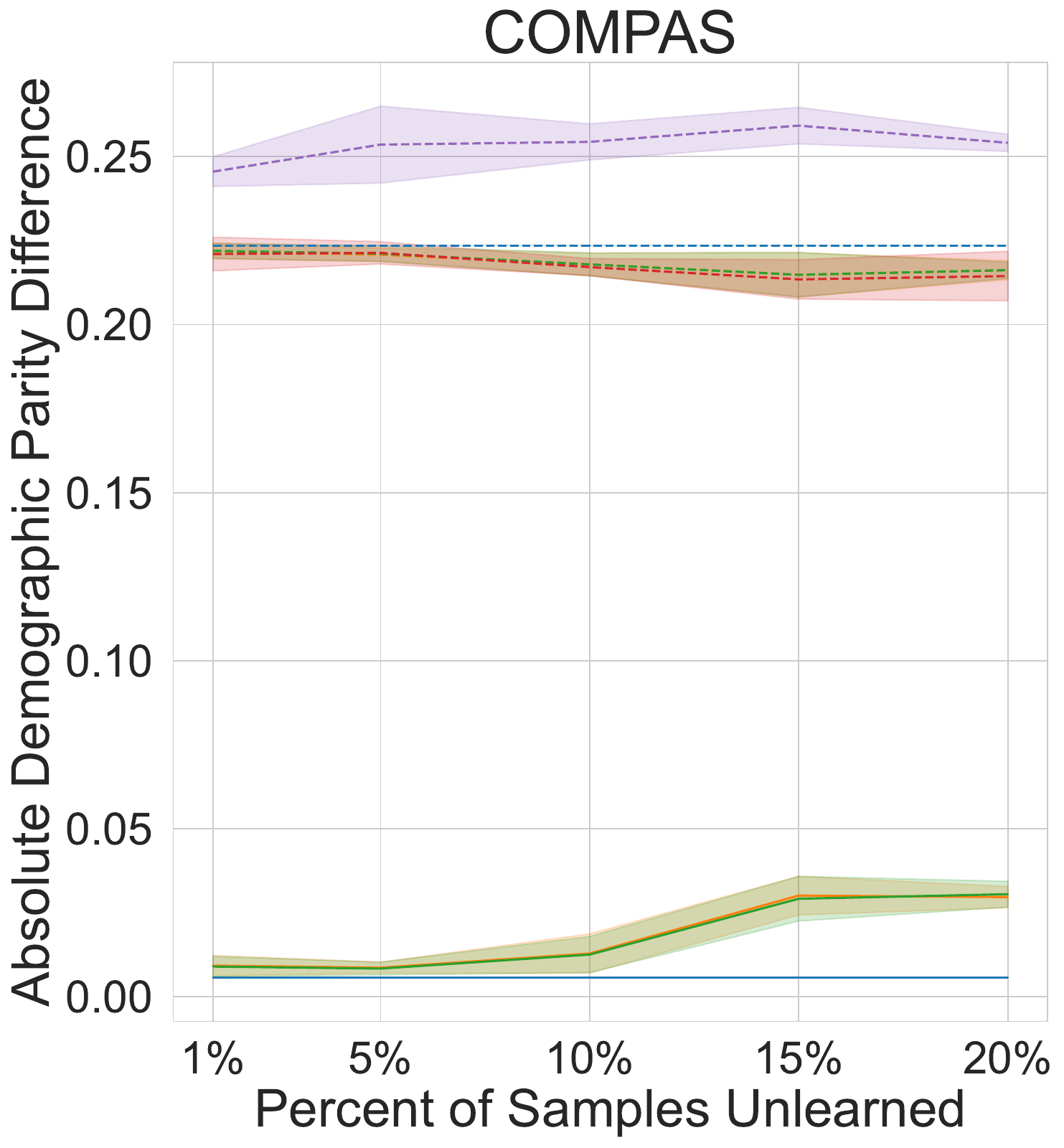}
     \end{subfigure}
     \hfill
     \begin{subfigure}{0.3\linewidth}
         \centering
         \includegraphics[width=\linewidth]{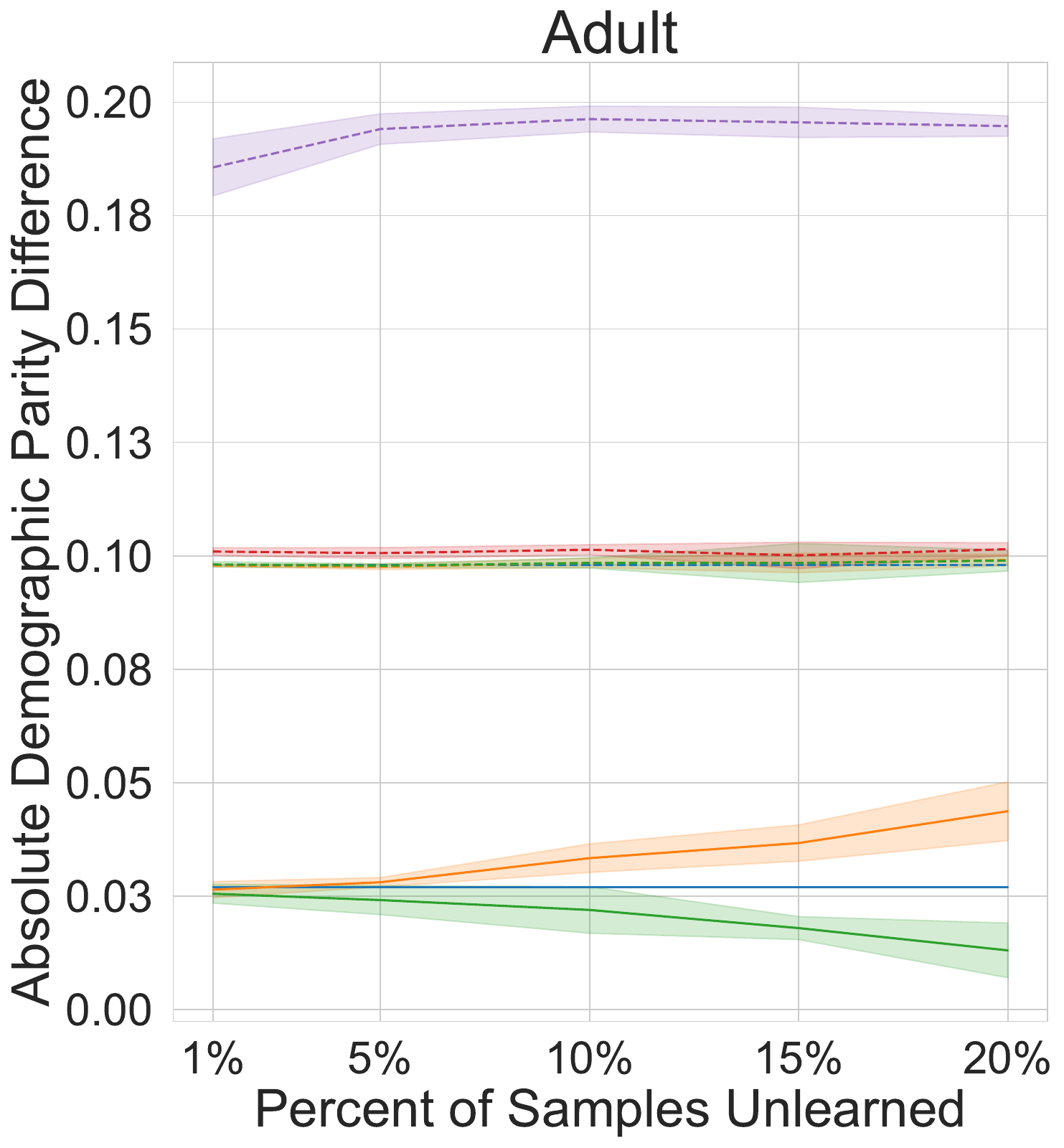}
     \end{subfigure}
     \hfill
     \begin{subfigure}{0.3\linewidth}
         \centering
         \includegraphics[width=\linewidth]{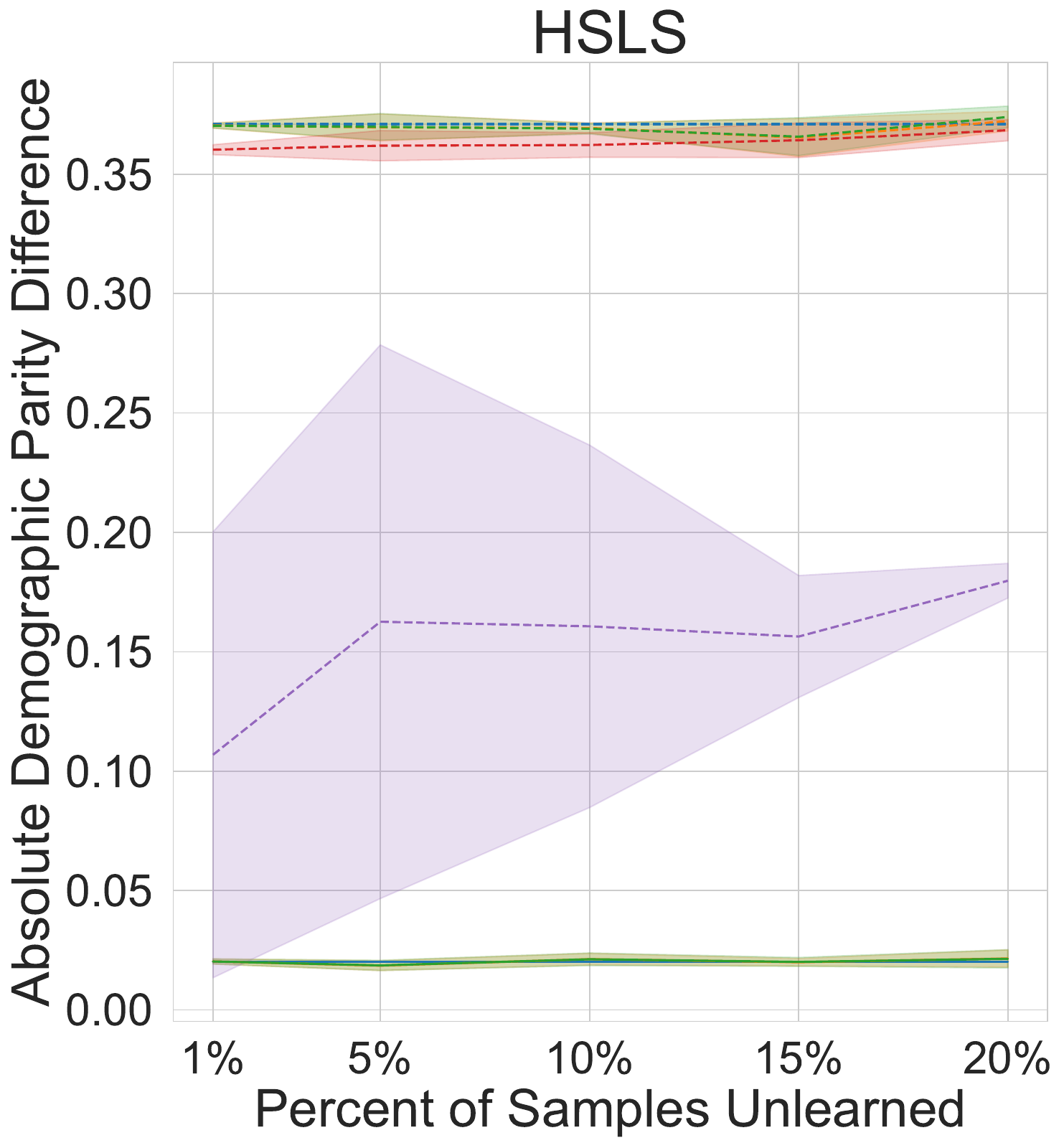}
     \end{subfigure}
     \\
     \begin{subfigure}{0.3\linewidth}
         \centering
         \includegraphics[width=\linewidth]{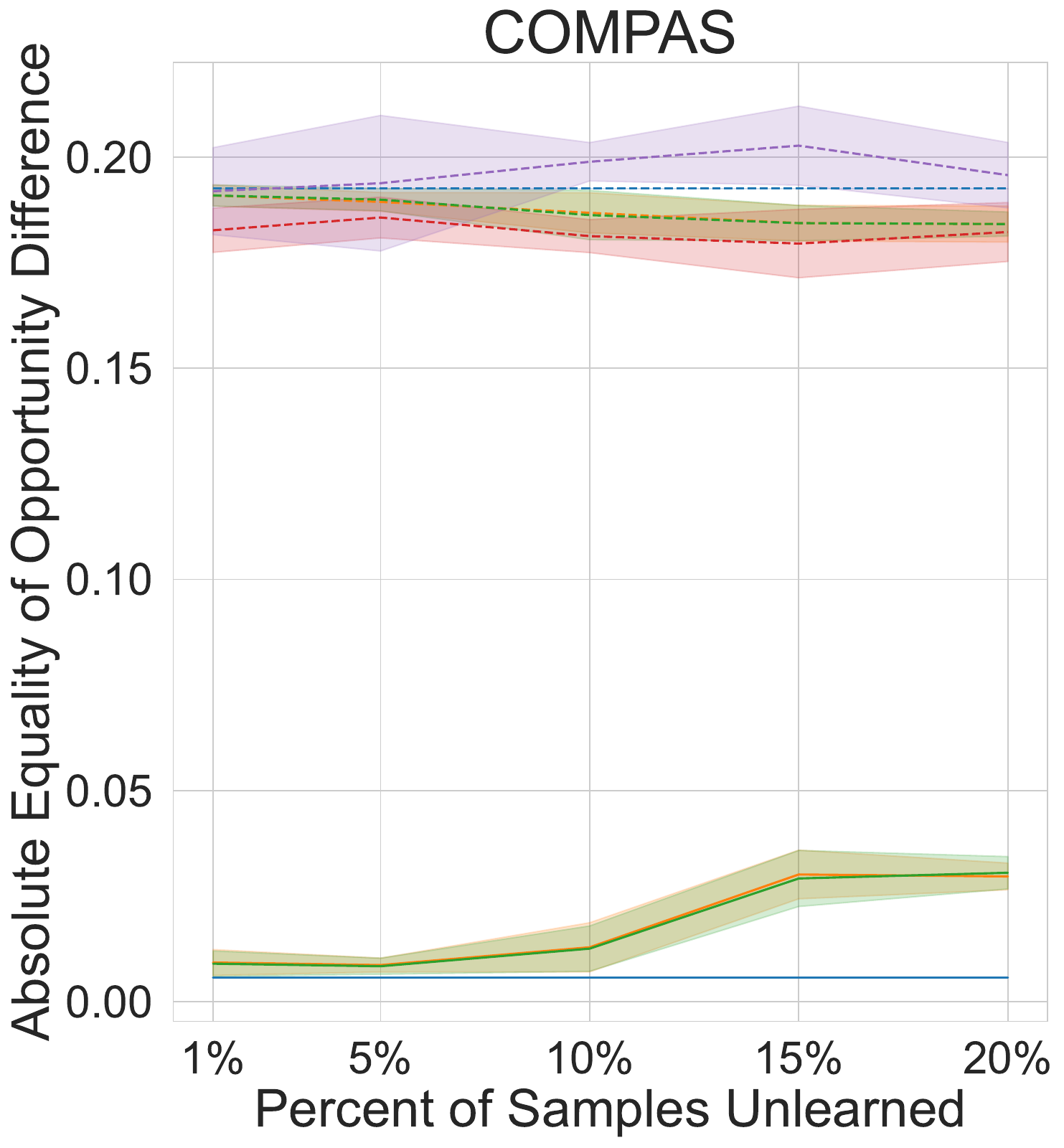}
     \end{subfigure}
     \hfill
     \begin{subfigure}{0.3\linewidth}
         \centering
         \includegraphics[width=\linewidth]{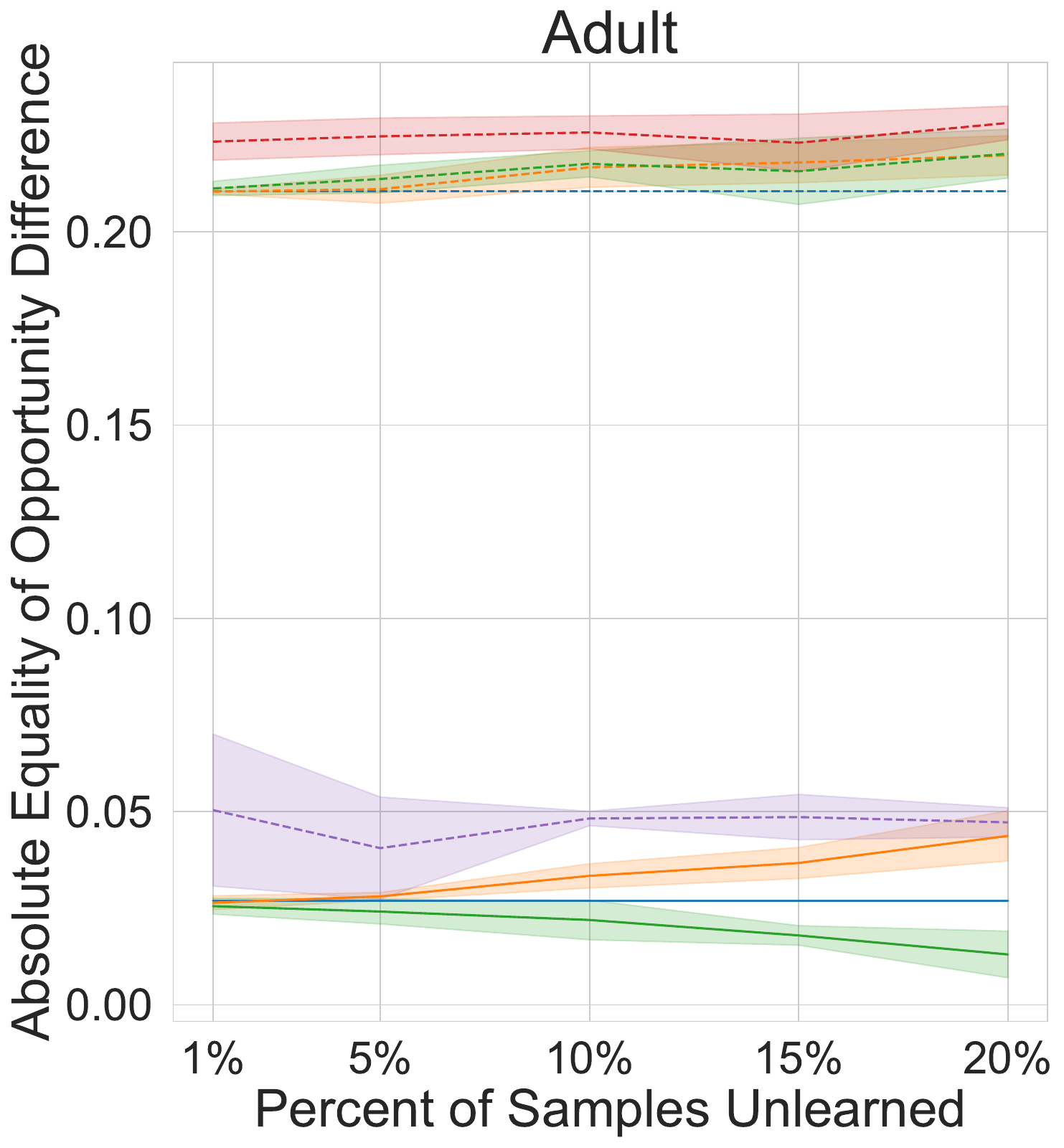}
     \end{subfigure}
     \hfill
     \begin{subfigure}{0.3\linewidth}
         \centering
         \includegraphics[width=\linewidth]{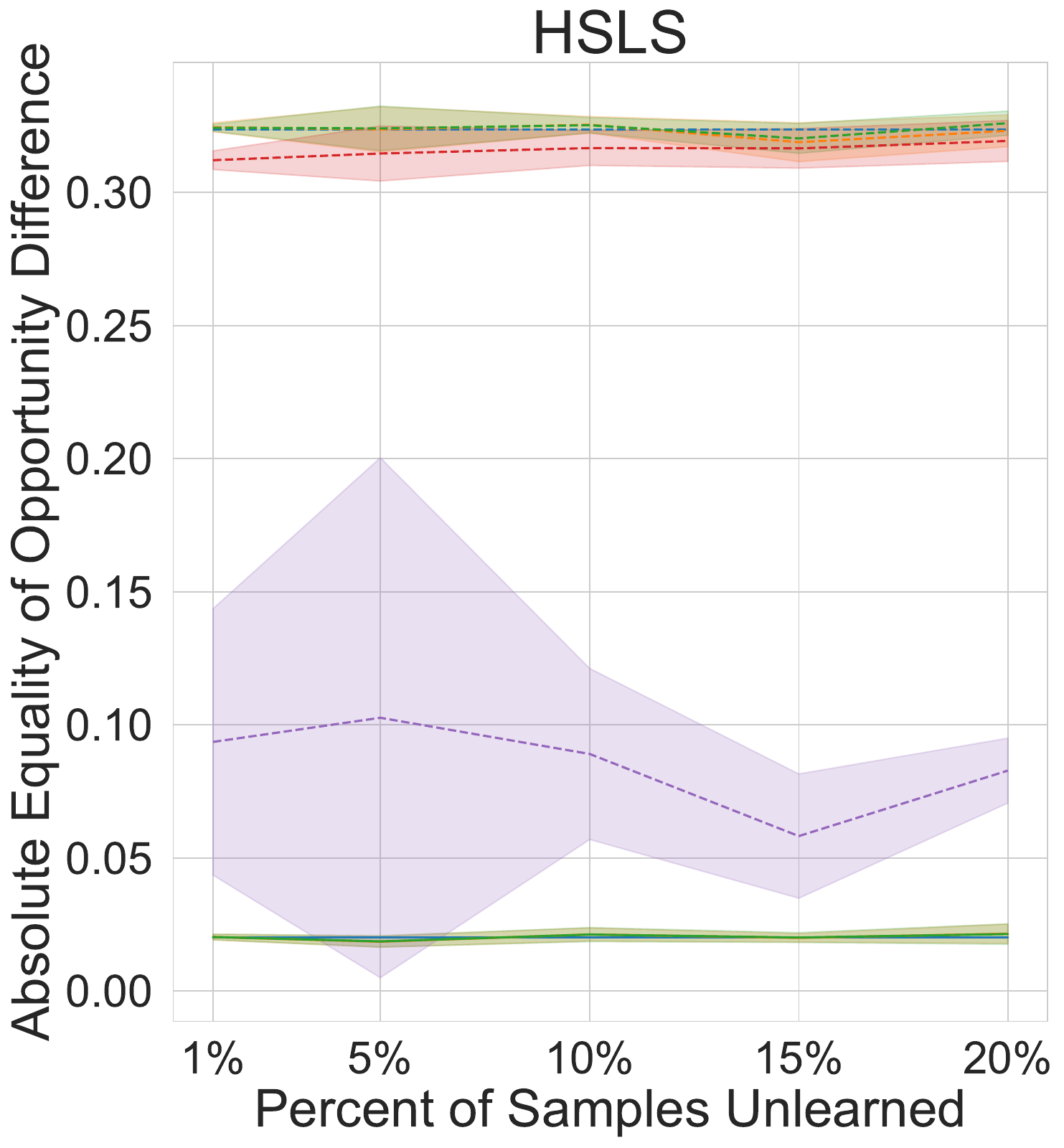}
     \end{subfigure}
     \\
     \begin{subfigure}{0.3\linewidth}
         \centering
         \includegraphics[width=\linewidth]{figures/COMPAS_fullbaselines_equality_of_opp_race_majority.pdf}
     \end{subfigure}
     \hfill
     \begin{subfigure}{0.3\linewidth}
         \centering
         \includegraphics[width=\linewidth]{figures/Adult_fullbaselines_equality_of_opp_race_majority.pdf}
     \end{subfigure}
     \hfill
     \begin{subfigure}{0.3\linewidth}
         \centering
         \includegraphics[width=\linewidth]{figures/HSLS_fullbaselines_equality_of_opp_race_majority.pdf}
     \end{subfigure}
     \includegraphics[width=0.8\linewidth]{figures/legend.png}
        \caption{Absolute demographic parity (top), equality of opportunity (middle), and subgroup test accuracy (bottom) differences (lower is better) for unlearning methods when unlearning from the majority subgroup on COMPAS, Adult, and HSLS.}
        \label{fig:appdx unlearning from majorities}
\end{figure}

\end{document}